%% file: arxiv_unite_full.tex
\documentclass[letterpaper,journal]{IEEEtran}
\RequirePackage[T1]{fontenc}

\usepackage{algorithm}
\usepackage{algpseudocode}
\input{math_commands.tex}

\usepackage{booktabs}
\usepackage[table]{xcolor}
\usepackage{graphicx}
\usepackage{array}
\usepackage{multirow}
\usepackage{adjustbox}
\usepackage{stfloats}
\usepackage{float}
\usepackage{placeins}
\usepackage{flushend}
\usepackage[nocompress]{cite}
\usepackage{url}
\usepackage[hidelinks,pdfversion=1.7]{hyperref}
\usepackage{titletoc}
\usepackage{tocloft}
\graphicspath{{figures/}}

\newif\ifuntkeep
\ifdefined\unttakesupplementonly
  \makeatletter
  \let\unt@deferred@write\protected@write
  \long\def\protected@write#1#2#3{%
    \begingroup
      \let\thepage\relax
      #2%
      \let\protect\@unexpandable@protect
      \edef\reserved@a{\immediate\write#1{#3}}\reserved@a
    \endgroup
    \if@nobreak\ifvmode\nobreak\fi\fi}
  \makeatother
  \AddToHook{shipout/before}{\ifuntkeep\else\DiscardShipoutBox\fi}
\fi

\title{UNITE: Consolidating Embodied VLM Specialists via Shared-Anchor One-Bit Factorization}
\author{Tao Tao, Anke Tang, Enneng Yang, Zhi Hou, Xiaogang Wang, and Dacheng Tao
  \IEEEcompsocitemizethanks{
    \IEEEcompsocthanksitem Tao Tao is with Sichuan University, Chengdu, China, and ACE Robotics, Shanghai, China. (e-mail: 2022141430130@stu.scu.edu.cn).
    \IEEEcompsocthanksitem Zhi Hou and Xiaogang Wang are with ACE Robotics, Shanghai, China. (e-mail: houzhi91@gmail.com, xg@acerobotics.com).
    \IEEEcompsocthanksitem Anke Tang and Dacheng Tao are with Nanyang Technological University, Singapore. (e-mail: anke.tang@ntu.edu.sg, dacheng.tao@ntu.edu.sg).
    \IEEEcompsocthanksitem Enneng Yang is with Shenzhen Campus of Sun Yat-sen University, Shenzhen 518107, China. (e-mail: yangenn@mail.sysu.edu.cn).
    \IEEEcompsocthanksitem Corresponding authors: Anke Tang, Zhi Hou.
  }
}
\IEEEtitleabstractindextext{
  \input{section/abstract}
  }

\begin{document}
\raggedend
\maketitle
\IEEEdisplaynontitleabstractindextext
\IEEEpeerreviewmaketitle
\flushbottom
\setlength{\parskip}{0pt}
\AddToHook{selectfont}[unite-column-spacing]{%
  \setlength{\baselineskip}{\dimexpr\baselineskip\relax plus 0.1pt}}
\setlength{\baselineskip}{\dimexpr\baselineskip\relax plus 0.1pt}

\input{section/introduction}
\input{section/related_work}
\input{section/method}
\input{section/experiments}
\input{section/conclusion}

\nocite{%
  song2025robospatial,zhou2025roborefer,cheng2025pointarena,roboafford2025,%
  ji2025robobrain,yang2025vsibench,yang2025mmsibench,wang2025mindcube,%
  du2024embspatial,ray2025sat,geminirobotics2025,azuma2022scanqa,%
  ma2023sqa3d,chen2021scan2cap,chen2019scanrefer,zhang2023multi3drefer,%
  zhang2025mmerealworld,cao2024maplm,driveaction2025,qian2024nuscenesqa,%
  park2025nuplanqa,marcu2024lingoqa,anderson2018r2r,openai2026gpt54,%
  comanici2025gemini25,anthropic2026sonnet46,gong2026acebrain,nvidia2026cosmos3,%
  yuan2026embodiedr15,dang2026rynnbrain,yang2025vlaser,hao2025mimoembodied,%
  zhang2025pelicanvl,luo2025vebrain,tan2026robobrain25,zhang2024navid,%
  zhang2024uninavid,cheng2025navila,wei2026streamvln,zhang2026navfom,%
  wang2026qwenvla%
}

\FloatBarrier
\flushend
\bibliographystyle{IEEEtran}
\bibliography{arxiv_unite_bib}
\clearpage\untkeeptrue
\raggedend
\RemoveFromHook{selectfont}[unite-column-spacing]
\setlength{\parskip}{0ex plus 0.2ex minus 0.1ex}
\setlength{\baselineskip}{\dimexpr\baselineskip\relax}
\ifdefined\unttakesupplementonly
  \makeatletter\let\protected@write\unt@deferred@write\makeatother
\fi
\input{section/supplement}
\end{document}

%% file: math_commands.tex
\usepackage{amsmath,amsfonts,bm}

\def\1{\bm{1}}

\DeclareMathAlphabet{\mathsfit}{\encodingdefault}{\sfdefault}{m}{sl}
\SetMathAlphabet{\mathsfit}{bold}{\encodingdefault}{\sfdefault}{bx}{n}



%% file: section/abstract.tex
\begin{abstract}
Embodied capabilities such as visual grounding, spatial reasoning, navigation, and driving understanding are often developed in separate vision-language specialists fine-tuned from a common pretrained model.
Deploying the specialists separately multiplies storage, whereas merging them into one dense model forces conflicting task updates to share parameters and loses part of each specialist's expertise.
We introduce UNITE, a framework that consolidates such specialists into a compact shared-backbone mixture of experts.
UNITE keeps the feed-forward block of each specialist as a routable expert and represents the expert bank through a shared-anchor one-bit factorization: a shared floating-point anchor plus expert-specific, row-scaled binary deviations.
The anchor is optimized jointly with the scales and signs through closed-form alternating updates, using only source weights and no calibration data.
Router adaptation then adjusts token-level expert selection with the backbone and experts frozen, updating fewer than $0.6$M parameters ($<0.01\%$ of the $8.8$B active parameters).
Built from four Qwen3-VL-8B specialists, the complete UNITE model occupies $20.3$ GB, $1.16\times$ the storage of one dense specialist and $59.6\%$ below the uncompressed MoE.
Across $22$ benchmarks and R2R Val-Unseen, UNITE matches six static merging methods in mean retention of best-specialist performance, leads all of them on Scan2Cap, ScanQA, and all four navigation metrics, and surpasses every source specialist on all five 3D scene understanding benchmarks (up to $+19.1$ on Scan2Cap) and on R2R navigation.
Relative to the uncompressed MoE, factorization is nearly lossless on $21$ of $22$ benchmarks; the exception is VSI-Bench, where frame-count experiments suggest impaired use of cross-frame evidence.
UNITE thus consolidates complementary embodied capabilities near single-model storage.
\end{abstract}

%% file: section/introduction.tex
\section{Introduction}
\label{sec:introduction}

\begin{figure}[!t]
    \centering
    \includegraphics[width=\columnwidth]{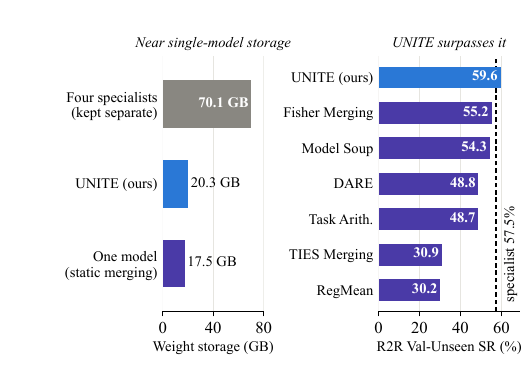}
    \caption{Storage versus capability retention for four Qwen3-VL-8B specialists. Keeping them separate (gray) costs $70.1$ GB; static merging (violet, Table~\ref{tab:storage-budget}) cuts this to $17.5$ GB, but every merging method then falls below the navigation specialist's $57.5\%$ R2R Val-Unseen success rate, reaching at best $55.2\%$ (Table~\ref{tab:r2r-main}). UNITE (blue) needs only $20.3$ GB, $1.16\times$ the single-specialist footprint, and surpasses the specialist at $59.6\%$ success rate.}
    \label{fig:tradeoff}
\end{figure}

\begin{figure*}[!t]
    \centering
    \includegraphics[width=0.9\textwidth]{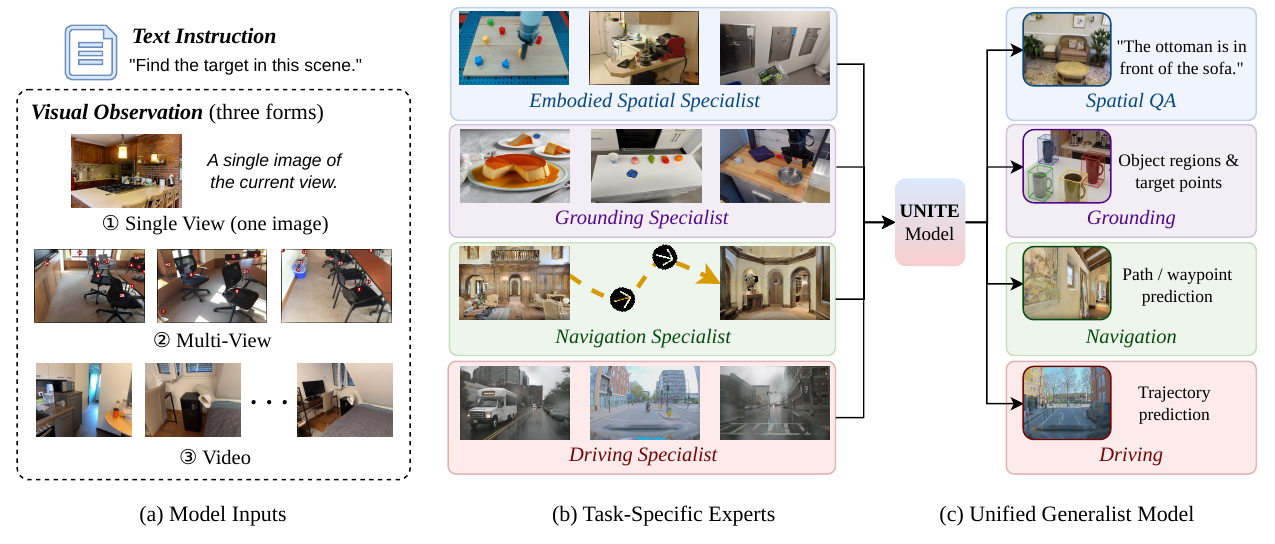}
    \caption{Consolidating embodied specialists with UNITE. Grounding, spatial reasoning, navigation, and driving understanding specialists from a common pretrained model contribute separate feed-forward experts to a shared-backbone MoE. Jointly fitted anchors and one-bit deviations compress the expert bank, bringing total model storage to $1.16\times$ that of one dense specialist; supervised router adaptation adjusts token-level expert selection.}
    \label{fig:overview}
\end{figure*}

\IEEEPARstart{E}{mbodied} intelligence aims to develop agents that can perceive and understand physical environments and use this understanding to guide their actions~\cite{geminirobotics2025}.
This requires complementary capabilities such as visual grounding, spatial reasoning, navigation, and driving understanding.
Recent advances in vision-language and vision-language-action foundation models have improved scene understanding, robot control, and action generation across diverse embodied tasks~\cite{black2025pi05,gong2026ace,gong2026acebrain,cen2025worldvla,cen2025rynnvla}.
These advances motivate bringing complementary embodied capabilities into a single model while retaining the expertise learned for individual tasks~\cite{community2026starvla}.

In practice, embodied tasks such as visual grounding~\cite{yu2016refcoco}, spatial reasoning~\cite{azuma2022scanqa}, navigation~\cite{anderson2018r2r}, and driving understanding~\cite{sima2024drivelm} differ in their observations, supervision signals, temporal horizons, and output formats~\cite{brohan2023rt2,octo2024,kim2024openvla,black2024pi0}.
The common practice is therefore to fine-tune a vision-language model (VLM) separately for each task, producing a collection of dense specialists that share an architecture and a pretrained initialization~\cite{yangModelMergingLLMs2024,yadavSurveyModelMoErging2024}.
Two common deployment choices expose a trade-off between storage and capability retention.
Storing the specialists side by side multiplies weight storage by the number of tasks: four Qwen3-VL-8B specialists occupy $70.1$ GB.
Statically merging them into one dense model retains the storage of a single specialist, but task-specific updates interfere~\cite{ilharco2023editingmodelstaskarithmetic,yadav2023tiesmergingresolvinginterferencemerging}.
This loss is evident in navigation: the navigation specialist reaches $57.5\%$ success rate on R2R Val-Unseen, whereas all six static merging methods fall below it, with the best reaching $55.2\%$ (Table~\ref{tab:r2r-main}).
Fig.~\ref{fig:tradeoff} summarizes this trade-off.
The challenge is to combine their capabilities near single-model storage (Fig.~\ref{fig:overview}).

Constructing a mixture of experts (MoE) from the specialists' feed-forward blocks avoids interference in these blocks by keeping each specialist's transformation as a separately routable expert~\cite{Sukhbaatar2024BranchTrainMiXME}, but retaining full-precision feed-forward weights for every specialist brings total model storage to $50.2$ GB in our setting.
Representing each expert as a shared component plus a compact expert-specific deviation removes this redundancy~\cite{Liu2024BitDeltaYF,Huang2025DeRSTE}.
Prior work fixes the shared component in advance, typically at the pretrained base or at a merged weight, and then compresses the deviation around it.
The key observation is that the shared anchor determines how well the expert-specific residuals can be represented at one bit per weight.
A row-scaled binary deviation represents only one magnitude per row, so the mean expert weight, which minimizes unquantized squared residuals, need not minimize binary reconstruction error.
Fitting the anchor jointly with the deviations accounts for this constraint and reduces reconstruction error by $20$--$22\%$ relative to a fixed mean anchor on the Qwen3-VL-8B expert bank, at identical storage.

We introduce \textbf{UNITE}, a framework that integrates this jointly optimized representation into a shared-backbone MoE for embodied VLM specialists.
Following the dense-to-MoE construction of Branch-Train-MiX~\cite{Sukhbaatar2024BranchTrainMiXME}, each decoder layer retains a complete feed-forward network (FFN) block from each specialist as a separately routable expert, while the remaining modules are merged into a shared backbone.
The factorization jointly fits a shared floating-point anchor, expert-specific row scales, and binary signs through closed-form alternating updates, each of which minimizes its conditional subproblem.
Fitting requires only source weights, with no calibration data or gradients, and each additional expert costs approximately one bit per weight.
Supervised router adaptation then adjusts token-level expert selection with the backbone and experts frozen, updating fewer than $0.6$M parameters ($<0.01\%$ of the $8.8$B active parameters).

We evaluate UNITE built from four Qwen3-VL-8B specialists~\cite{qwen3vl2025} for grounding, spatial reasoning, navigation, and driving understanding, against the source specialists and six static merging methods on $22$ understanding benchmarks and R2R Val-Unseen.
The complete four-expert UNITE model occupies $20.3$ GB, $1.16\times$ the footprint of one dense specialist and $59.6\%$ below the uncompressed MoE (Table~\ref{tab:storage-budget}).
At this footprint, UNITE matches static merging methods in mean retention of the best specialist's score across the $22$ benchmarks, and it leads all static merging methods on Scan2Cap ($102.1$ versus $86.4$), ScanQA ($105.6$ versus $101.8$), and all four navigation metrics.
UNITE also surpasses every source specialist on all five 3D scene understanding benchmarks and all four navigation metrics, improving Scan2Cap by $19.1$ points over the best specialist and reaching $59.6\%$ R2R success rate.
Stage-wise evaluation shows that factorization is nearly lossless relative to the uncompressed MoE on $21$ of $22$ benchmarks; on the exception, VSI-Bench, frame-count experiments suggest impaired use of cross-frame evidence.

\noindent\textbf{Our main contributions are threefold:}
(1) a shared-anchor one-bit factorization with closed-form updates that jointly optimize the anchor and expert-specific binary deviations using only source weights;
(2) UNITE, a shared-backbone MoE framework with compressed specialist FFNs and supervised router adaptation, achieving $1.16\times$ the storage of a single dense model; and
(3) a stage-wise evaluation across $22$ embodied understanding benchmarks and R2R navigation, which identifies the gains and losses from expert construction, factorization, and router adaptation.

%% file: section/related_work.tex
\section{Related Work}
\label{sec:related_work}

\textbf{Embodied foundation models and capability reuse.}
Vision-language-action (VLA) and world action models pursue broader embodied capabilities through multimodal learning and predictive modeling, as exemplified by $\pi_{0.5}$, WorldVLA, and DreamZero~\cite{black2025pi05,cen2025worldvla,ye2026worldactionmodelszeroshot}.
These models acquire breadth through large-scale joint training across tasks.
Parameter reuse provides a complementary route that composes already-trained skills.
RETAIN interpolates a pretrained VLA policy with its fine-tuned counterpart to retain general capabilities during task adaptation~\cite{yadav2026retain}, and MergeVLA combines masked parameter merging, cross-attention action experts, and task routing to merge multiple task-specialized VLA models for skill composition~\cite{fu2026mergevla}.
Both produce a single merged model and do not address the storage cost of holding many specialists.
UNITE targets the many-specialist setting: it consolidates existing dense VLM specialists derived from a common pretrained model, with an emphasis on compact expert storage and learned token-level selection.

\textbf{Static and dynamic parameter merging.}
Parameter merging methods differ in whether they produce a fixed model or compose expertise according to the input.
Model Soups averages fine-tuned model weights, while Task Arithmetic combines task vectors relative to a shared pretrained model~\cite{wortsman2022modelsoupsaveragingweights,ilharco2023editingmodelstaskarithmetic}.
Fisher Merging and RegMean use parameter importance and activation statistics, respectively~\cite{michaelmatenaMergingModelsFisherWeighted2022,xisenjinDatalessKnowledgeFusion2023}.
TIES-Merging resolves conflicts among update signs, and DARE sparsifies and rescales task updates~\cite{yadav2023tiesmergingresolvinginterferencemerging,yuLanguageModelsAre2023}.
All of these methods produce a single dense weight set, so capabilities whose updates conflict must share the same parameters; the resulting interference is evident in Section~\ref{sec:embodied_consolidation}, where every static method falls below the navigation specialist on R2R.
Dynamic variants reduce interference by conditioning on the input.
WEMoE constructs input-conditioned FFN weights from pretrained weights and task vectors, while Twin-Merging dynamically combines a shared merged model with low-rank task residuals~\cite{tang2024mergingmultitaskmodelsweightensembling,lu2024twinmergingdynamicintegrationmodular}.
These methods synthesize a new weight matrix per input, which requires the task vectors or residuals to be stored and combined at inference.
UNITE instead composes the outputs of complete feed-forward experts through token-level routing, so each expert is stored once in a compressed form and evaluated only when selected.

\textbf{Expert construction and routing.}
Constructing mixtures of experts (MoEs) from dense checkpoints reuses learned parameters to form separately routable expert branches.
Sparse Upcycling initializes experts by replicating feed-forward network (FFN) blocks from a single dense model~\cite{komatsuzakiSparseUpcyclingTraining2023}.
Branch-Train-MiX assembles FFN blocks from independently specialized models and averages the remaining parameters into a shared backbone~\cite{Sukhbaatar2024BranchTrainMiXME}.
MergeME addresses both homogeneous and heterogeneous expert models, using projection layers and sequence-level routing to connect heterogeneous model branches~\cite{Zhou2025MergeMEMM}.
These constructions retain every expert at full precision, so the expert bank grows linearly with the number of specialists; in our setting, the resulting $50.2$-GB uncompressed MoE is $2.9\times$ the size of one specialist.
UNITE adopts this construction and reduces its storage through jointly fitted anchors and row-scaled binary deviations.

Routing determines how the constructed expert branches are selected and used.
PHATGOOSE learns gates post hoc on each specialist's source data with the backbone and parameter-efficient modules frozen, then combines these gates for token- and module-level routing~\cite{Muqeeth2024LearningTR}.
SMILE selects experts using projection norms onto SVD-derived task-update subspaces without additional router training~\cite{tangZeroShotSparseMixture2025}.
MADE-IT uses feature-subspace alignment for training-free routing within an adaptive continual model-merging framework~\cite{qiu2026madeit}.
Training-free routers are fitted to the uncompressed specialist weights and do not see the reconstruction error introduced when the experts are compressed.
UNITE therefore uses singular directions of the centered source weights only to initialize the routers, then fine-tunes them on the compressed bank with supervision.

\textbf{Compact specialist and expert representations.}
Compact parameter representations reduce storage by separating shared weights from specialist-specific deviations.
BitDelta compresses fine-tuning updates around a fixed pretrained base using binary signs and matrix-level scales, while T-Switch represents task vectors through binary masks, signs, and scalar magnitudes~\cite{Liu2024BitDeltaYF,qi2024moreefficientmodelmerging}.
At the expert-bank level, DeRS represents upcycled experts using their common pretrained initialization and sparse or quantized deviations~\cite{Huang2025DeRSTE}.
Low-rank approaches offer another compact representation: D$^2$-MoE constructs a shared base through Fisher-weighted expert merging and factorizes expert-specific residuals~\cite{Gu2025DeltaDF}.
In all of these methods the shared component is fixed before the deviations are fitted, either at the pretrained base or at a merged weight.
When deviations are constrained to one bit per weight, this fixed choice determines the residuals the binary code must approximate, and the shared component cannot compensate for the representation's limited expressiveness.
UNITE removes this restriction by treating the anchor as an optimization variable: it jointly fits a floating-point anchor, expert-specific row scales, and binary deviation signs through alternating closed-form updates, allowing the shared anchor to adapt to the one-bit deviation budget while preserving a separate reconstruction for each expert.

%% file: section/method.tex
\section{Method}
\label{sec:methodology}

We consider $E$ dense vision-language specialists for complementary embodied capabilities, including grounding, spatial reasoning, navigation, and driving understanding.
They share an architecture and are fine-tuned from a common pretrained model.
UNITE consolidates them into a shared-backbone mixture of experts (MoE), reducing weight duplication while retaining conditional access to specialist transformations.

\begin{figure*}[!t]
    \centering
    \includegraphics[width=\linewidth]{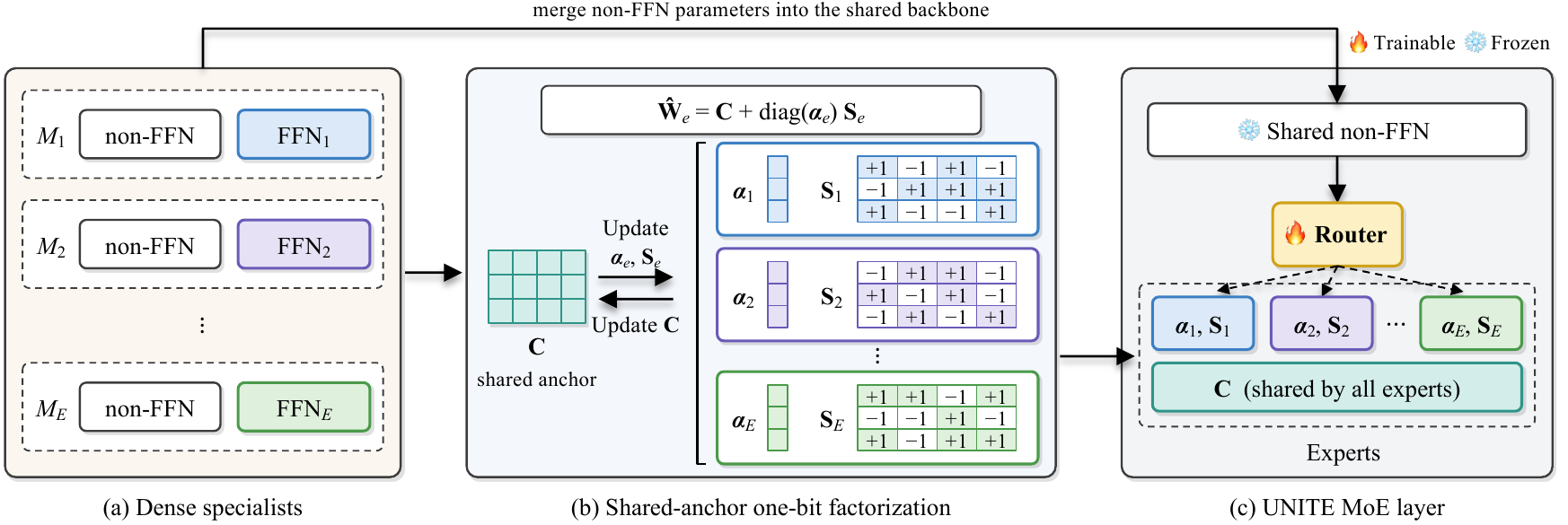}
    \caption{Overview of UNITE.
    (a) The non-FFN parameters of the dense specialists are merged into a shared backbone, while each specialist FFN is kept as an expert.
    (b) Alternating updates jointly fit the shared anchor $\mathbf{C}$, row scales $\boldsymbol{\alpha}_e$, and binary signs $\mathbf{S}_e$ for each layer and weight group.
    (c) In the resulting MoE layer, every expert is represented by its one-bit factors $(\boldsymbol{\alpha}_e, \mathbf{S}_e)$ on top of the shared anchor $\mathbf{C}$.
    The trainable/frozen markers refer to router adaptation, which trains only the routers while the backbone and the compressed experts stay frozen.}
    \label{fig:method}
\end{figure*}

The three stages serve distinct roles (Fig.~\ref{fig:method}).
First, expert construction retains complete specialist feed-forward blocks and merges the remaining parameters into the shared backbone.
Second, shared-anchor one-bit factorization compresses the expert weights using source weights alone.
Finally, supervised router adaptation learns token-level expert selection with the backbone and compressed experts frozen.

\subsection{Shared-Backbone Expert Construction}
\label{sec:moe_construction}

Each decoder layer retains one complete feed-forward network (FFN) block from each specialist.
An expert contains the gate, up, and down projections of its source specialist; the router selects these projections together.
Matching input and output dimensions allow the experts to process token representations within a shared decoder.

We form the shared backbone by merging corresponding non-FFN parameters.
We use norm-rescaled averaging for self-attention projection weights, whose Frobenius norms can shrink when source matrices differ in direction.
For corresponding source projections $\{\mathbf{P}_e\}_{e=1}^{E}$, let $\boldsymbol{\mu}=E^{-1}\sum_e\mathbf{P}_e$ and $\overline{n}=E^{-1}\sum_e\|\mathbf{P}_e\|_F$.
We rescale the mean to the average source norm, except near zero:
\begin{equation*}
    \mathbf{P}_{\mathrm{shared}}
    =
    \begin{cases}
        \boldsymbol{\mu}\,\overline{n}/\|\boldsymbol{\mu}\|_F,
        & \|\boldsymbol{\mu}\|_F\geq10^{-10},\\
        \boldsymbol{\mu}, & \text{otherwise}.
    \end{cases}
\end{equation*}
The remaining shared parameters (visual encoder, normalization layers, embeddings, and output head) are averaged.

Each MoE layer routes tokens independently to a subset of experts.
At layer $\ell$, expert $e$ maps an input token representation $\mathbf{x}\in\mathbb{R}^{d}$ to $F_{\ell,e}(\mathbf{x})$, where $d$ is the hidden dimension.
A linear router with weights $\mathbf{R}_{\ell}\in\mathbb{R}^{E\times d}$ assigns probabilities $\mathbf{p}_{\ell}(\mathbf{x})=\operatorname{softmax}(\mathbf{R}_{\ell}\mathbf{x})$.
The indices of the $k$ highest-probability experts form $\mathcal{T}_{\ell}(\mathbf{x})$.
We renormalize their probabilities and combine the selected expert outputs:

\begin{equation}
    \operatorname{MoE}_{\ell}(\mathbf{x})
    =
    \sum_{e\in\mathcal{T}_{\ell}(\mathbf{x})}
    \frac{p_{\ell,e}(\mathbf{x})}
    {\sum_{j\in\mathcal{T}_{\ell}(\mathbf{x})}p_{\ell,j}(\mathbf{x})}
    F_{\ell,e}(\mathbf{x}).
    \label{eq:router}
\end{equation}
Only the selected expert blocks are evaluated for each token.

\subsection{Joint Shared-Anchor One-Bit Factorization}
\label{sec:factorization}

We represent every expert matrix as a shared anchor plus an expert-specific binary deviation.
The anchor determines the residuals that the binary deviations must approximate, and with one nonnegative scale per output row, a binary deviation can represent only a single magnitude within that row.
An anchor fixed in advance therefore leaves reconstruction error that the deviations cannot absorb.
We instead treat the anchor as an optimization variable and formulate reconstruction as a joint optimization over the anchor, row scales, and binary signs.

We concatenate each expert's gate and up projections by rows into one weight group and treat its down projection as a second group.
For a fixed layer and weight group, expert $e$ has a weight matrix $\mathbf{W}_e\in\mathbb{R}^{O\times I}$ with $O$ output rows and $I$ input columns.
We approximate each expert matrix as
\begin{equation}
    \widehat{\mathbf{W}}_e
    =
    \mathbf{C}+\operatorname{diag}(\boldsymbol{\alpha}_e)\mathbf{S}_e,
    \label{eq:onebit_reconstruction}
\end{equation}
where $\mathbf{C}\in\mathbb{R}^{O\times I}$ is the anchor shared across experts within this layer and weight group.
The matrix $\mathbf{S}_e\in\{-1,+1\}^{O\times I}$ contains expert-specific binary signs.
The nonnegative vector $\boldsymbol{\alpha}_e\in\mathbb{R}_{\geq0}^{O}$ provides one scale per output row.
Anchors and row scales remain floating point; only the signs use one bit.

The joint objective minimizes the total squared reconstruction error across experts:
\begin{equation}
\begin{aligned}
    &\min_{\mathbf{C},\{\boldsymbol{\alpha}_e,\mathbf{S}_e\}_{e=1}^{E}}
    \sum_{e=1}^{E}
    \left\|
        \mathbf{W}_e-\mathbf{C}
        -\operatorname{diag}(\boldsymbol{\alpha}_e)\mathbf{S}_e
    \right\|_F^2,\\
    &\text{s.t.}\quad
    \boldsymbol{\alpha}_e\geq0,\qquad
    \mathbf{S}_e\in\{-1,+1\}^{O\times I}.
\end{aligned}
\label{eq:onebit_objective}
\end{equation}
We optimize every layer and weight group on its own.

\begin{figure}[!t]
  \centering
  \includegraphics[width=\linewidth]{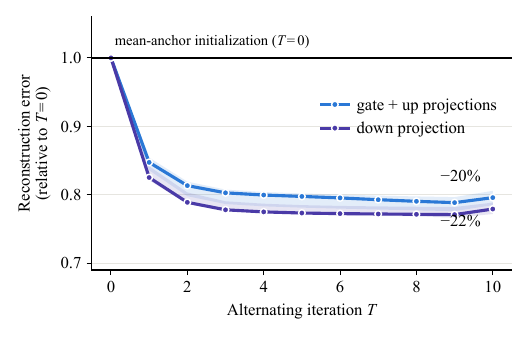}
  \caption{Joint anchor fitting versus a fixed anchor.
    Summed reconstruction error $\sum_e\|\mathbf{W}_e-\widehat{\mathbf{W}}_e\|_F^2$ as a function of the alternating iteration $T$, normalized per layer to its value at $T=0$ (fixed mean anchor with a single sign-and-scale fit).
    Lines are medians across the 36 decoder layers of the Qwen3-VL-8B expert bank; bands span the 10th to 90th percentile.
    The last point ($T=10$) is measured after rounding the anchor and row scales to BF16 and refitting the signs and scales, i.e., the archive used at inference; the rounding costs less than one percentage point.}
  \label{fig:recon-vs-T}
\end{figure}

We optimize Eq.~\eqref{eq:onebit_objective} by alternating between the expert-specific factors and the shared anchor for $T$ iterations.
We initialize the anchor with the mean expert weight, $\mathbf{C}^{(0)}=\frac{1}{E}\sum_{e=1}^{E}\mathbf{W}_e$.
At iteration $t\in\{0,\ldots,T-1\}$, fixing the anchor gives residuals $\boldsymbol{\Delta}_e^{(t)}=\mathbf{W}_e-\mathbf{C}^{(t)}$.
The conditionally optimal binary signs and row scales are
\begin{equation}
    \mathbf{S}_e^{(t)}
    =\operatorname{sign}\!\left(\boldsymbol{\Delta}_e^{(t)}\right),\quad
    \alpha_{e,o}^{(t)}
    =\frac{1}{I}\sum_{i=1}^{I}\left|\Delta_{e,o,i}^{(t)}\right|,
    \label{eq:sign_scale_update}
\end{equation}
where $o$ and $i$ index rows and columns, and $\operatorname{sign}(\cdot)$ is applied entrywise with $\operatorname{sign}(x)=1$ for $x\geq0$ and $-1$ for $x<0$.
The signs follow the residual entries; each row scale is the row's mean absolute residual.
With these factors fixed, we subtract the reconstructed deviations from each source matrix and average the remaining weights to update the anchor:
\begin{equation}
    \mathbf{C}^{(t+1)}
    =
    \frac{1}{E}\sum_{e=1}^{E}
    \left[
        \mathbf{W}_e
        -\operatorname{diag}\!\left(\boldsymbol{\alpha}_e^{(t)}\right)
        \mathbf{S}_e^{(t)}
    \right].
    \label{eq:anchor_update}
\end{equation}
Each closed-form update minimizes its conditional subproblem, so the objective cannot increase in exact arithmetic.
The objective values therefore form a non-increasing sequence bounded below by zero and hence converge.
Each iteration costs $O(EOI)$ elementwise operations and requires no data.
Iteration $t=0$ corresponds to a fixed anchor: a single sign-and-scale fit of the residuals around the mean expert weight.
The mean minimizes the total squared deviation $\sum_e\|\mathbf{W}_e-\mathbf{C}\|_F^2$ and is therefore a natural fixed choice, but it is not optimal once the deviations are restricted to row-scaled signs.
Fig.~\ref{fig:recon-vs-T} reports the reduction on the Qwen3-VL-8B expert bank: after ten iterations and the final BF16 rounding, the reconstruction error is $20\%$ (gate-up) and $22\%$ (down projection) lower than with the fixed mean anchor (median over the 36 decoder layers; the 10th--90th percentile range spans $19$--$23\%$), and over $90\%$ of the reduction is obtained within the first three iterations.

After $T$ alternating updates in 32-bit floating point (FP32), we round the final anchor $\mathbf{C}^{(T)}$ to bfloat16 (BF16).
We recompute signs and row scales relative to this stored anchor so that the deviations are fitted to the anchor used during inference.
The recomputed row scales are then stored in BF16.
Algorithm~\ref{alg:joint_factorization} summarizes the procedure for one layer and weight group.
\begin{algorithm}[!t]
\caption{Alternating optimization for shared-anchor one-bit factorization}
\label{alg:joint_factorization}
\small
\algrenewcommand{\algorithmicrequire}{\textbf{Input:}}
\algrenewcommand{\algorithmicensure}{\textbf{Output:}}
\begin{algorithmic}[1]
\Require Expert matrices $\{\mathbf{W}_e\}_{e=1}^{E}$, $\mathbf{W}_e\in\mathbb{R}^{O\times I}$; iteration count $T$.
\Ensure Stored anchor $\mathbf{C}_{\mathrm{store}}$, signs $\{\mathbf{S}_e\}_{e=1}^{E}$, and row scales $\{\boldsymbol{\alpha}_{e,\mathrm{store}}\}_{e=1}^{E}$.
\Statex Use FP32 for alternating updates and $\operatorname{sign}(0)=+1$.
\State $\mathbf{C}^{(0)}\gets \frac{1}{E}\sum_{e=1}^{E}\mathbf{W}_e$
\For{$t=0,\ldots,T-1$}
    \For{$e=1,\ldots,E$}
        \State $\boldsymbol{\Delta}_e^{(t)}\gets\mathbf{W}_e-\mathbf{C}^{(t)}$
        \State $\mathbf{S}_e^{(t)}\gets\operatorname{sign}(\boldsymbol{\Delta}_e^{(t)})$
        \State $\alpha_{e,o}^{(t)}\gets\frac{1}{I}\sum_{i=1}^{I}|\Delta_{e,o,i}^{(t)}|$, $o=1,\ldots,O$
    \EndFor
    \State $\mathbf{C}^{(t+1)}\gets\frac{1}{E}\sum_{e=1}^{E}\bigl[\mathbf{W}_e-\operatorname{diag}(\boldsymbol{\alpha}_e^{(t)})\mathbf{S}_e^{(t)}\bigr]$
\EndFor
\State $\mathbf{C}_{\mathrm{store}}\gets\operatorname{BF16}(\mathbf{C}^{(T)})$
\For{$e=1,\ldots,E$}
    \State $\boldsymbol{\Delta}_e\gets\mathbf{W}_e-\mathbf{C}_{\mathrm{store}}$
    \State $\mathbf{S}_e\gets\operatorname{sign}(\boldsymbol{\Delta}_e)$
    \State $\alpha_{e,o}\gets\frac{1}{I}\sum_{i=1}^{I}|\Delta_{e,o,i}|$, $o=1,\ldots,O$
    \State $\boldsymbol{\alpha}_{e,\mathrm{store}}\gets\operatorname{BF16}(\boldsymbol{\alpha}_e)$
\EndFor
\State \Return $\mathbf{C}_{\mathrm{store}},\{\mathbf{S}_e,\boldsymbol{\alpha}_{e,\mathrm{store}}\}_{e=1}^{E}$
\end{algorithmic}
\end{algorithm}
Storing $E$ expert matrices requires $16OI+EOI+16EO$ bits before packing alignment.
The three terms account for the shared anchor, expert-specific binary signs, and row scales, respectively.
Dividing by the $EOI$ logical expert weights gives $1+16/E+16/I$ bits per weight, including the shared anchor and scales.
The $16/E$ term reflects the anchor cost amortized across experts.
For $E=4$, this is approximately five bits per expert weight, compared with $16$ bits for an uncompressed BF16 expert bank; the shared backbone and routers are counted separately.
Because the anchor is shared, the marginal cost of adding one more expert is only $1+16/I$ bits per weight, so the representation becomes more economical as more specialists are consolidated.

\subsection{Router Initialization and Adaptation}
\label{sec:router_sft}

Routing determines how tokens use the constructed and compressed expert bank.
We initialize each router from differences among the original specialist weights, without calibration data.
At layer $\ell$, let $\mathbf{W}_{\ell,e}^{\mathrm{gate}}$ and $\mathbf{W}_{\ell,e}^{\mathrm{up}}$ denote expert $e$'s original gate and up matrices.
Let $\overline{\mathbf{W}}_{\ell}^{\mathrm{gate}}$ and $\overline{\mathbf{W}}_{\ell}^{\mathrm{up}}$ be their respective means across specialists.
We subtract these means and stack the centered matrices by rows:
\begin{equation*}
    \mathbf{X}_{\ell,e}
    =
    \begin{bmatrix}
        \mathbf{W}_{\ell,e}^{\mathrm{gate}}-\overline{\mathbf{W}}_{\ell}^{\mathrm{gate}}\\
        \mathbf{W}_{\ell,e}^{\mathrm{up}}-\overline{\mathbf{W}}_{\ell}^{\mathrm{up}}
    \end{bmatrix}.
\end{equation*}
We approximate the leading right singular vector $\mathbf{v}_{\ell,e,1}$ using randomized singular value decomposition (SVD).
This vector initializes the corresponding router row:
\begin{equation}
    \mathbf{R}_{\ell}[e,:]
    =
    \mathbf{v}_{\ell,e,1}^{\mathsf T}.
    \label{eq:svd_router_init}
\end{equation}
It captures a dominant input direction along which the specialist's gate and up projections differ from their cross-specialist means.
We compute the SVD in FP32 and retain the sign returned by the numerical routine.
The router row is initialized to zero if $\max_{o,i}|X_{\ell,e,o,i}|<10^{-8}$.

Supervised router adaptation adjusts expert selection to the compressed model.
The shared backbone, anchors, row scales, and binary signs remain frozen.
Training combines causal language-modeling loss on target answer tokens with an auxiliary load-balancing loss, without explicit expert-assignment labels.
During router adaptation, we select multiple experts per token so gradients from the language-modeling loss can pass through the selected mixture weights.

%% file: section/experiments.tex
\section{Experiments}
\label{sec:experiments}

Our experiments answer four questions.
(i) How much storage does UNITE save relative to the alternatives of keeping separate specialists or an uncompressed MoE (Section~\ref{sec:storage_memory})?
(ii) At that storage, does UNITE retain the capabilities that static merging loses, and does consolidation add capability beyond the source specialists (Section~\ref{sec:embodied_consolidation})?
(iii) At which stage do gains and losses arise, and how does router adaptation interact with one-bit compression (Sections~\ref{sec:controlled_ablations} and~\ref{sec:routing_topk})?
(iv) Does the factorization transfer to a second model family without router training (Section~\ref{sec:language_extension})?

\subsection{Experimental Setup}
\label{sec:experimental_setup}

\noindent\textbf{Source models.}
We fine-tune four Qwen3-VL-8B specialists for grounding, embodied spatial reasoning, navigation, and driving understanding~\cite{qwen3vl2025}.
UNITE and the static merging baselines use these same four source checkpoints.
The language-model extension uses Llama-3.1-8B specialists from MergeBench~\cite{he2025mergebench} without router adaptation (Section~\ref{sec:language_extension}).

\noindent\textbf{Tasks and metrics.}
The embodied evaluation covers $22$ understanding benchmarks and navigation on R2R Val-Unseen~\cite{anderson2018r2r}.
(1) \textit{The grounding benchmarks} are RoboSpatial~\cite{song2025robospatial}, RefSpatial~\cite{zhou2025roborefer}, PointArena~\cite{cheng2025pointarena}, RoboAfford~\cite{roboafford2025}, and the ShareRobot trajectory task~\cite{ji2025robobrain}.
(2) \textit{The embodied spatial benchmarks} are VSI-Bench (VSI)~\cite{yang2025vsibench}, MMSI-Bench (MMSI)~\cite{yang2025mmsibench}, MindCube~\cite{wang2025mindcube}, EmbSpatial-Bench~\cite{du2024embspatial}, ERQA~\cite{geminirobotics2025}, and SAT~\cite{ray2025sat}, together with five 3D scene understanding benchmarks: ScanQA~\cite{azuma2022scanqa}, SQA3D~\cite{ma2023sqa3d}, Scan2Cap~\cite{chen2021scan2cap}, ScanRefer~\cite{chen2019scanrefer}, and Multi3DRefer (Multi3DRef)~\cite{zhang2023multi3drefer}.
(3) \textit{The real-world and driving benchmarks} are MME-RealWorld~\cite{zhang2025mmerealworld}, MAPLM~\cite{cao2024maplm}, DriveAction~\cite{driveaction2025}, NuScenes-QA (NuScenesQA)~\cite{qian2024nuscenesqa}, NuPlanQA~\cite{park2025nuplanqa}, and LingoQA~\cite{marcu2024lingoqa}.

We use each benchmark's reported metric, including overall accuracy for MMSI and DriveAction; MindCube uses the tiny configuration.
For 3D scene understanding, we report CIDEr for ScanQA, EM-1 for SQA3D, and CIDEr@0.25 for Scan2Cap.
ScanRefer and Multi3DRef use Acc@0.25 and F1@0.25, respectively.
For R2R, we report navigation error (NE), oracle success (OS), success rate (SR), and success weighted by path length (SPL).
NE is in meters; OS, SR, and SPL are percentages.
ShareRobot reports mean normalized discrete Fr\'echet distance (nDFD) for trajectory prediction.
Lower is better for NE and nDFD; higher is better otherwise.
The supplement details benchmark variants and protocols.

\noindent\textbf{Baselines.}
We compare with Model Soup~\cite{wortsman2022modelsoupsaveragingweights}, Task Arithmetic~\cite{ilharco2023editingmodelstaskarithmetic}, TIES Merging~\cite{yadav2023tiesmergingresolvinginterferencemerging}, DARE~\cite{yuLanguageModelsAre2023}, Fisher Merging~\cite{michaelmatenaMergingModelsFisherWeighted2022}, and RegMean~\cite{xisenjinDatalessKnowledgeFusion2023}.
All six methods merge the same source checkpoints into one dense model of $17.5$ GB.
For context, we also evaluate general-purpose and embodied reference models on the same benchmark data with the same metrics.
They differ from UNITE in architecture, scale, and training data. Tables~\ref{tab:unite-leading-benchmarks} and~\ref{tab:r2r-main} compare their task-level scores with UNITE, while Tables~\ref{tab:setup-models} and~\ref{tab:setup-r2r} compare methods built from the same source checkpoints.
The supplement details baseline implementations and calibration settings.

\noindent\textbf{Training and evaluation.}
We apply ten alternating factorization updates and store anchors and row scales in BF16.
Router adaptation updates 589,824 parameters with the backbone and expert bank frozen.
We use top-$2$ routing during adaptation and top-$1$ routing for the main evaluations.
Top-$2$ training allows gradients through the selected mixture weights, whereas the normalized top-$1$ coefficient in Eq.~\eqref{eq:router} is constant.
Uncompressed and compressed router-tuned models share the same training budget and supervision mixture; full optimization settings are provided in the supplement.

\noindent\textbf{Router supervision.}
We use $55{,}757$ samples from $18$ sources covering navigation, grounding, spatial reasoning, and related vision-language tasks.
The mixture includes training data associated with several evaluated tasks (R2R, RefSpatial, ShareRobot affordance and trajectory, VSI-590K, MMSI, MindCube, and SAT).
Router adaptation updates only the routing parameters, and its largest gain occurs on NuScenesQA, which is absent from the mixture (Section~\ref{sec:controlled_ablations}).
The supplement details the full mixture.

\subsection{Weight Storage and GPU Memory}
\label{sec:storage_memory}

Table~\ref{tab:storage-budget} places UNITE between the two deployment options discussed in Section~\ref{sec:introduction}.
Four standalone specialists require $70.1$ GB.
Sharing the non-FFN backbone alone reduces this to $50.2$ GB for the uncompressed MoE, still $2.9\times$ one dense specialist, because the expert bank remains at full precision.
Shared-anchor one-bit factorization reduces expert-bank storage from $43.5$ to $13.6$ GB; adding $6.7$ GB for shared modules and routers gives $20.3$ GB of total weight storage.
This is a $59.6\%$ reduction relative to the uncompressed MoE and a $71.0\%$ reduction relative to the standalone collection.
It is only $1.16\times$ the $17.5$-GB footprint of a single dense specialist or any static merging baseline, while retaining four routable experts.
\input{table/storage_memory_table}

Peak allocated GPU memory during evaluation decreases from $50.6$ GB for the uncompressed MoE to $21.0$ GB for UNITE, confirming that the storage reduction carries over to deployment memory.
We next examine the embodied capabilities retained within this footprint.

\subsection{Consolidation Performance}
\label{sec:embodied_consolidation}

Tables~\ref{tab:unite-leading-benchmarks} and~\ref{tab:r2r-main} compare UNITE with reference models on understanding and navigation. Tables~\ref{tab:setup-models} and~\ref{tab:setup-r2r} compare static merging with UNITE and its source experts on the same tasks.
In mean retention of best-specialist performance (defined under \emph{Remaining deficits} below), UNITE matches static merging methods; it leads on all four navigation metrics and trails the strongest static method by over one point only on VSI, MME-RealWorld, RoboSpatial, and RefSpatial.
\input{table/moe-qwen3vl8b-main.tex}

\begin{figure*}[!t]
  \centering
  \includegraphics[width=\linewidth]{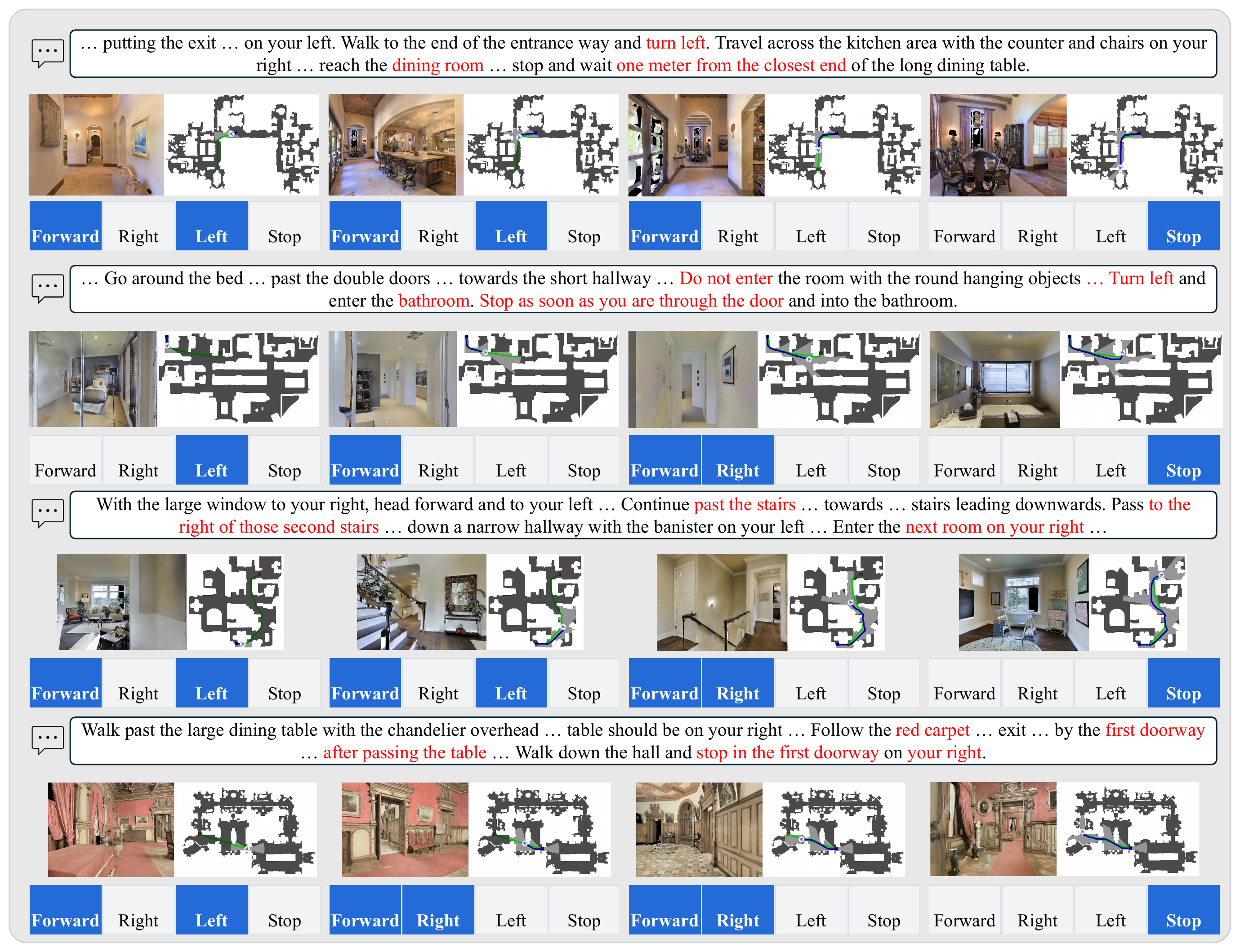}
  \caption{Navigation in R2R Val-Unseen environments.
    Each row shows an instruction excerpt and four snapshots pairing egocentric observations with trajectory maps and action annotations.
    Green and blue indicate the ground-truth and UNITE trajectories, respectively.
  Tables~\ref{tab:r2r-main} and~\ref{tab:setup-r2r} report navigation results.}
  \label{fig:r2r-unseen}
\end{figure*}

\noindent\textbf{UNITE leads static merging most on 3D scene understanding.}
On Scan2Cap, UNITE reaches $102.1$ against the best static score of $86.4$, and on ScanQA $105.6$ against $101.8$.

\noindent\textbf{Driving question answering depends on router adaptation.}
The driving specialist reaches $58.8$ on NuScenesQA (Table~\ref{tab:setup-models}); after static merging, Model Soup retains $40.1$, Fisher Merging $19.2$, and the best static method, Task Arithmetic, $48.6$.
Keeping the driving FFN as a routable expert is not sufficient on its own: before router adaptation, UNITE scores $27.7$; adapting the router raises this to $48.7$, matching the best static method.
TIES Merging and RegMean degrade broadly on this specialist collection under their standard settings, which is consistent with four task updates that conflict too strongly for sign resolution or activation-matched averaging to reconcile within a single weight set.

\noindent\textbf{Consolidation is additive rather than merely lossless.}
On several tasks, UNITE exceeds every source specialist.
The best specialist scores on ScanQA, SQA3D, and Scan2Cap are $97.7$, $61.7$, and $83.0$; UNITE reaches $105.6$, $62.8$, and $102.1$.
PointArena rises from $68.3$ for the grounding specialist to $71.3$, RoboAfford from $77.7$ to $80.3$, and MAPLM from $88.4$ to $93.8$.
These margins are substantially larger than those of the best static method ($+19.1$ versus $+3.4$ on Scan2Cap and $+7.9$ versus $+4.1$ on ScanQA), consistent with the additional contribution of token-level routing, which allows tokens within a single forward pass to draw on different experts.

\input{table/r2r-main.tex}

\noindent\textbf{Navigation benefits from both expert retention and router adaptation.}
On R2R Val-Unseen (Table~\ref{tab:setup-r2r}, Fig.~\ref{fig:r2r-unseen}), UNITE leads all static baselines on every metric, with $59.6\%$ SR and $53.4\%$ SPL against the best static scores of $55.2\%$ and $50.5\%$, and a navigation error of $4.5$ m against $5.0$ m.
UNITE also surpasses the navigation specialist on all four metrics ($4.7$ m NE, $63.7\%$ OS, $57.5\%$ SR, $51.2\%$ SPL), whereas every static method remains below it.
The router mixture contains R2R training data, and before adaptation the one-bit MoE ($57.0\%$ SR) is just below the specialist, so the margin over the specialist comes from router adaptation.
UNITE also exceeds all navigation reference models in SR and SPL (Table~\ref{tab:r2r-main}).

\noindent\textbf{Remaining deficits.}
The largest gap is on the video benchmark VSI, where UNITE scores $46.7$ against $63.6$ for Model Soup.
On the remaining nine benchmarks on which UNITE does not lead, the strongest static method is higher by between $0.3$ and $4.9$ points; the gap exceeds one point only on MME-RealWorld ($4.9$), RoboSpatial ($4.3$), and RefSpatial ($3.3$), and is below one point on MMSI, DriveAction, MindCube, Multi3DRef, LingoQA, and ScanRefer.
The RoboSpatial and MME-RealWorld deficits are already present in the uncompressed MoE (Section~\ref{sec:controlled_ablations}) and therefore do not originate in compression.
In mean retention (the score divided by the best specialist's score, with the ratio inverted for nDFD, averaged over the $22$ benchmarks), UNITE ($98.8\%$) and the strongest static methods ($98.6$--$98.9\%$) are close.
The difference lies in \emph{which} capabilities are retained: UNITE's main loss is concentrated in one video benchmark, whereas its advantages are concentrated in 3D captioning and question answering and in navigation.
The VSI decline arises during factorization and persists after router adaptation.

\subsection{Stage-wise Analysis}
\label{sec:controlled_ablations}

\begin{figure}[t]
  \centering
  \includegraphics[width=\linewidth]{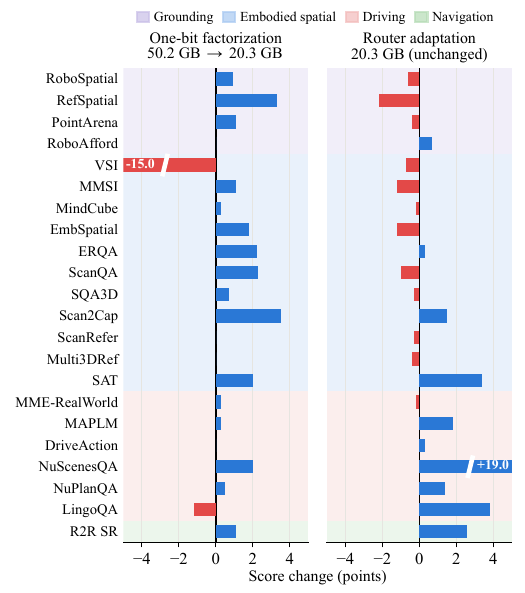}
  \caption{Score change contributed by each consolidation stage (top-$1$ routing throughout).
    Left: one-bit factorization of the expert bank, reducing total model weight storage from $50.2$ to $20.3$ GB. Right: router adaptation at constant storage.
    Bars beyond $\pm5$ points are truncated and labeled with their value.
    ShareRobot (Traj., on a different scale) improves from $0.34$ to $0.26$ and $0.25$.
    Full numbers are in Tables~\ref{tab:setup-models} and~\ref{tab:setup-r2r}.}
  \label{fig:stagewise}
\end{figure}

We trace when capability gains and losses arise through the three consolidation stages: the uncompressed MoE, the shared-anchor one-bit MoE, and the router-adapted one-bit MoE.
Tables~\ref{tab:setup-models} and~\ref{tab:setup-r2r} compare these stages with source experts and static merging on understanding and navigation. Figure~\ref{fig:stagewise} shows the change contributed by each stage across benchmarks.
All three stages use top-$1$ routing, so the number of activated expert branches is fixed.
\input{table/setup_models_table}
\input{table/stage_r2r}

\noindent\textbf{Source specialists are complementary.}
The navigation and embodied spatial specialists record $57.5\%$ and $56.8\%$ R2R SR, whereas the grounding specialist reaches $28.2\%$ and the driving specialist $0.1\%$.
Conversely, the driving specialist leads on NuScenesQA ($58.8$), LingoQA ($65.8$), and SAT ($92.0$), where the other specialists are far behind.
A consolidated model must preserve capabilities held by a single source.

\noindent\textbf{Expert construction already composes capabilities.}
Before any compression or router training, the uncompressed MoE scores $104.3$ on ScanQA and $97.1$ on Scan2Cap, exceeding every source specialist.
Its R2R SR ($55.9\%$), NuScenesQA ($27.7$), RoboSpatial ($58.0$), and MME-RealWorld ($65.6$) scores are below the corresponding best specialists ($57.5\%$, $58.8$, $68.0$, and $71.2$).
Router adaptation later recovers R2R and most of NuScenesQA, whereas the RoboSpatial and MME-RealWorld gaps persist; they therefore arise at expert construction rather than from compression.

\noindent\textbf{One-bit factorization is nearly lossless on 21 of 22 benchmarks.}
Factorizing the expert bank reduces total model weight storage from $50.2$ to $20.3$ GB, lowers no score other than VSI by more than $1.2$ points, and improves $16$ of the $22$: ScanQA increases from $104.3$ to $106.6$, Scan2Cap from $97.1$ to $100.6$, RefSpatial from $56.1$ to $59.4$, and ShareRobot (Traj.) from $0.34$ to $0.26$.
R2R SR also improves, from $55.9\%$ to $57.0\%$.
At matched storage, the one-bit bank also outperforms data-free INT4 quantization of the same expert weights on $17$ of $22$ understanding benchmarks and all navigation metrics (Detailed Results supplement).
Both representations occupy $20.3$ GB; INT4 uses round-to-nearest quantization with group size $16$.
The sole exception is VSI, which falls from $62.4$ to $47.4$.
We next examine how this deficit scales with frame count.

\begin{figure}[t]
  \centering
  \includegraphics[width=\linewidth]{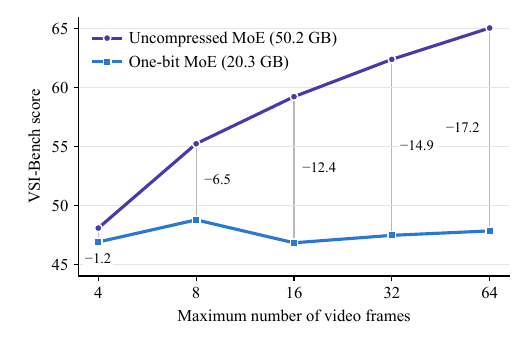}
  \caption{VSI-Bench score as a function of the maximum number of sampled video frames for the uncompressed and the one-bit expert bank (same backbone, top-$1$ routing, no router adaptation; all $5{,}130$ questions).
    The two models are within $1.2$ points at four frames.
    The uncompressed MoE gains $17.0$ points from $4$ to $64$ frames, whereas the one-bit MoE stays within a two-point band, indicating limited benefit from additional frames after factorization.
    The main tables use $32$ frames.}
  \label{fig:vsi-frames}
\end{figure}

\begin{figure*}[!t]
  \centering
  \includegraphics[width=0.85\linewidth]{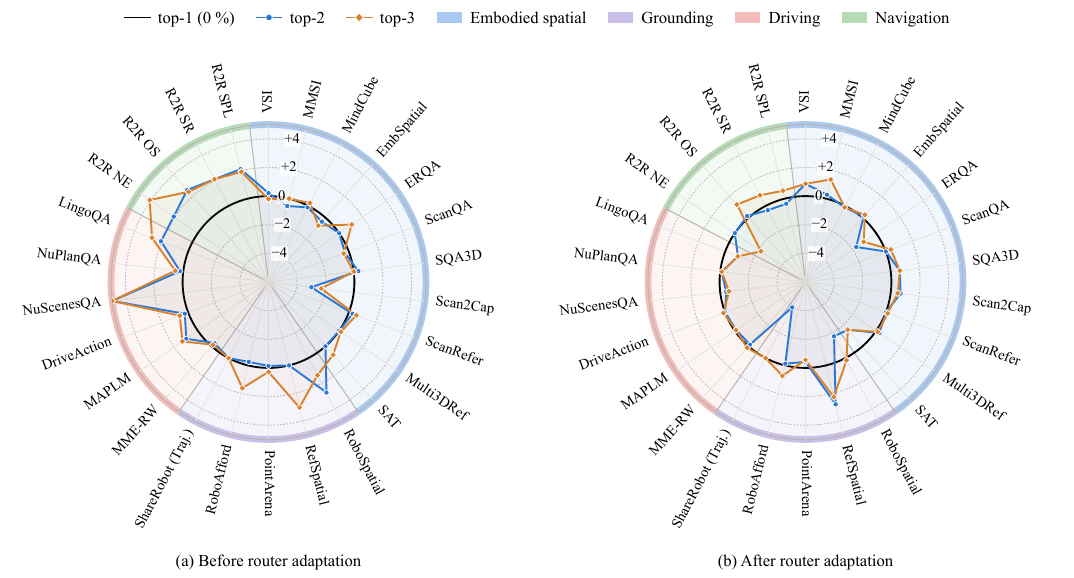}
  \caption{Top-$k$ routing before and after router adaptation.
    The radius is the direction-adjusted percentage change of top-$k$ relative to top-$1$ at the same stage, so the bold circle marks top-$1$ and points outside it indicate improvement for every metric, including ShareRobot (Traj.) and R2R NE.
    The radial axis spans $-6\%$ to $+5\%$; NuScenesQA before router adaptation ($+6.1\%$ and $+11.8\%$) exceeds this range and is clipped at the outer edge.}
  \label{fig:routing-curves}
\end{figure*}

\noindent\textbf{The one-bit MoE gains little from additional frames on VSI.}
To characterize this deficit, we evaluate both expert banks on VSI while varying the number of sampled frames from $4$ to $64$ (Fig.~\ref{fig:vsi-frames}).
At four frames the uncompressed and the one-bit MoE score $48.1$ and $46.9$, a gap of $1.2$ points in line with the image benchmarks.
With more frames the uncompressed model improves monotonically to $55.2$, $59.2$, $62.4$, and $65.0$, while the one-bit model stays between $46.8$ and $48.8$ with no trend; the gap therefore widens to $17.2$ points at $64$ frames.
The sub-task breakdown is consistent with this reading.
On object counting and object-size estimation, scores change by at most $5$ points across frame counts for either model, and the two models differ by at most about six points at any frame count.
By contrast, the uncompressed model improves from $49.4$ to $80.3$ on relative direction, from $51.0$ to $77.8$ on appearance order, from $28.3$ to $49.1$ on absolute distance, and from $45.6$ to $62.2$ on room size, whereas the one-bit model varies by less than six points on each of them.
These results suggest that one-bit factorization limits the model's use of additional cross-frame evidence.
At the same total model weight storage, the INT4 bank retains VSI at $61.9$ while trailing on most other benchmarks, showing that the two representations have different error profiles.
Router adaptation does not recover the VSI deficit (Table~\ref{tab:compression-router}); testing larger bit budgets in selected experts or layers is left to future work.

\noindent\textbf{Router adaptation refines selection.}
With the backbone and compressed experts frozen, router adaptation raises SAT from $87.3$ to $90.7$, LingoQA from $59.2$ to $63.0$, MAPLM from $92.0$ to $93.8$, NuScenesQA from $29.7$ to $48.7$, and R2R SR from $57.0\%$ to $59.6\%$.
It lowers $12$ of the $22$ benchmarks, all but RefSpatial ($59.4$ to $57.2$) by at most $1.2$ points.

\noindent\textbf{Compression and router adaptation interact.}
To separate the effect of compression from that of adaptation, we also train the routers of the uncompressed MoE under the same supervision mixture and training budget (Table~\ref{tab:compression-router}).
\input{table/compression_router}

After adaptation, the $20.3$-GB compressed MoE outperforms the $50.2$-GB uncompressed one on Scan2Cap ($102.1$ versus $94.1$), NuScenesQA ($48.7$ versus $24.4$), and R2R SR and SPL, while the uncompressed model leads on ScanQA ($111.1$ versus $105.6$), SQA3D, and VSI, and marginally on ScanRefer and Multi3DRef.
Two observations follow.
First, one-bit compression does not systematically limit what router adaptation can achieve: on four of the nine reported metrics, the compressed model is the better one after adaptation.
Second, pre-adaptation differences do not reliably predict post-adaptation ones: the compressed bank leads on ScanQA and SQA3D before adaptation and trails after it, while its NuScenesQA lead grows from $2.0$ to $24.3$ points.

\subsection{\texorpdfstring{Routing Top-$k$}{Routing Top-k}}
\label{sec:routing_topk}

We evaluate whether activating additional experts, $k\in\{1,2,3\}$, improves the consolidated model before and after router adaptation (Fig.~\ref{fig:routing-curves}; absolute scores in the Detailed Results supplement).
Activating $k$ experts raises the active parameter count from $8.8$B to $14.2$B and $19.6$B.

Before router adaptation, the effect of larger $k$ is task-dependent.
Navigation, grounding, and driving question answering benefit: R2R SR rises from $57.0\%$ to $58.2\%$, RefSpatial from $59.4$ to $61.2$, and NuScenesQA from $29.7$ to $33.2$ at $k=3$.
With weight-derived routers, no single expert yet serves all tasks well; more experts partially compensate.

After router adaptation, this sensitivity is largely eliminated.
Router adaptation alone raises top-$1$ NuScenesQA from $29.7$ to $48.7$, far more than activating three experts before adaptation, and NuScenesQA then stays within $0.3$ points across $k$ ($48.7$, $48.5$, $48.4$); all other benchmarks change by at most approximately two points.
R2R SR moves from $59.6\%$ to $60.1\%$ at $k=3$.
The adapted top-$1$ configuration, which activates the same $8.8$B parameters as a dense specialist, therefore captures nearly all of the benefit, and additional expert activation is not required for the reported results.

\subsection{Extension to Language Models}
\label{sec:language_extension}

We also evaluate the factorization on five Llama-3.1-8B specialists from MergeBench~\cite{he2025mergebench}, covering instruction following, mathematics, multilingual understanding, coding, and safety (Table~\ref{tab:llm-extension}).
Both UNITE stages use weight-derived SVD initialization and top-$1$ routing without router adaptation.
Math, Multilingual, Coding, and Safety report within-group averages; Instruction reports IFEval prompt-level loose accuracy.

\begin{table}[!htbp]
  \centering
  \caption{Llama-3.1-8B extension on MergeBench. Source names denote specialization. Both UNITE stages use top-$1$ without router adaptation; highlights rank merging methods and UNITE. Higher is better in all columns.}
  \label{tab:llm-extension}
  \small
  \setlength{\tabcolsep}{2pt}
  \renewcommand{\arraystretch}{1.10}
  \begin{adjustbox}{max width=\linewidth,center}
    \begin{tabular}{@{}lccccc@{}}
      \toprule
      \textbf{Model}            & \textbf{Math}    & \textbf{Multilingual} & \textbf{Instruction} & \textbf{Coding}  & \textbf{Safety} \\
      \midrule
      \multicolumn{6}{l}{\textit{Base model and source specialists}} \\
      Pretrained base           & 28.9             & 53.9                  & 10.5                 & 36.4             & 36.1 \\
      Instruction expert        & 34.9             & 55.3                  & 54.2                 & 47.4             & 43.8 \\
      Math expert               & 62.9             & 48.1                  & 14.6                 & 31.4             & 36.6 \\
      Multilingual expert       & 32.4             & 55.9                  & 4.1                  & 29.2             & 41.4 \\
      Coding expert             & 30.5             & 53.1                  & 27.4                 & 56.2             & 42.6 \\
      Safety expert             & 27.6             & 48.7                  & 7.8                  & 42.6             & 77.6 \\
      \midrule
      \multicolumn{6}{l}{\textit{Static merging methods}} \\
      Model Soup                & 47.8             & \underline{54.4}      & 16.8                 & \textbf{53.4}    & 56.4 \\
      Task Arithmetic           & \underline{55.0} & 52.1                  & \textbf{35.5}        & 50.3             & 57.6 \\
      TIES Merging              & \textbf{55.9}    & 54.1                  & 19.0                 & 50.0             & \textbf{63.2} \\
      DARE                      & 54.1             & 51.5                  & \underline{33.6}     & 49.4             & 58.4 \\
      Fisher Merging            & 45.5             & \underline{54.4}      & 17.4                 & \underline{51.1} & 55.5 \\
      RegMean                   & 49.0             & \textbf{55.1}         & 25.1                 & 49.7             & \underline{61.7} \\
      \midrule
      \multicolumn{6}{l}{\textit{UNITE stages}} \\
      Uncompressed MoE          & 49.4             & \textbf{55.1}         & 21.6                 & 46.9             & 53.1 \\
      Shared-anchor one-bit MoE & 49.9             & \textbf{55.1}         & 22.3                 & 48.2             & 53.3 \\
      \bottomrule
    \end{tabular}
  \end{adjustbox}
\end{table}

This experiment tests the factorization alone on a second model family with five experts, where the amortized anchor cost is lower ($1+16/5$ bits per weight before scales).
Keeping the routers unchanged isolates the effect of factorization relative to the uncompressed MoE.
Factorization matches or improves the uncompressed MoE in every group: Math, Instruction, and Coding increase by $0.5$ to $1.3$ points, Multilingual is unchanged, and Safety increases by $0.2$.
The one-bit representation thus preserves the uncompressed model's quality on Llama as well.

%% file: table/storage_memory_table.tex
\begin{table}[!t]
\centering
\caption{Weight storage and peak allocated GPU memory. Storage counts safetensors weight payloads only. ``Uncompressed MoE'' denotes the shared-backbone model before factorization.}
\label{tab:storage-budget}
\footnotesize
\setlength{\tabcolsep}{3pt}
\renewcommand{\arraystretch}{1.05}
\begin{adjustbox}{max width=\linewidth,center}
\begin{tabular}{@{}>{\raggedright\arraybackslash}p{0.42\linewidth}cc@{}}
\toprule
\textbf{Model} & \shortstack{\textbf{Weight storage}\\\textbf{(GB)}$\downarrow$} & \shortstack{\textbf{Peak allocated}\\\textbf{memory (GB)}$\downarrow$} \\
\midrule
One dense specialist & 17.5 ($1.0\times$) & 17.9 ($1.0\times$) \\
Four standalone specialists & 70.1 ($4.0\times$) & 65.6 ($3.7\times$) \\
Uncompressed MoE & 50.2 ($2.9\times$) & 50.6 ($2.8\times$) \\
\textbf{UNITE (one-bit + router)} & \textbf{20.3 ($1.16\times$)} & \textbf{21.0 ($1.17\times$)} \\
\bottomrule
\end{tabular}
\end{adjustbox}
\end{table}

%% file: table/moe-qwen3vl8b-main.tex
\begin{table*}[!t]
  \centering
  \caption{Task-level comparison of reference models and UNITE across 22 benchmarks.
    Reference models are evaluated by us on the same benchmark data with the same metrics; their architectures and training data differ.
    Bold and underlined scores denote the best and second-best distinct values across all rows, including ties.
    UNITE uses top-$1$ routing after router adaptation. Higher is better except ShareRobot (Traj.).
    Column tints mark the benchmark groups.}
  \label{tab:unite-leading-benchmarks}
  \fontsize{8.2}{10.2}\selectfont
  \setlength{\tabcolsep}{2.0pt}
  \def\unitegroupgap{6.8pt}
  \def\unitetintL{\dimexpr\tabcolsep+0.3pt\relax}
  \def\unitegroupgaptrim{4.8pt}
  \renewcommand{\arraystretch}{1.15}
  \definecolor{tintGrounding}{HTML}{C9BFE8}
  \definecolor{tintSpatial}{HTML}{A9C9F2}
  \definecolor{tintDriving}{HTML}{F3BCBB}
  \providecommand{\unitemainhead}[1]{\rotatebox{90}{\strut #1}}
  \begin{adjustbox}{max width=\textwidth,center}
    \begin{tabular}{@{}l@{\hspace{\unitegroupgap}}*{5}{>{\columncolor{tintGrounding!30}[\unitetintL][\tabcolsep]}c}@{\hspace{\unitegroupgap}}*{11}{>{\columncolor{tintSpatial!30}[\unitetintL][\tabcolsep]}c}@{\hspace{\unitegroupgap}}*{6}{>{\columncolor{tintDriving!30}[\unitetintL][\tabcolsep]}c}@{\hspace{\tabcolsep}}}
      \toprule
                                                                                                  & \multicolumn{5}{@{}>{\columncolor{tintGrounding!30}[\unitetintL][\tabcolsep]}c@{\hspace{\unitegroupgap}}}{\textbf{Grounding}} & \multicolumn{11}{@{}>{\columncolor{tintSpatial!30}[\unitetintL][\tabcolsep]}c@{\hspace{\unitegroupgap}}}{\textbf{Embodied Spatial}} & \multicolumn{6}{@{}>{\columncolor{tintDriving!30}[\unitetintL][\tabcolsep]}c@{\hspace{\tabcolsep}}}{\textbf{Driving}} \\
      \cmidrule(l{-\unitetintL}r{\unitegroupgaptrim}){2-6}\cmidrule(l{-\unitetintL}r{\unitegroupgaptrim}){7-17}\cmidrule(l{-\unitetintL}r{0pt}){18-23}
      \multicolumn{1}{@{}c@{\hspace{\unitegroupgap}}}{\raisebox{3.2em}[0pt][0pt]{\textbf{Model}}} & \unitemainhead{RoboSpatial}                                                                                                   & \unitemainhead{RefSpatial}                                                                                                          & \unitemainhead{PointArena} & \unitemainhead{RoboAfford} & \unitemainhead{ShareRobot ($\downarrow$)} & \unitemainhead{VSI} & \unitemainhead{MMSI} & \unitemainhead{MindCube} & \unitemainhead{EmbSpatial} & \unitemainhead{ERQA} & \unitemainhead{ScanQA} & \unitemainhead{SQA3D} & \unitemainhead{Scan2Cap} & \unitemainhead{ScanRefer} & \unitemainhead{Multi3DRef} & \unitemainhead{SAT} & \unitemainhead{MME-RealWorld} & \unitemainhead{MAPLM} & \unitemainhead{DriveAction} & \unitemainhead{NuScenesQA} & \unitemainhead{NuPlanQA} & \unitemainhead{LingoQA} \\
      \midrule
      \multicolumn{23}{@{}l@{}}{\textit{Closed-source reference models}} \\
      GPT-5.4~\cite{openai2026gpt54} & 53.5 & 15.7 & 37.9 & 29.4 & 0.32 & 52.6 & 31.3 & 45.3 & 73.2 & \underline{50.5} & 78.3 & 45.8 & 14.0 & 61.7 & 45.1 & 73.3 & 32.4 & 57.9 & \textbf{82.5} & 37.1 & \underline{87.5} & \textbf{76.8} \\
      Gemini 2.5 Pro~\cite{comanici2025gemini25} & 53.7 & 36.5 & 62.8 & 15.0 & 0.34 & 43.4 & \underline{38.0} & 57.6 & 78.7 & \textbf{55.7} & 67.0 & 37.3 & 16.8 & 65.9 & 54.4 & 78.7 & \underline{67.0} & 26.1 & 73.5 & 16.1 & 64.2 & 64.1 \\
      Claude Sonnet 4.6~\cite{anthropic2026sonnet46} & 37.3 & 4.0 & 14.9 & 15.2 & 0.39 & 17.6 & 30.9 & 43.7 & \textbf{87.9} & 41.0 & 31.5 & 14.0 & 8.3 & 51.5 & 40.2 & 64.7 & 24.4 & 56.7 & 80.0 & 22.9 & 78.2 & 60.0 \\
      \midrule
      \multicolumn{23}{@{}l@{}}{\textit{Embodied reference models}} \\
      ACE-Brain-0.5-8B~\cite{gong2026acebrain} & 60.1 & 55.6 & 68.5 & 75.1 & 0.32 & \underline{62.2} & 35.5 & \underline{86.3} & 75.9 & 46.3 & 99.2 & \underline{62.6} & 83.3 & \underline{70.2} & \underline{72.4} & \underline{82.7} & 66.6 & 71.3 & 78.2 & 44.7 & 86.7 & 55.6 \\
      Cosmos3 Nano~\cite{nvidia2026cosmos3} & 61.0 & 53.1 & 69.4 & \textbf{84.0} & 0.30 & 54.9 & 36.2 & 34.8 & \underline{82.9} & 46.0 & 60.9 & 44.2 & 9.2 & 5.4 & 8.1 & 80.7 & 57.0 & 26.7 & 79.9 & 16.6 & 82.5 & \underline{71.4} \\
      Embodied-R1.5~\cite{yuan2026embodiedr15} & 69.7 & 54.2 & \textbf{71.4} & 80.0 & 0.31 & 56.1 & 29.5 & 29.4 & 78.1 & 46.0 & 67.1 & 48.2 & 0.3 & 5.1 & 7.7 & 62.7 & 61.7 & 63.1 & 77.1 & 33.5 & 82.0 & 61.0 \\
      RynnBrain-8B~\cite{dang2026rynnbrain} & \textbf{73.1} & \underline{59.2} & 65.8 & 50.3 & 0.35 & \textbf{71.0} & \textbf{39.6} & 56.6 & 80.0 & 46.8 & 60.2 & 43.3 & 2.8 & 5.4 & 8.1 & 78.0 & \textbf{68.7} & 68.2 & 76.3 & 29.5 & 81.8 & 57.8 \\
      Vlaser-8B~\cite{yang2025vlaser} & 61.7 & \underline{59.2} & 60.3 & 18.3 & 0.35 & 60.3 & 27.2 & 34.6 & 75.1 & 41.0 & 55.5 & 44.6 & 0.2 & 3.1 & 8.2 & 66.7 & 41.6 & 29.1 & 78.1 & 33.1 & 78.3 & 59.6 \\
      MiMo-Embodied-7B~\cite{hao2025mimoembodied} & 61.8 & 48.0 & 3.9 & 69.8 & \textbf{0.15} & 48.5 & 31.7 & 32.3 & 76.2 & 46.8 & 83.7 & 54.2 & 61.8 & 64.7 & 35.9 & 78.7 & 60.3 & \underline{74.5} & \underline{81.0} & \textbf{56.7} & 73.7 & 69.9 \\
      Pelican-VL-7B~\cite{zhang2025pelicanvl} & 57.5 & 22.3 & 24.6 & 26.0 & 0.46 & 52.8 & 26.0 & 31.0 & 73.2 & 39.8 & 65.6 & 42.5 & 0.5 & 5.4 & 7.9 & 67.3 & 57.9 & 24.4 & 77.2 & 14.8 & 83.4 & 56.0 \\
      VeBrain-7B~\cite{luo2025vebrain} & 44.9 & 9.8 & 12.1 & 18.9 & 0.38 & 39.9 & 27.3 & 30.1 & 30.6 & 33.3 & \underline{101.5} & 61.6 & \underline{89.7} & 66.4 & 67.8 & 73.3 & 60.1 & 22.9 & 78.3 & 29.3 & 82.9 & 55.0 \\
      RoboBrain 2.5-8B~\cite{tan2026robobrain25} & \underline{73.0} & \textbf{60.5} & 59.3 & 74.9 & \underline{0.24} & 41.0 & 29.3 & 28.1 & 75.6 & 22.3 & 63.1 & 28.1 & 4.8 & 2.3 & 8.1 & 63.3 & 60.0 & 22.5 & 80.5 & 33.2 & 79.3 & 48.0 \\
      \midrule
      UNITE (ours) & 58.3 & 57.2 & \underline{71.3} & \underline{80.3} & 0.25 & 46.7 & 36.3 & \textbf{87.9} & 78.9 & 47.3 & \textbf{105.6} & \textbf{62.8} & \textbf{102.1} & \textbf{71.0} & \textbf{73.2} & \textbf{90.7} & 65.7 & \textbf{93.8} & 79.8 & \underline{48.7} & \textbf{91.2} & 63.0 \\
      \bottomrule
    \end{tabular}
  \end{adjustbox}
\end{table*}

%% file: table/r2r-main.tex
\begin{table}[t]
\centering
\caption{R2R Val-Unseen. Navigation references are evaluated by us on the same episodes. NE is in meters; other metrics are percentages. Bold and underlined scores mark the best and second-best distinct values across all rows.}
\label{tab:r2r-main}
\fontsize{9}{11}\selectfont
\setlength{\tabcolsep}{3pt}
\renewcommand{\arraystretch}{1.00}
\begin{adjustbox}{max width=\linewidth,center}
\begin{tabular}{@{}p{92pt}rrrr@{}}
\toprule
\textbf{Model} & NE$\downarrow$ & OS$\uparrow$ & SR$\uparrow$ & SPL$\uparrow$ \\
\midrule
\multicolumn{5}{l}{\textit{Navigation references}} \\
\raggedright NaVid~\cite{zhang2024navid} & 5.7 & 49.2 & 41.9 & 36.5 \\
\raggedright Uni-NaVid~\cite{zhang2024uninavid} & 5.6 & 53.3 & 47.0 & 42.7 \\
\raggedright NaVILA~\cite{cheng2025navila} & 5.2 & 62.5 & 54.0 & 49.0 \\
\raggedright StreamVLN~\cite{wei2026streamvln} & 5.0 & 64.2 & 56.9 & \underline{51.9} \\
\raggedright NavFoM~\cite{zhang2026navfom} & 5.0 & 64.9 & 56.2 & 51.2 \\
\raggedright RynnBrain-Nav~\cite{dang2026rynnbrain} & 4.9 & \textbf{71.6} & \underline{58.6} & 49.6 \\
\raggedright Qwen-VLA-Instruct~\cite{wang2026qwenvla} & 5.1 & \underline{69.0} & 57.5 & 51.2 \\
\raggedright ACE-Brain-0.5~\cite{gong2026acebrain} & \underline{4.8} & 63.7 & 57.4 & 51.7 \\
\midrule
\raggedright UNITE (ours) & \textbf{4.5} & 65.7 & \textbf{59.6} & \textbf{53.4} \\
\bottomrule
\end{tabular}
\end{adjustbox}
\end{table}

%% file: table/setup_models_table.tex
\begin{table*}[!t]
  \centering
  \caption{Controlled comparison of source specialists, static merging methods, and successive UNITE stages across all 22 benchmarks.
    All merging methods and UNITE use the same source specialists; UNITE uses top-$1$ routing at every stage.
    Bold and underlined scores denote the best and second-best distinct values, including ties: source specialists are ranked separately, while static merging methods and UNITE stages are ranked jointly. The base model is excluded from highlighting.
    Higher is better except ShareRobot (Traj.); MMSI and DriveAction report overall accuracy.}
  \label{tab:setup-models}
  \fontsize{8.2}{10.2}\selectfont
  \setlength{\tabcolsep}{1.0pt}
  \def\unitegroupgap{4.8pt}
  \def\unitetintL{\dimexpr\tabcolsep+0.3pt\relax}
  \def\unitegroupgaptrim{\dimexpr\unitegroupgap-\tabcolsep\relax}
  \providecolor{tintGrounding}{HTML}{C9BFE8}
  \providecolor{tintSpatial}{HTML}{A9C9F2}
  \providecolor{tintDriving}{HTML}{F3BCBB}
  \renewcommand{\arraystretch}{1.15}
  \providecommand{\unitebenchmarkhead}[1]{\rotatebox{90}{\strut #1}}
  \begin{adjustbox}{max width=\textwidth,center}
    \begin{tabular}{@{}l@{\hspace{\unitegroupgap}}*{5}{>{\columncolor{tintGrounding!30}[\unitetintL][\tabcolsep]}c}@{\hspace{\unitegroupgap}}*{11}{>{\columncolor{tintSpatial!30}[\unitetintL][\tabcolsep]}c}@{\hspace{\unitegroupgap}}*{6}{>{\columncolor{tintDriving!30}[\unitetintL][\tabcolsep]}c}@{\hspace{\tabcolsep}}}
      \toprule
                                                                                                  & \multicolumn{5}{@{}>{\columncolor{tintGrounding!30}[\unitetintL][\tabcolsep]}c@{\hspace{\unitegroupgap}}}{\textbf{Grounding}} & \multicolumn{11}{@{}>{\columncolor{tintSpatial!30}[\unitetintL][\tabcolsep]}c@{\hspace{\unitegroupgap}}}{\textbf{Embodied Spatial}} & \multicolumn{6}{@{}>{\columncolor{tintDriving!30}[\unitetintL][\tabcolsep]}c@{\hspace{\tabcolsep}}}{\textbf{Driving}} \\
      \cmidrule(l{-\unitetintL}r{\unitegroupgaptrim}){2-6}\cmidrule(l{-\unitetintL}r{\unitegroupgaptrim}){7-17}\cmidrule(l{-\unitetintL}r{0pt}){18-23}
      \multicolumn{1}{@{}c@{\hspace{\unitegroupgap}}}{\raisebox{3.2em}[0pt][0pt]{\textbf{Model}}} & \unitebenchmarkhead{RoboSpatial}                                                                                              & \unitebenchmarkhead{RefSpatial}                                                                                                     & \unitebenchmarkhead{PointArena} & \unitebenchmarkhead{RoboAfford} & \unitebenchmarkhead{ShareRobot ($\downarrow$)} & \unitebenchmarkhead{VSI} & \unitebenchmarkhead{MMSI} & \unitebenchmarkhead{MindCube} & \unitebenchmarkhead{EmbSpatial} & \unitebenchmarkhead{ERQA} & \unitebenchmarkhead{ScanQA} & \unitebenchmarkhead{SQA3D} & \unitebenchmarkhead{Scan2Cap} & \unitebenchmarkhead{ScanRefer} & \unitebenchmarkhead{Multi3DRef} & \unitebenchmarkhead{SAT} & \unitebenchmarkhead{MME-RealWorld} & \unitebenchmarkhead{MAPLM} & \unitebenchmarkhead{DriveAction} & \unitebenchmarkhead{NuScenesQA} & \unitebenchmarkhead{NuPlanQA} & \unitebenchmarkhead{LingoQA} \\
      \midrule
      \multicolumn{23}{l}{\textit{Base model and source specialists}} \\
      \textbf{Base} (Qwen3-VL-8B-Instruct) & 62.4 & 53.1 & 62.3 & 75.4 & 0.37 & 59.4 & 31.0 & 29.4 & 78.5 & 45.8 & 72.9 & 49.5 & 2.6 & 0.1 & 0.1 & 66.0 & 63.3 & 31.5 & 79.0 & 22.8 & 82.2 & 57.0 \\
      Grounding expert & \textbf{68.0} & \textbf{57.8} & \textbf{68.3} & \textbf{77.7} & \textbf{0.25} & 56.3 & \underline{34.1} & \textbf{87.6} & \textbf{79.4} & \textbf{47.3} & 92.8 & 55.2 & 75.7 & 65.6 & 66.4 & 86.7 & \underline{68.9} & \textbf{88.4} & \underline{79.1} & \underline{28.9} & 86.7 & 55.6 \\
      Navigation expert & 60.6 & \underline{54.2} & 67.7 & 75.1 & \underline{0.26} & 62.9 & 33.3 & 85.3 & 74.5 & \underline{44.8} & 93.5 & \underline{61.4} & \underline{78.8} & \underline{70.1} & \underline{72.1} & 84.7 & 67.4 & \underline{83.6} & 78.7 & 6.6 & 86.3 & 53.6 \\
      Embodied Spatial expert & \underline{60.7} & 53.4 & \underline{68.0} & \underline{77.2} & \textbf{0.25} & \textbf{63.8} & \textbf{36.5} & \underline{85.8} & 76.1 & \underline{44.8} & \textbf{97.7} & \textbf{61.7} & \textbf{83.0} & \textbf{70.2} & \textbf{72.4} & \underline{87.3} & 66.3 & 81.3 & 78.6 & 7.8 & \underline{86.9} & \underline{56.2} \\
      Driving expert & 55.6 & 26.0 & 44.7 & 56.5 & 0.46 & \underline{63.1} & 32.2 & 82.1 & \underline{77.8} & 41.5 & \underline{97.3} & 54.5 & 75.2 & 61.4 & 55.9 & \textbf{92.0} & \textbf{71.2} & 77.8 & \textbf{81.3} & \textbf{58.8} & \textbf{91.7} & \textbf{65.8} \\
      \midrule
      \multicolumn{23}{@{}l@{}}{\textit{Static merging methods}} \\
      Model Soup~\cite{wortsman2022modelsoupsaveragingweights} & \textbf{62.6} & \textbf{60.5} & 69.3 & 79.8 & \textbf{0.25} & \textbf{63.6} & \underline{37.1} & \textbf{88.3} & 77.9 & 44.8 & 101.8 & 61.6 & 85.6 & \textbf{71.3} & 73.2 & \textbf{90.7} & \underline{70.5} & 93.1 & \textbf{80.4} & 40.1 & 90.2 & 62.2 \\
      Task Arithmetic~\cite{ilharco2023editingmodelstaskarithmetic} & 61.3 & \underline{59.7} & 68.3 & 78.4 & \textbf{0.25} & \underline{63.5} & 37.0 & 88.0 & 78.0 & 45.0 & 101.8 & 61.6 & 85.6 & \textbf{71.3} & \underline{73.5} & \textbf{90.7} & \underline{70.5} & 93.1 & \underline{80.2} & \underline{48.6} & 90.2 & 62.0 \\
      TIES Merging~\cite{yadav2023tiesmergingresolvinginterferencemerging} & 47.7 & 11.6 & 17.4 & 27.2 & 0.37 & 41.0 & 27.8 & 57.8 & 46.5 & 32.3 & 5.9 & 19.3 & 10.1 & 19.7 & 23.0 & 72.0 & 39.8 & 88.2 & 45.7 & 9.7 & 73.1 & 33.8 \\
      DARE~\cite{yuLanguageModelsAre2023} & 61.7 & 58.1 & 69.8 & \underline{80.2} & \textbf{0.25} & 63.2 & 36.2 & 87.1 & 77.9 & 45.0 & 98.2 & 60.7 & 86.4 & \underline{71.1} & \textbf{73.6} & \textbf{90.7} & \textbf{70.6} & \underline{93.7} & \textbf{80.4} & 47.3 & \underline{90.5} & \textbf{63.4} \\
      Fisher Merging~\cite{michaelmatenaMergingModelsFisherWeighted2022} & \underline{62.1} & 58.5 & 69.5 & 77.6 & 0.27 & 63.3 & 35.1 & \underline{88.2} & 77.3 & 45.8 & 98.5 & \underline{62.8} & 84.4 & 70.8 & 73.3 & \underline{89.3} & 68.7 & 90.1 & 79.2 & 19.2 & 87.7 & 59.0 \\
      RegMean~\cite{xisenjinDatalessKnowledgeFusion2023} & 50.5 & 41.5 & 55.6 & 56.4 & 0.36 & 55.4 & 32.1 & 41.5 & 68.6 & 38.5 & 74.4 & 45.2 & 42.5 & 56.6 & 52.5 & 72.0 & 49.6 & 82.4 & 63.6 & 34.5 & 73.3 & 52.8 \\
      \midrule
      \multicolumn{23}{l}{\textit{UNITE stages}} \\
      Uncompressed MoE & 58.0 & 56.1 & 70.6 & 79.6 & 0.34 & 62.4 & 36.4 & 87.8 & 78.3 & 44.8 & 104.3 & 62.4 & 97.1 & \textbf{71.3} & \textbf{73.6} & 85.3 & 65.6 & 91.7 & 79.5 & 27.7 & 89.3 & 60.4 \\
      Shared-anchor one-bit MoE & 58.9 & 59.4 & \textbf{71.7} & 79.6 & \underline{0.26} & 47.4 & \textbf{37.5} & 88.1 & \textbf{80.1} & \underline{47.0} & \textbf{106.6} & \textbf{63.1} & \underline{100.6} & \textbf{71.3} & \textbf{73.6} & 87.3 & 65.9 & 92.0 & 79.5 & 29.7 & 89.8 & 59.2 \\
      One-bit MoE + router adaptation & 58.3 & 57.2 & \underline{71.3} & \textbf{80.3} & \textbf{0.25} & 46.7 & 36.3 & 87.9 & \underline{78.9} & \textbf{47.3} & \underline{105.6} & \underline{62.8} & \textbf{102.1} & 71.0 & 73.2 & \textbf{90.7} & 65.7 & \textbf{93.8} & 79.8 & \textbf{48.7} & \textbf{91.2} & \underline{63.0} \\
      \bottomrule
    \end{tabular}
  \end{adjustbox}
\end{table*}

%% file: table/stage_r2r.tex
\begin{table}[t]
\centering
\caption{R2R Val-Unseen: source experts, static merging methods, and UNITE stages with top-$1$ routing. NE is in meters; other metrics are percentages. Bold and underlined scores mark the best and second-best distinct values. Experts are ranked separately; static merging and UNITE stages are ranked jointly. The base model is excluded from highlighting.}
\label{tab:setup-r2r}
\footnotesize
\setlength{\tabcolsep}{2pt}
\renewcommand{\arraystretch}{1.10}
\begin{adjustbox}{max width=\linewidth,center}
\begin{tabular}{@{}p{0.47\linewidth}cccc@{}}
\toprule
\multicolumn{1}{c}{\multirow{2}{*}{\textbf{Model}}} & \multicolumn{4}{c}{\textbf{R2R Val-Unseen}} \\
\cmidrule(lr){2-5}
 & NE$\downarrow$ & OS$\uparrow$ & SR$\uparrow$ & SPL$\uparrow$ \\
\midrule
\multicolumn{5}{l}{\textit{Base model and source specialists}} \\
\textbf{Base} Qwen3-VL-8B-Instruct & 9.1 & 1.1 & 0.1 & 0.1 \\
Grounding expert & 7.5 & 31.0 & 28.2 & 26.3 \\
Navigation expert & \textbf{4.7} & \textbf{63.7} & \textbf{57.5} & \textbf{51.2} \\
Embodied Spatial expert & \underline{4.9} & \underline{62.9} & \underline{56.8} & \underline{50.3} \\
Driving expert & 8.9 & 0.5 & 0.1 & 0.0 \\
\midrule
\multicolumn{5}{l}{\textit{Static merging methods}} \\
\raggedright Model Soup~\cite{wortsman2022modelsoupsaveragingweights} & 5.0 & 59.1 & 54.3 & 50.5 \\
\raggedright Task Arithmetic~\cite{ilharco2023editingmodelstaskarithmetic} & 5.7 & 60.2 & 48.7 & 43.0 \\
\raggedright TIES Merging~\cite{yadav2023tiesmergingresolvinginterferencemerging} & 7.7 & 50.6 & 30.9 & 23.1 \\
\raggedright DARE~\cite{yuLanguageModelsAre2023} & 5.8 & 60.3 & 48.8 & 43.1 \\
\raggedright Fisher Merging~\cite{michaelmatenaMergingModelsFisherWeighted2022} & 5.2 & \underline{63.5} & 55.2 & 48.0 \\
\raggedright RegMean~\cite{xisenjinDatalessKnowledgeFusion2023} & 8.0 & 40.5 & 30.2 & 26.3 \\
\midrule
\multicolumn{5}{l}{\textit{UNITE stages}} \\
Uncompressed MoE & 5.0 & 61.6 & 55.9 & 50.6 \\
Shared-anchor one-bit MoE & \underline{4.9} & 62.1 & \underline{57.0} & \underline{51.9} \\
One-bit MoE + router adaptation & \textbf{4.5} & \textbf{65.7} & \textbf{59.6} & \textbf{53.4} \\
\bottomrule
\end{tabular}
\end{adjustbox}
\end{table}

%% file: table/compression_router.tex
\begin{table}[t]
\centering
\caption{Factorization and router adaptation under the same training budget and supervision mixture. All models use top-$1$; R2R uses Val-Unseen.}
\label{tab:compression-router}
\footnotesize
\setlength{\tabcolsep}{3pt}
\renewcommand{\arraystretch}{1.05}
\begin{tabular}{@{}lcccc@{}}
\toprule
 & \multicolumn{2}{c}{\textbf{Before adaptation}} & \multicolumn{2}{c}{\textbf{After adaptation}} \\
\cmidrule(lr){2-3}\cmidrule(lr){4-5}
\textbf{Benchmark} & Uncomp. & Comp. & Uncomp. & Comp. \\
\midrule
ScanQA (CIDEr) & 104.3 & 106.6 & \textbf{111.1} & 105.6 \\
SQA3D (EM-1) & 62.4 & 63.1 & \textbf{63.5} & 62.8 \\
Scan2Cap (CIDEr@0.25) & 97.1 & 100.6 & 94.1 & \textbf{102.1} \\
ScanRefer (Acc@0.25) & \textbf{71.3} & \textbf{71.3} & \textbf{71.3} & 71.0 \\
Multi3DRef (F1@0.25) & 73.6 & 73.6 & \textbf{73.7} & 73.2 \\
VSI & 62.4 & 47.4 & \textbf{63.7} & 46.7 \\
NuScenesQA & 27.7 & 29.7 & 24.4 & \textbf{48.7} \\
R2R SR (\%) & 55.9 & 57.0 & 59.2 & \textbf{59.6} \\
R2R SPL (\%) & 50.6 & 51.9 & 53.2 & \textbf{53.4} \\
\bottomrule
\end{tabular}
\end{table}

%% file: section/conclusion.tex
\section{Conclusion}
\label{sec:conclusion}

We presented UNITE, a framework that consolidates dense embodied VLM specialists into a shared-backbone mixture of experts whose expert bank is stored through a shared-anchor one-bit factorization.
Under a one-bit deviation budget, the shared anchor is optimized jointly with the expert-specific row scales and binary signs through closed-form alternating updates that require only the source weights.
Combined with router adaptation of fewer than $0.6$M parameters ($<0.01\%$ of the $8.8$B active parameters), UNITE consolidates four Qwen3-VL-8B specialists into a model whose total weight storage is $1.16\times$ that of a single specialist.
Across $22$ embodied understanding benchmarks and R2R navigation, UNITE surpasses every source specialist on all five 3D scene understanding benchmarks and all four navigation metrics; the factorization itself is nearly lossless relative to the uncompressed MoE on $21$ of the $22$ understanding benchmarks.

For the remaining exception, the video benchmark VSI-Bench, frame-count experiments show that the compressed model gains little from additional frames, suggesting impaired use of cross-frame evidence.
These findings motivate investigating whether a larger bit budget for selected experts or layers can recover this capability.
Extending the framework to vision-language-action policies with action heads and to more divergent specialists is a natural direction for future work.

%% file: section/supplement.tex
\onecolumn
\markboth{UNITE: Supplementary Material}{UNITE: Supplementary Material}
\raggedbottom
\begin{center}
{\LARGE\bfseries Supplementary Material\par}
\end{center}

\newcommand{\mainref}[1]{\ref*{#1}}
\newcommand{\mainequation}[1]{\textup{(\ref*{#1})}}

\setlength{\cftbeforesecskip}{0.2em}
\setlength{\cftbeforesubsecskip}{0.2em}
\setlength{\cftbeforesubsubsecskip}{0.2em}
\setlength{\cftsecnumwidth}{2.2em}
\setlength{\cftsubsecindent}{2.2em}
\setlength{\cftsubsecnumwidth}{2.8em}

This supplement contains \hyperref[sec:appendix-method]{Methods and Implementation} (M), \hyperref[sec:appendix-topk]{Detailed Results} (D), and \hyperref[sec:qualitative-results]{Qualitative Examples} (Q).
The letter prefixes distinguish supplementary sections, tables, figures, and equations; unprefixed references point to the main manuscript.

\vskip 0.1in
\startcontents[sections]
\printcontents[sections]{}{1}{\setcounter{tocdepth}{2}}
\vskip 0.1in
\hrule
\vskip 0.1in

\newcommand{\supplementprefix}[1]{%
  \setcounter{section}{0}\setcounter{table}{0}%
  \setcounter{figure}{0}\setcounter{equation}{0}%
  \renewcommand{\thesection}{#1\arabic{section}}%
  \renewcommand{\thesubsection}{\thesection.\arabic{subsection}}%
  \renewcommand{\thetable}{#1\arabic{table}}%
  \renewcommand{\thefigure}{#1\arabic{figure}}%
  \renewcommand{\theequation}{#1\arabic{equation}}%
  \renewcommand{\theHsection}{#1.\arabic{section}}%
  \renewcommand{\theHtable}{#1.\arabic{table}}%
  \renewcommand{\theHfigure}{#1.\arabic{figure}}%
  \renewcommand{\theHequation}{#1.\arabic{equation}}%
}
\supplementprefix{M}

\section{Method Details}
\label{sec:appendix-method}

\paragraph*{Expert parameterization.}
Each specialist contributes a complete SwiGLU block at every decoder layer.
Let $d$ and $m$ denote the hidden and intermediate dimensions, respectively.
For specialist $e$ at layer $\ell$, the gate and up matrices satisfy $\mathbf{W}_{\ell,e}^{\mathrm{gate}},\mathbf{W}_{\ell,e}^{\mathrm{up}}\in\mathbb{R}^{m\times d}$, and the down matrix satisfies $\mathbf{W}_{\ell,e}^{\mathrm{down}}\in\mathbb{R}^{d\times m}$.
For input $\mathbf{x}\in\mathbb{R}^{d}$, the expert computes
\begin{equation}
    F_{\ell,e}(\mathbf{x})
    =
    \mathbf{W}_{\ell,e}^{\mathrm{down}}
    \left[
        \operatorname{SiLU}\!\left(\mathbf{W}_{\ell,e}^{\mathrm{gate}}\mathbf{x}\right)
        \odot
        \left(\mathbf{W}_{\ell,e}^{\mathrm{up}}\mathbf{x}\right)
    \right],
    \label{eq:app_swiglu}
\end{equation}
where $\odot$ denotes elementwise multiplication.
We store the gate and up matrices in Eq.~\eqref{eq:app_swiglu} as the row-concatenated matrix $\mathbf{W}_{\ell,e}^{\mathrm{gate\text{-}up}}=[(\mathbf{W}_{\ell,e}^{\mathrm{gate}})^{\mathsf T},(\mathbf{W}_{\ell,e}^{\mathrm{up}})^{\mathsf T}]^{\mathsf T}\in\mathbb{R}^{2m\times d}$.
The two factorization groups at each layer are therefore the gate-up stack of shape $E\times2m\times d$ and the down stack of shape $E\times d\times m$.
Gate and up retain separate row scales despite being packed into one group, and the router selects both projections together with the corresponding down projection.

\paragraph*{Shared-backbone fusion.}
We apply norm-rescaled averaging to decoder self-attention projections.
For corresponding query, key, value, or output matrices $\{\mathbf{P}_e\}_{e=1}^{E}$, define $\boldsymbol{\mu}=E^{-1}\sum_e\mathbf{P}_e$ and $\overline{n}=E^{-1}\sum_e\|\mathbf{P}_e\|_F$.
The shared projection is
$$
\mathbf{P}_{\mathrm{shared}}
=
\begin{cases}
        \boldsymbol{\mu}\,\overline{n}/\|\boldsymbol{\mu}\|_F,
        & \|\boldsymbol{\mu}\|_F\geq10^{-10},\\
        \boldsymbol{\mu}, & \text{otherwise}.
\end{cases}
$$
We compute means and norms in FP32 and cast fused tensors to the source storage dtype.
The visual encoder, normalization parameters, token embeddings, output head, and any decoder attention biases use the elementwise mean across specialists.

\paragraph*{Weight-derived router initialization.}
The router initializer extracts expert differences in the input space of the feed-forward block.
Writing $\overline{\mathbf{W}}_{\ell}^{\mathrm{gate}}$ and $\overline{\mathbf{W}}_{\ell}^{\mathrm{up}}$ for the respective means across specialists, we form
$$
\mathbf{X}_{\ell,e}
=
\begin{bmatrix}
        \mathbf{W}_{\ell,e}^{\mathrm{gate}}-\overline{\mathbf{W}}_{\ell}^{\mathrm{gate}}\\
        \mathbf{W}_{\ell,e}^{\mathrm{up}}-\overline{\mathbf{W}}_{\ell}^{\mathrm{up}}
\end{bmatrix}
\in\mathbb{R}^{2m\times d}.
$$
We compute an approximate leading right singular vector using randomized SVD in FP32, with a subspace size $q=\min(6,2m,d)$ and four subspace iterations, and use it in Eq.~\mainequation{eq:svd_router_init}.
If $\max_{o,i}|X_{\ell,e,o,i}|<10^{-8}$, we instead initialize that router row to zero.
We retain the singular-vector sign returned by the numerical routine without sign canonicalization, then cast the router weights to the source storage dtype.
An exact leading right singular vector maximizes the centered projections' combined output norm over unit-norm input directions.
This construction provides a weight-derived starting point for expert selection; router adaptation (Section~\mainref{sec:router_sft}) then refines it.

\paragraph*{Conditional least-squares updates.}
Each alternating step solves a conditional subproblem of Eq.~\mainequation{eq:onebit_objective}.
For a fixed anchor, consider one residual row $\boldsymbol{\delta}\in\mathbb{R}^{I}$, its nonnegative scale $a$, and its binary sign vector $\mathbf{s}\in\{-1,+1\}^{I}$.
Its squared error is
$$
\|\boldsymbol{\delta}-a\mathbf{s}\|_2^2
=\|\boldsymbol{\delta}\|_2^2+Ia^2-2a\boldsymbol{\delta}^{\mathsf T}\mathbf{s}.
$$
For any $a\geq0$, choosing $s_i=\operatorname{sign}(\delta_i)$ maximizes the last inner product, with either sign optimal when $\delta_i=0$.
Using the positive sign at zero, minimization over $a$ gives $a=I^{-1}\sum_i|\delta_i|$, yielding Eq.~\mainequation{eq:sign_scale_update}.
For fixed signs and scales, differentiating the objective with respect to the anchor and setting the derivative to zero gives the average in Eq.~\mainequation{eq:anchor_update}.
Consequently, the objective is non-increasing over the alternating updates in exact arithmetic.

The numerical settings used in the experiments (ten FP32 iterations followed by BF16 storage) are given in Section~\ref{sec:appendix-implementation}, and the measured reconstruction error per iteration in Table~\ref{tab:recon-vs-T}.

\paragraph*{Direct evaluation of factorized experts.}
The factorized implementation applies each factorized projection as $\widehat{\mathbf{W}}_e\mathbf{x}=\mathbf{C}\mathbf{x}+\boldsymbol{\alpha}_e\odot(\mathbf{S}_e\mathbf{x})$, without constructing a full dense expert parameter.
Within a layer forward pass, tokens are grouped by selected expert, and the default PyTorch path temporarily expands the signs for each expert selected by at least one token.
Thus, top-$1$ routing selects one expert per token, while different tokens in a batch may collectively select every expert.

\section{Implementation and Training Details}
\label{sec:appendix-implementation}

\paragraph*{Model construction and compression.}
The embodied model uses four Qwen3-VL-8B specialists, each with $36$ decoder layers, hidden dimension $d=4096$, and intermediate dimension $m=12288$.
Each layer contains two independently factorized weight groups, gate-up and down, each spanning all four experts, giving $72$ groups in total.
For each group, we initialize the anchor with the expert mean, perform $T=10$ alternating updates in FP32 (Algorithm~\mainref{alg:joint_factorization}), and round the final anchor to BF16.
We then recompute the binary signs and row scales relative to the stored anchor and store the recomputed row scales in BF16.
For storage, we encode each sign $S_{e,o,i}\in\{-1,+1\}$ as a bit $b_{e,o,i}=(S_{e,o,i}+1)/2\in\{0,1\}$, mapping negative residuals to $0$ and nonnegative residuals to $1$.
We pack the bits into 32-bit words along each matrix's input dimension and decode them at inference time as $S_{e,o,i}=2b_{e,o,i}-1$.
For a group of $E$ expert matrices with shape $O\times I$, the anchor, signs, and row scales require $16OI+EOI+16EO$ bits before packing alignment.
This is $1+16/E+16/I$ bits per logical expert weight, or approximately five bits for the four-expert bank, including the amortized anchor and scales.

\paragraph*{Reconstruction error per iteration.}
Table~\ref{tab:recon-vs-T} lists the values underlying Fig.~\mainref{fig:recon-vs-T}.
With the fixed mean anchor ($t=0$), the summed one-bit reconstruction error is a median $36.7\%$ (gate-up) and $34.5\%$ (down) of the experts' total squared deviation from their mean, $\sum_e\|\mathbf{W}_e-\overline{\mathbf{W}}\|_F^2$.
After joint fitting and BF16 storage, it is $29.2\%$ and $26.9\%$, respectively.

\begin{table}[ht]
  \centering
  \caption{Summed reconstruction error $\sum_e\|\mathbf{W}_e-\widehat{\mathbf{W}}_e\|_F^2$ relative to the fixed mean anchor ($t=0$, $=100$), median over the $36$ decoder layers. Column $t$ uses the anchor $\mathbf{C}^{(t)}$ with its conditionally optimal signs and scales; ``Stored'' is the BF16 archive used at inference, with 10th--90th percentile ranges of $78.5$--$80.6$ (gate-up) and $77.2$--$78.9$ (down).}
  \label{tab:recon-vs-T}
  \footnotesize
  \setlength{\tabcolsep}{4pt}
  \begin{adjustbox}{max width=\linewidth,center}
    \begin{tabular}{@{}lcccccc@{}}
      \toprule
      Group   & $t=1$  & $t=2$  & $t=3$  & $t=5$  & $t=9$  & Stored \\
      \midrule
      Gate-up & $84.7$ & $81.3$ & $80.3$ & $79.8$ & $78.8$ & $79.6$ \\
      Down    & $82.5$ & $78.9$ & $77.8$ & $77.3$ & $77.1$ & $77.9$ \\
      \bottomrule
    \end{tabular}
  \end{adjustbox}
\end{table}

\paragraph*{Storage accounting.}
Weight storage counts the safetensors payloads of all parameters, including the vision encoder, embeddings, shared backbone, routers, and expert bank; peak GPU memory is the peak allocated memory during evaluation (Table~\mainref{tab:storage-budget}).
With $d=4096$ and $m=12288$, each layer holds $1.5\times10^{8}$ expert weights per specialist across its gate-up and down groups.
The four-expert bank therefore occupies $43.5$ GB in BF16 and $13.6$ GB after factorization, and the shared backbone, vision encoder, embeddings, and routers add $6.7$ GB to both, giving the $50.2$ GB and $20.3$ GB totals.
Table~\ref{tab:supp-storage-params} lists storage together with logical parameter counts.

\begin{table}[!ht]
  \centering
  \caption{Weight storage and logical parameter counts. Active counts include the shared parameters and $k$ complete expert branches.}
  \label{tab:supp-storage-params}
  \footnotesize
  \setlength{\tabcolsep}{4pt}
  \renewcommand{\arraystretch}{1.08}
  \begin{tabular}{@{}lr@{}}
    \toprule
    \multicolumn{2}{@{}l}{\textit{Weight storage (GB)$\downarrow$}} \\
    One dense specialist             & 17.5 \\
    Four standalone specialists      & 70.1 \\
    Uncompressed shared-backbone MoE & 50.2 \\
    Shared-anchor one-bit MoE        & \textbf{20.3} \\
    \midrule
    \multicolumn{2}{@{}l}{\textit{Logical parameters (B)}} \\
    Dense specialist, total          & 8.8 \\
    MoE, total logical               & 25.1 \\
    MoE, active top-$1$              & 8.8 \\
    MoE, active top-$2$              & 14.2 \\
    MoE, active top-$3$              & 19.6 \\
    \bottomrule
  \end{tabular}
\end{table}

\paragraph*{Router adaptation.}
We train only the $36$ router matrices ($589{,}824$ parameters) with target-token language-modeling loss and a load-balancing loss weighted by $10^{-3}$.
Training runs for one epoch on eight A800 GPUs with learning rate $10^{-3}$, cosine decay, warmup ratio $0.03$, and seed $42$.
The per-device batch size is one, with four gradient accumulation steps, giving an effective global batch size of $32$.
Training selects two experts per token, while the exported model uses top-$1$ unless the evaluation explicitly varies $k$.
Under Eq.~\mainequation{eq:router} of the main manuscript, top-$2$ routing permits gradients through the selected mixture weights, whereas the normalized mixture coefficient is identically one for top-$1$.

\paragraph*{Router supervision mixture.}
Router adaptation uses $55{,}757$ supervised samples from eighteen sources spanning navigation, 3D referring, grounding, affordance, trajectory, video spatial reasoning, spatial question answering, egocentric planning, and general visual question answering.
The uncompressed MoE (Table~\mainref{tab:compression-router}) uses the same mixture and training budget.

\paragraph*{Static merging configurations.}
The six merging baselines use the same four source checkpoints as UNITE.
Model Soup averages the source weights without including the pretrained base as an additional model.
Task Arithmetic uses a task-vector scale of $0.2$.
DARE uses a drop probability of $0.5$, a task-vector scale of $0.25$, and seed $42$.
TIES retains $20\%$ of task-vector entries, uses sum aggregation after sign resolution, and applies scale $1.0$.

\paragraph*{Calibration data.}
Fisher Merging and RegMean use the same $256$-example multimodal calibration set, drawn from the training splits of $20$ sources with fixed quotas and seed $1234$ (Table~\ref{tab:merging-calibration}).
The quotas allocate $64$ examples to each of the four specialist domains, so that every specialist's capability is represented equally in the Fisher and Gram statistics.
NuScenes-QA examples are drawn from training scenes with available camera images.
All four specialists receive the same preprocessed inputs, one example per batch, and compute their own Fisher or Gram statistics.
An example may contain multiple images; $9$ of the $256$ examples are VSI-590K videos, and all other visual inputs are images.

\begin{table}[ht]
  \centering
  \caption{Calibration-set composition for Fisher Merging and RegMean.}
  \label{tab:merging-calibration}
  \footnotesize
  \setlength{\tabcolsep}{3pt}
  \renewcommand{\arraystretch}{1.05}
  \begin{tabular}{@{}p{0.78\linewidth}r@{}}
    \toprule
    Source                             & Examples \\
    \midrule
    \multicolumn{2}{@{}l}{\textit{Grounding}} \\
    Point XY                           & $4$ \\
    ShareRobot Affordance              & $14$ \\
    ShareRobot Trajectory              & $14$ \\
    PixMo-Points (at most $10$ points) & $8$ \\
    PACO-LVIS v1 train                 & $8$ \\
    RefSpatial 3D                      & $8$ \\
    RefSpatial Simulator               & $8$ \\
    \midrule
    \multicolumn{2}{@{}l}{\textit{Navigation}} \\
    R2R (action-word annotations)      & $40$ \\
    RxR (action-word annotations)      & $24$ \\
    \midrule
    \multicolumn{2}{@{}l}{\textit{Embodied spatial}} \\
    VSI-590K                           & $12$ \\
    MMSI                               & $12$ \\
    MindCube train                     & $12$ \\
    SAT train                          & $12$ \\
    GPT4Scene train                    & $16$ \\
    \midrule
    \multicolumn{2}{@{}l}{\textit{Driving}} \\
    NuScenes-QA train                  & $16$ \\
    NuPlanQA train                     & $12$ \\
    MAPLM train                        & $12$ \\
    LingoQA Action train               & $8$ \\
    LingoQA Scenery train              & $8$ \\
    CODA-LM train                      & $8$ \\
    \midrule
    Total                              & $256$ \\
    \bottomrule
  \end{tabular}
\end{table}

\paragraph*{Fisher Merging.}
For each specialist and calibration example, we select the highest-probability next token at the final sequence position as a hard self-label and compute its negative log-likelihood.
The diagonal Fisher estimate is the mean of the per-example squared parameter gradients over the $256$ examples, without global L2 normalization.
Each parameter coordinate is merged by averaging the four specialist values with weights proportional to their Fisher estimates plus $\delta=10^{-10}$.

\paragraph*{RegMean.}
For each specialist's linear layer, we collect the uncentered input Gram matrix $\mathbf{G}_i=\mathbf{X}_i^{\mathsf T}\mathbf{X}_i/N_i$, where $N_i$ is the total number of input activation rows observed for that layer.
Before merging, we multiply the off-diagonal entries by $0.5$ while leaving the diagonal unchanged, yielding $\mathbf{G}'_i$.
For output-by-input weight matrices, the merged weight satisfies
\begin{equation*}
  \mathbf{W}_{\mathrm{merge}}^{\mathsf T}
  =\left(\sum_{i=1}^{4}\mathbf{G}'_i\right)^{-1}
  \left(\sum_{i=1}^{4}\mathbf{G}'_i\mathbf{W}_i^{\mathsf T}\right).
\end{equation*}
We use a pseudoinverse if the summed Gram matrix has condition number greater than $10^8$ or any zero diagonal entry; otherwise, we use the ordinary inverse.
This rule applies to two-dimensional linear weights with Gram statistics; all remaining parameters are averaged across the four specialists.

\section{Benchmark and Evaluation Protocols}
\label{sec:appendix-protocol}

Tables~\mainref{tab:unite-leading-benchmarks} and~\mainref{tab:r2r-main} report the $22$ understanding benchmarks and R2R Val-Unseen.
Table~\ref{tab:evaluation-sizes} lists the evaluation split and size of each benchmark; all models in the main tables are evaluated on these data.

\paragraph*{Benchmark and reference-model sources.}
The grounding evaluations include RoboSpatial~\cite{song2025robospatial}, RefSpatial~\cite{zhou2025roborefer}, PointArena~\cite{cheng2025pointarena}, and RoboAfford~\cite{roboafford2025}.
The ShareRobot trajectory task uses the dataset introduced with RoboBrain~\cite{ji2025robobrain}.
The spatial evaluations include VSI-Bench~\cite{yang2025vsibench}, MMSI-Bench~\cite{yang2025mmsibench}, MindCube~\cite{wang2025mindcube}, EmbSpatial-Bench~\cite{du2024embspatial}, SAT (Spatial Aptitude Training)~\cite{ray2025sat}, and the ERQA benchmark introduced with Gemini Robotics~\cite{geminirobotics2025}.
VSI-Bench and MMSI-Bench are abbreviated as VSI and MMSI in the tables.
The 3D scene-understanding tasks use ScanQA~\cite{azuma2022scanqa}, SQA3D~\cite{ma2023sqa3d}, Scan2Cap~\cite{chen2021scan2cap}, ScanRefer~\cite{chen2019scanrefer}, and Multi3DRefer, abbreviated as Multi3DRef in the tables~\cite{zhang2023multi3drefer}.
The real-world and driving evaluations include MME-RealWorld~\cite{zhang2025mmerealworld}, MAPLM~\cite{cao2024maplm}, DriveAction~\cite{driveaction2025}, NuScenes-QA, denoted NuScenesQA in the tables~\cite{qian2024nuscenesqa}, NuPlanQA~\cite{park2025nuplanqa}, and LingoQA~\cite{marcu2024lingoqa}.
Navigation is evaluated on R2R~\cite{anderson2018r2r}.
The closed-source reference models in Table~\mainref{tab:unite-leading-benchmarks} are GPT-5.4~\cite{openai2026gpt54}, Gemini 2.5 Pro~\cite{comanici2025gemini25}, and Claude Sonnet 4.6~\cite{anthropic2026sonnet46}.
The embodied reference models are ACE-Brain-0.5-8B~\cite{gong2026acebrain}, Cosmos3 Nano~\cite{nvidia2026cosmos3}, Embodied-R1.5~\cite{yuan2026embodiedr15}, RynnBrain-8B~\cite{dang2026rynnbrain}, Vlaser-8B~\cite{yang2025vlaser}, MiMo-Embodied-7B~\cite{hao2025mimoembodied}, Pelican-VL-7B~\cite{zhang2025pelicanvl}, VeBrain-7B~\cite{luo2025vebrain}, and RoboBrain 2.5-8B~\cite{tan2026robobrain25}.
The navigation reference models in Table~\mainref{tab:r2r-main} are NaVid~\cite{zhang2024navid}, Uni-NaVid~\cite{zhang2024uninavid}, NaVILA~\cite{cheng2025navila}, StreamVLN~\cite{wei2026streamvln}, NavFoM~\cite{zhang2026navfom}, RynnBrain-Nav~\cite{dang2026rynnbrain}, Qwen-VLA-Instruct~\cite{wang2026qwenvla}, and ACE-Brain-0.5~\cite{gong2026acebrain}.
We evaluate all reference models ourselves on the same benchmark data with the same metrics as UNITE; they differ from UNITE in architecture, scale, and training data and are therefore excluded from the rankings.

\paragraph*{Metric definitions.}
RoboSpatial, RefSpatial, PointArena, and RoboAfford use their task-level aggregate scores, and ShareRobot reports mean normalized discrete Fr\'echet distance (nDFD).
For 3D scene understanding, ScanQA reports CIDEr, SQA3D reports EM-1, Scan2Cap reports CIDEr@0.25, ScanRefer reports Acc@0.25, and Multi3DRef reports F1@0.25.
VSI uses its overall benchmark score; MMSI and DriveAction report overall accuracy across questions, and NuScenesQA reports strict exact-match accuracy.
We also evaluate MindCube, EmbSpatial, ERQA, SAT, MME-RealWorld, MAPLM, NuPlanQA, and LingoQA, each with its official evaluation metric.
R2R reports navigation error in meters and OS, SR, and SPL in percent.
Lower values are better for navigation error and ShareRobot (Traj.), and higher values are better for the other reported scores.

\paragraph*{ShareRobot trajectory metric.}
We report mean normalized discrete Fr\'echet distance (nDFD; lower is better) between image-plane trajectories.
For image width $W$ and height $H$, ground-truth pixels $(u,v)$ are mapped to $(\operatorname{int}(1000u/W),\operatorname{int}(1000v/H))$ by integer truncation.
Predictions use the same coordinate system and are rounded to integers; both sequences are then divided by $1000$ without further normalization or clipping.

Let $P=(p_1,\ldots,p_m)$ and $Q=(q_1,\ldots,q_n)$ denote the normalized prediction and reference.
We linearly resample $P$ along its point-index axis to the reference length $n$, obtaining $\widetilde P$ while keeping $Q$ unchanged.
A single prediction point is repeated; a one-point reference uses the first prediction point.
The dimensionless per-example error is
\[
  d_{\mathrm{DF}}(\widetilde P,Q)
  = \min_{\pi\in\Pi_n}\;\max_{(i,j)\in\pi}
    \|\widetilde p_i-q_j\|_2,
\]
where $\Pi_n$ contains monotone paths from $(1,1)$ to $(n,n)$ with steps $(1,0)$, $(0,1)$, or $(1,1)$.
We report their mean over examples.

\begin{table}[ht]
\centering
\caption{Evaluation settings and sizes for the $22$ understanding benchmarks and R2R.}
\label{tab:evaluation-sizes}
\footnotesize
\setlength{\tabcolsep}{3pt}
\begin{tabular}{@{}>{\raggedright\arraybackslash}p{0.25\linewidth}>{\raggedright\arraybackslash}p{0.29\linewidth}>{\raggedright\arraybackslash}p{0.38\linewidth}@{}}
\toprule
Benchmark & Evaluation setting & Evaluation size \\
\midrule
RoboSpatial & Grounding evaluation & $350$ \\
RefSpatial & Grounding evaluation & $277$ \\
PointArena & Grounding evaluation & $982$ \\
RoboAfford & Grounding evaluation & $338$ \\
ShareRobot & Trajectory evaluation & $200$ \\
ScanQA & Val & $4{,}675$ \\
SQA3D & Val & $3{,}261$ \\
Scan2Cap & Val, Mask3D proposals & $2{,}007$ inference inputs; $2{,}068$ scoring reference objects \\
ScanRefer & Val, Mask3D proposals & $9{,}508$ \\
Multi3DRef & Val, Mask3D proposals & $11{,}120$ \\
VSI & Overall benchmark score & $5{,}130$ \\
MMSI & Overall benchmark accuracy & $1{,}000$ \\
MindCube & Tiny & $1{,}200$ \\
EmbSpatial & Test evaluation & $3{,}640$ \\
ERQA & Test evaluation & $400$ \\
SAT & Val & $150$ \\
MME-RealWorld & Overall benchmark score & $23{,}609$ \\
MAPLM & Multiple-choice evaluation & $1{,}500$ \\
DriveAction & True/false and multiple-choice & $16{,}185$ \\
NuScenesQA & Full evaluation task & $83{,}337$ \\
NuPlanQA & Test & $1{,}801$ \\
LingoQA & Official evaluation/test & $500$ questions; $100$ video clips \\
R2R & Val-Unseen & $1{,}839$ episodes; $11$ scenes \\
\bottomrule
\end{tabular}
\end{table}

\paragraph*{Input processing.}
UNITE evaluations on VSI, MMSI, and MindCube use a minimum image area of $200{,}704$ pixels, a maximum of $802{,}816$ pixels, and at most $32$ video frames.
EmbSpatial and ERQA set both the minimum and maximum image-area budgets to $1{,}048{,}576$ pixels and use the same frame cap.
Their preprocessing preserves aspect ratio and aligns image dimensions to the model's image grid, so the actual area may differ slightly from the target budget due to rounding.
These evaluations use batch size one and the FlashAttention-2 attention implementation.
Grounding inputs use a maximum image area of $16{,}777{,}216$ pixels.
The minimum is $16{,}384$ pixels for RoboSpatial and $1{,}048{,}576$ pixels for the other four grounding tasks.
Model evaluation and router training use Transformers $5.13.0$.

\paragraph*{Generation settings.}
VSI and MME-RealWorld use deterministic decoding with a maximum of $16$ generated tokens.
MMSI and MindCube use deterministic decoding with a maximum of $4{,}096$ tokens.
NuScenesQA and NuPlanQA use a maximum of $512$ tokens.
The EmbSpatial and ERQA task configurations specify at most $1{,}024$ tokens.
The five grounding evaluations use temperature $0.2$, top-$p=0.95$, top-$k=50$, and at most $4{,}096$ generated tokens.
This sampling top-$k$ controls the vocabulary distribution and is distinct from the number of activated MoE experts.

\paragraph*{Reporting conventions.}
Boldface and underlining mark the best and second-best distinct displayed scores within each caption's comparison group.
Ties receive the same formatting, and scores are rounded once to the displayed precision.

\paragraph*{Parameter accounting.}
Active parameters (Table~\ref{tab:supp-storage-params}) are logical counts: the shared non-FFN parameters plus $k$ complete FFN expert branches per token, independently of their compressed storage representation.
The equivalent dense MoE architecture contains approximately $25.1$ billion logical parameters in total.

\FloatBarrier
\supplementprefix{D}
\begin{center}\large\bfseries Detailed Results
\end{center}

\section{Routing Top-\texorpdfstring{$k$}{k}}
\label{sec:appendix-topk}

\begingroup
\setlength{\intextsep}{6pt}
\setlength{\floatsep}{6pt}
Tables~\ref{tab:topk-ablation} and~\ref{tab:topk-r2r-ablation} give the scores underlying Fig.~\mainref{fig:routing-curves}: top-$k\in\{1,2,3\}$ before and after router adaptation, on all $22$ benchmarks and R2R Val-Unseen.
Source specialists and consolidation stages are compared in Tables~\mainref{tab:setup-models} and~\mainref{tab:setup-r2r} of the main manuscript.
\begin{table}[H]
  \centering
  \caption{Top-$k$ comparison at two stages of UNITE. Active parameters are logical counts. Higher is better except ShareRobot (Traj.). MMSI and DriveAction report overall accuracy. Routing settings are ranked separately within each stage.}
  \label{tab:topk-ablation}
  \fontsize{8.2}{10.2}\selectfont
  \setlength{\tabcolsep}{1.0pt}
  \def\unitegroupgap{4.8pt}
  \def\unitetintL{\dimexpr\tabcolsep+0.3pt\relax}
  \def\unitegroupgaptrim{\dimexpr\unitegroupgap-\tabcolsep\relax}
  \providecolor{tintGrounding}{HTML}{C9BFE8}
  \providecolor{tintSpatial}{HTML}{A9C9F2}
  \providecolor{tintDriving}{HTML}{F3BCBB}
  \renewcommand{\arraystretch}{1.15}
  \providecommand{\unitebenchmarkhead}[1]{\rotatebox{90}{\strut #1}}
  \begin{adjustbox}{max width=\textwidth,center}
    \begin{tabular}{@{}lc@{\hspace{\unitegroupgap}}*{5}{>{\columncolor{tintGrounding!30}[\unitetintL][\tabcolsep]}c}@{\hspace{\unitegroupgap}}*{11}{>{\columncolor{tintSpatial!30}[\unitetintL][\tabcolsep]}c}@{\hspace{\unitegroupgap}}*{6}{>{\columncolor{tintDriving!30}[\unitetintL][\tabcolsep]}c}@{\hspace{\tabcolsep}}}
      \toprule
                                                           &                                            & \multicolumn{5}{@{}>{\columncolor{tintGrounding!30}[\unitetintL][\tabcolsep]}c@{\hspace{\unitegroupgap}}}{\textbf{Grounding}} & \multicolumn{11}{@{}>{\columncolor{tintSpatial!30}[\unitetintL][\tabcolsep]}c@{\hspace{\unitegroupgap}}}{\textbf{Embodied Spatial}} & \multicolumn{6}{@{}>{\columncolor{tintDriving!30}[\unitetintL][\tabcolsep]}c@{\hspace{\tabcolsep}}}{\textbf{Driving}} \\
      \cmidrule(l{-\unitetintL}r{\unitegroupgaptrim}){3-7}\cmidrule(l{-\unitetintL}r{\unitegroupgaptrim}){8-18}\cmidrule(l{-\unitetintL}r{0pt}){19-24}
      \multicolumn{1}{c}{\raisebox{3.2em}[0pt][0pt]{\textbf{Model}}} & \unitebenchmarkhead{Active (B)} & \unitebenchmarkhead{RoboSpatial}       & \unitebenchmarkhead{RefSpatial}                & \unitebenchmarkhead{PointArena} & \unitebenchmarkhead{RoboAfford} & \unitebenchmarkhead{ShareRobot ($\downarrow$)} & \unitebenchmarkhead{VSI} & \unitebenchmarkhead{MMSI} & \unitebenchmarkhead{MindCube} & \unitebenchmarkhead{EmbSpatial} & \unitebenchmarkhead{ERQA} & \unitebenchmarkhead{ScanQA} & \unitebenchmarkhead{SQA3D} & \unitebenchmarkhead{Scan2Cap} & \unitebenchmarkhead{ScanRefer} & \unitebenchmarkhead{Multi3DRef} & \unitebenchmarkhead{SAT} & \unitebenchmarkhead{MME-RealWorld} & \unitebenchmarkhead{MAPLM} & \unitebenchmarkhead{DriveAction} & \unitebenchmarkhead{NuScenesQA} & \unitebenchmarkhead{NuPlanQA} & \unitebenchmarkhead{LingoQA} \\
      \midrule
      \multicolumn{24}{l}{\textit{Shared-anchor one-bit MoE}} \\
      UNITE (top-$k=1$)                                    & 8.8                                        & 58.9                                   & \underline{59.4}                               & \underline{71.7}                & \underline{79.6}                & \textbf{0.26}                               & \underline{47.4}         & \textbf{37.5}             & \underline{88.1}              & \textbf{80.1}                   & \underline{47.0}          & \textbf{106.6}              & \underline{63.1}           & \textbf{100.6}                & 71.3                           & \underline{73.6}                & \underline{87.3}         & \textbf{65.9}                      & 92.0                       & 79.5                             & 29.7                            & 89.8                         & 59.2 \\
      UNITE (top-$k=2$)                                    & 14.2                                       & \textbf{60.5}                          & \underline{59.4}                               & 71.6                            & 79.4                            & \textbf{0.26}                               & \textbf{47.5}            & \underline{37.3}          & 88.0                          & \underline{79.8}                & \underline{47.0}          & \underline{106.3}           & \textbf{63.3}              & 97.6                          & \underline{71.5}               & \underline{73.6}                & \underline{87.3}         & 65.7                               & \underline{92.9}           & \underline{79.7}                 & \underline{31.5}                & \underline{90.0}              & \underline{60.4} \\
      UNITE (top-$k=3$)                                    & 19.6                                       & \underline{59.7}                       & \textbf{61.2}                                  & \textbf{71.9}                   & \textbf{80.9}                   & \textbf{0.26}                               & 47.3                     & \textbf{37.5}             & \textbf{88.3}                 & 79.5                            & \textbf{47.5}             & 106.2                       & \underline{63.1}           & \underline{98.3}              & \textbf{71.7}                  & \textbf{73.7}                   & \textbf{88.0}            & \underline{65.8}                   & \textbf{93.2}              & \textbf{80.0}                    & \textbf{33.2}                   & \textbf{90.3}                 & \textbf{60.8} \\
      \midrule
      \multicolumn{24}{l}{\textit{One-bit MoE + router adaptation}} \\
      UNITE (top-$k=1$)                                    & 8.8                                        & \underline{58.3}                       & 57.2                                           & \textbf{71.3}                   & \underline{80.3}                & \textbf{0.25}                               & \underline{46.7}         & 36.3                      & \textbf{87.9}                 & 78.9                            & \textbf{47.3}             & \underline{105.6}           & \underline{62.8}           & 102.1                         & \underline{71.0}               & 73.2                            & \textbf{90.7}            & \underline{65.7}                   & \textbf{93.8}              & \underline{79.8}                 & \textbf{48.7}                   & \textbf{91.2}                 & \textbf{63.0} \\
      UNITE (top-$k=2$)                                    & 14.2                                       & 57.3                                   & \textbf{58.8}                                  & \underline{71.0}                & 80.2                            & \underline{0.26}                            & \textbf{47.1}            & \underline{36.4}          & \underline{87.8}              & \underline{79.0}                & 46.5                      & \underline{105.6}           & \textbf{63.2}              & \textbf{102.8}                & \underline{71.0}               & \textbf{73.4}                   & \underline{89.3}         & 65.6                               & \underline{93.7}           & \underline{79.8}                 & \underline{48.5}                & \underline{91.1}              & \underline{62.4} \\
      UNITE (top-$k=3$)                                    & 19.6                                       & \textbf{58.4}                          & \underline{58.5}                               & 70.9                            & \textbf{80.9}                   & \textbf{0.25}                               & \textbf{47.1}            & \textbf{36.8}             & \underline{87.8}              & \textbf{79.1}                   & \underline{46.8}          & \textbf{106.0}              & \textbf{63.2}              & \underline{102.6}             & \textbf{71.1}                  & \underline{73.3}                & \underline{89.3}         & \textbf{65.8}                      & \underline{93.7}           & \textbf{79.9}                    & 48.4                            & \underline{91.1}              & \underline{62.4} \\
      \bottomrule
    \end{tabular}
  \end{adjustbox}
\end{table}
\noindent\begin{minipage}{\linewidth}
\noindent
Before router adaptation, NuScenesQA rises from $29.7$ ($k=1$) to $33.2$ ($k=3$) and R2R SR from $57.0\%$ to $58.2\%$.
After adaptation, top-$1$ reaches $48.7$ on NuScenesQA and $59.6\%$ R2R SR, and larger $k$ changes NuScenesQA by at most $0.3$ points.

\begin{table}[H]
  \centering
  \caption{R2R Val-Unseen, same episodes across settings. Best and second-best scores are marked within each stage. NE: meters; OS, SR, SPL: percentages.}
  \label{tab:topk-r2r-ablation}
  \footnotesize
  \setlength{\tabcolsep}{4pt}
  \renewcommand{\arraystretch}{1.08}
  \begin{adjustbox}{max width=\linewidth,center}
  \begin{tabular}{@{}lcccc@{}}
    \toprule
    \textbf{Routing} & NE$\downarrow$ & OS$\uparrow$ & SR$\uparrow$ & SPL$\uparrow$ \\
    \midrule
    \multicolumn{5}{l}{\textit{Shared-anchor one-bit MoE}} \\
    Top-$1$                            & 4.9                & 62.1              & \underline{57.0}  & 51.9 \\
    Top-$2$                            & \underline{4.8}    & \textbf{63.7}     & \textbf{58.2}     & \textbf{53.0} \\
    Top-$3$                            & \textbf{4.7}       & \underline{63.6}  & \textbf{58.2}     & \underline{52.9} \\
    \midrule
    \multicolumn{5}{l}{\textit{One-bit MoE + router adaptation}} \\
    Top-$1$                            & \textbf{4.5}       & 65.7              & \underline{59.6}  & \underline{53.4} \\
    Top-$2$                            & \textbf{4.5}       & \underline{65.8}  & 59.4              & 53.2 \\
    Top-$3$                            & \underline{4.6}    & \textbf{66.5}     & \textbf{60.1}     & \textbf{53.7} \\
    \bottomrule
  \end{tabular}
  \end{adjustbox}
\end{table}
\par
\end{minipage}
\par
\FloatBarrier

\section{VSI Frame-Count Sweep by Sub-Task}
\label{sec:appendix-vsi-frames}

Table~\ref{tab:vsi-frames} lists the overall VSI scores plotted in Fig.~\mainref{fig:vsi-frames} together with the eight sub-task scores.
Both expert banks share the backbone and use top-$1$ routing without router adaptation on all $5{,}130$ questions; only the maximum number of sampled frames varies.
On object counting and object size, scores change by at most $5$ points across frame counts for both models.
The uncompressed MoE gains most on sub-tasks that relate observations across frames, such as relative direction ($49.4\rightarrow80.3$) and appearance order ($51.0\rightarrow77.8$), whereas the one-bit MoE varies by less than six points on each of them.

\begin{table}[!ht]
\centering
\caption{VSI-Bench overall and sub-task scores versus the maximum number of sampled frames (all $5{,}130$ questions; top-$1$ routing, no router adaptation). The main tables use $32$ frames. Abs.\ dist.: absolute distance; Rel.\ dist./dir.: relative distance/direction; Appear.\ order: appearance order.}
\label{tab:vsi-frames}
\footnotesize
\setlength{\tabcolsep}{4pt}
\renewcommand{\arraystretch}{1.08}
\newcommand{\vsihead}[2]{\begin{tabular}[b]{@{}c@{}}#1\\#2\end{tabular}}
\begin{adjustbox}{max width=\textwidth,center}
\begin{tabular}{@{}llccccccccc@{}}
\toprule
\textbf{Expert bank} & \textbf{Frames} & \textbf{Overall} & \vsihead{Obj.}{count} & \vsihead{Abs.}{dist.} & \vsihead{Obj.}{size} & \vsihead{Room}{size} & \vsihead{Rel.}{dist.} & \vsihead{Rel.}{dir.} & \vsihead{Route}{plan} & \vsihead{Appear.}{order} \\
\midrule
\multirow{5}{*}{Uncompressed MoE}
 & 4  & 48.1 & 64.9 & 28.3 & 68.9 & 45.6 & 42.5 & 49.4 & 34.0 & 51.0 \\
 & 8  & 55.2 & 67.4 & 35.7 & 71.9 & 59.4 & 47.5 & 58.0 & 38.1 & 63.8 \\
 & 16 & 59.2 & 68.8 & 43.1 & 71.8 & 60.6 & 54.6 & 59.3 & 43.3 & 72.2 \\
 & 32 & 62.4 & 66.6 & 46.4 & 73.9 & 61.1 & 61.5 & 69.8 & 46.4 & 73.3 \\
 & 64 & 65.0 & 66.1 & 49.1 & 73.8 & 62.2 & 63.0 & 80.3 & 47.9 & 77.8 \\
      \midrule
      \multirow{5}{*}{One-bit MoE}
 & 4  & 46.9 & 63.9 & 26.8 & 68.1 & 37.0 & 42.4 & 48.1 & 38.1 & 50.5 \\
 & 8  & 48.8 & 64.0 & 27.7 & 69.1 & 36.7 & 48.7 & 47.9 & 40.2 & 55.8 \\
 & 16 & 46.8 & 63.9 & 26.9 & 68.3 & 36.7 & 42.4 & 47.9 & 38.1 & 50.3 \\
 & 32 & 47.4 & 63.0 & 26.7 & 67.8 & 36.8 & 45.2 & 48.0 & 39.2 & 52.9 \\
 & 64 & 47.8 & 64.5 & 29.8 & 68.9 & 40.1 & 44.8 & 45.3 & 33.5 & 55.5 \\
      \bottomrule
    \end{tabular}
  \end{adjustbox}
\end{table}
\endgroup

\FloatBarrier
\supplementprefix{Q}
\section{Qualitative Results}
\label{sec:qualitative-results}

All outputs are produced by the final UNITE model (One-bit MoE + router adaptation) with top-$1$ routing and are shown alongside the reference answers or task-level scores.

\subsection{Grounding}
\label{sec:qualitative-grounding}

\begin{figure}[!htbp]
    \centering
    \begin{adjustbox}{max totalsize={\textwidth}{0.82\textheight},center}
    \begin{minipage}{\textwidth}
    \includegraphics[width=\textwidth]{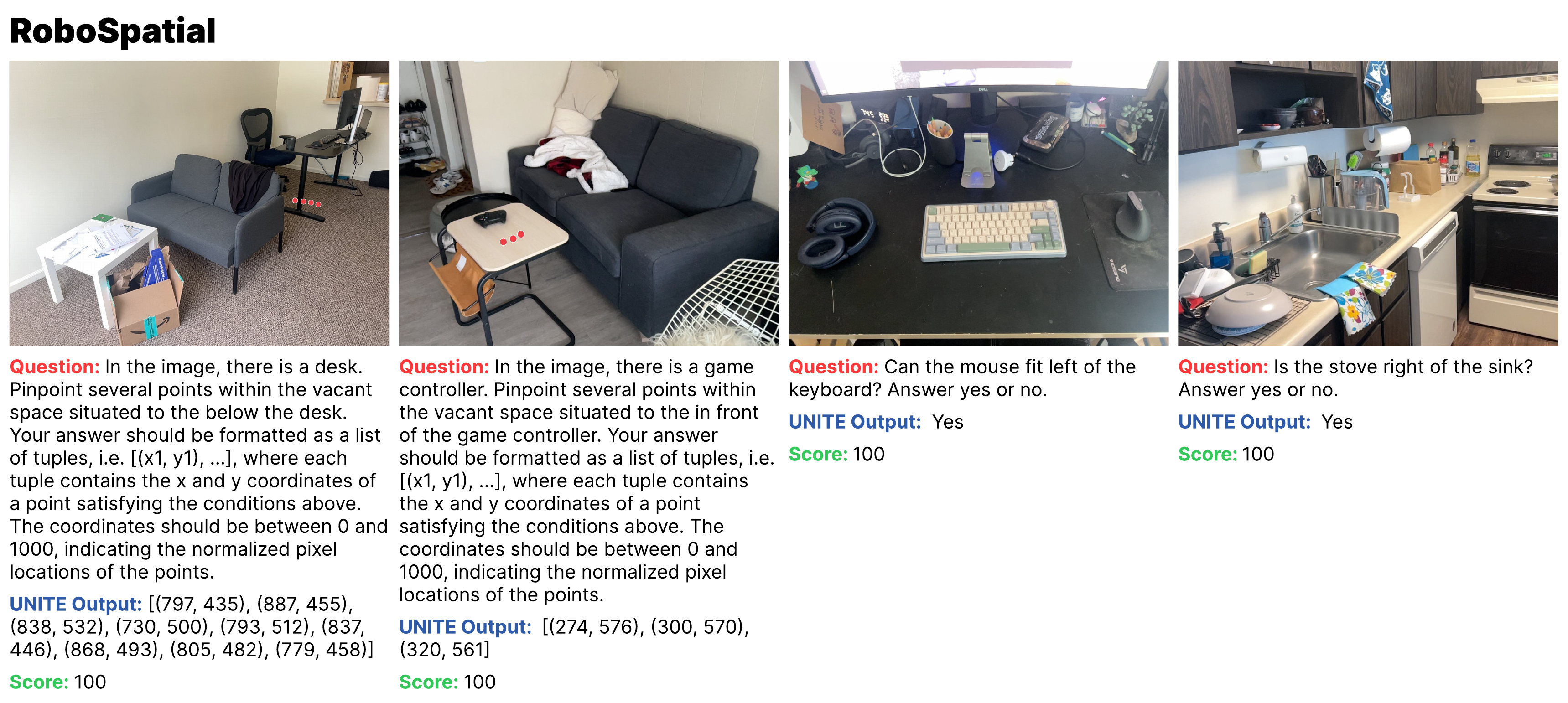}
    \end{minipage}
    \end{adjustbox}
    \caption{Qualitative examples on RoboSpatial. The bracket mismatch in the displayed output is retained from the original model response.}
    \label{fig:qual-robospatial}
\end{figure}

\begin{figure}[!htbp]
    \centering
    \begin{adjustbox}{max totalsize={\textwidth}{0.82\textheight},center}
    \begin{minipage}{\textwidth}
    \includegraphics[width=\textwidth]{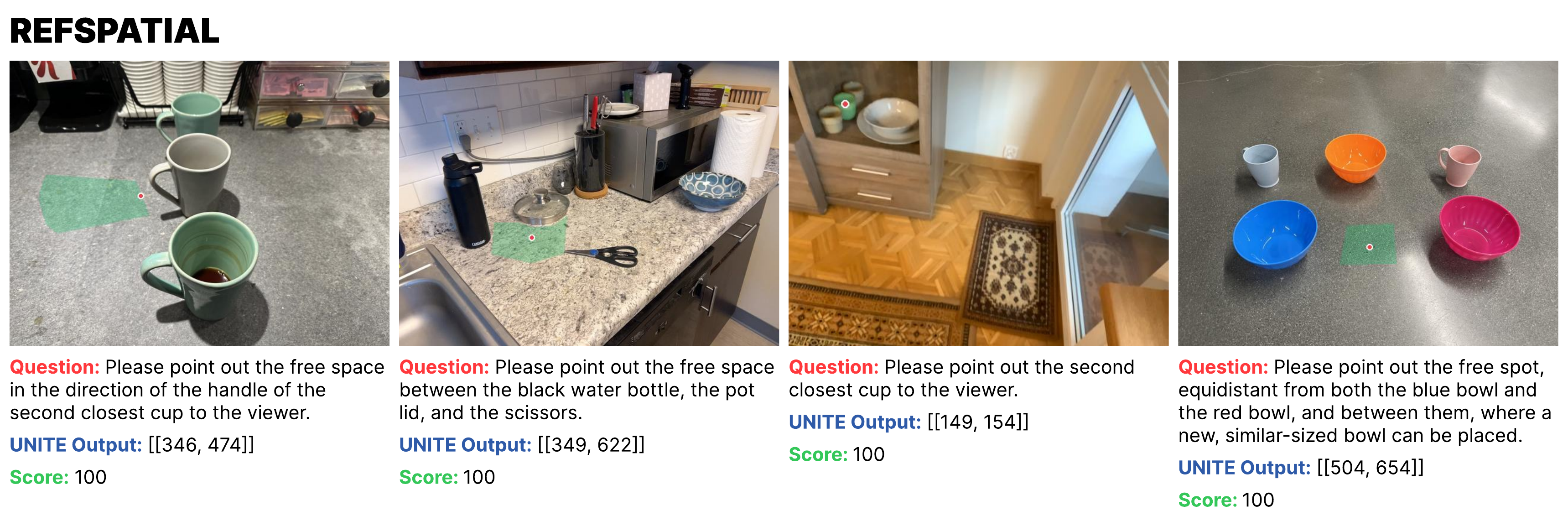}
    \end{minipage}
    \end{adjustbox}
    \caption{Qualitative examples on RefSpatial.}
    \label{fig:qual-refspatial}
\end{figure}

\begin{figure}[!htbp]
    \centering
    \begin{adjustbox}{max totalsize={\textwidth}{0.82\textheight},center}
    \begin{minipage}{\textwidth}
    \includegraphics[width=\textwidth]{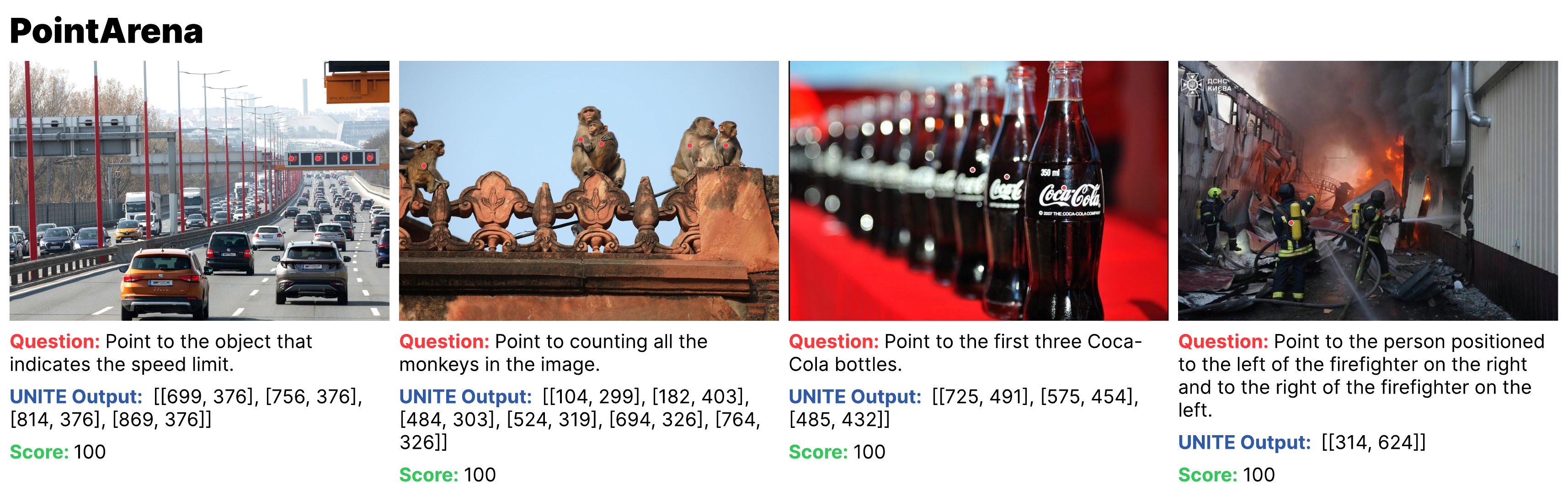}
    \end{minipage}
    \end{adjustbox}
    \caption{Qualitative examples on PointArena.}
    \label{fig:qual-pointarena}
\end{figure}

\begin{figure}[!htbp]
    \centering
    \begin{adjustbox}{max totalsize={\textwidth}{0.82\textheight},center}
    \begin{minipage}{\textwidth}
    \includegraphics[width=\textwidth]{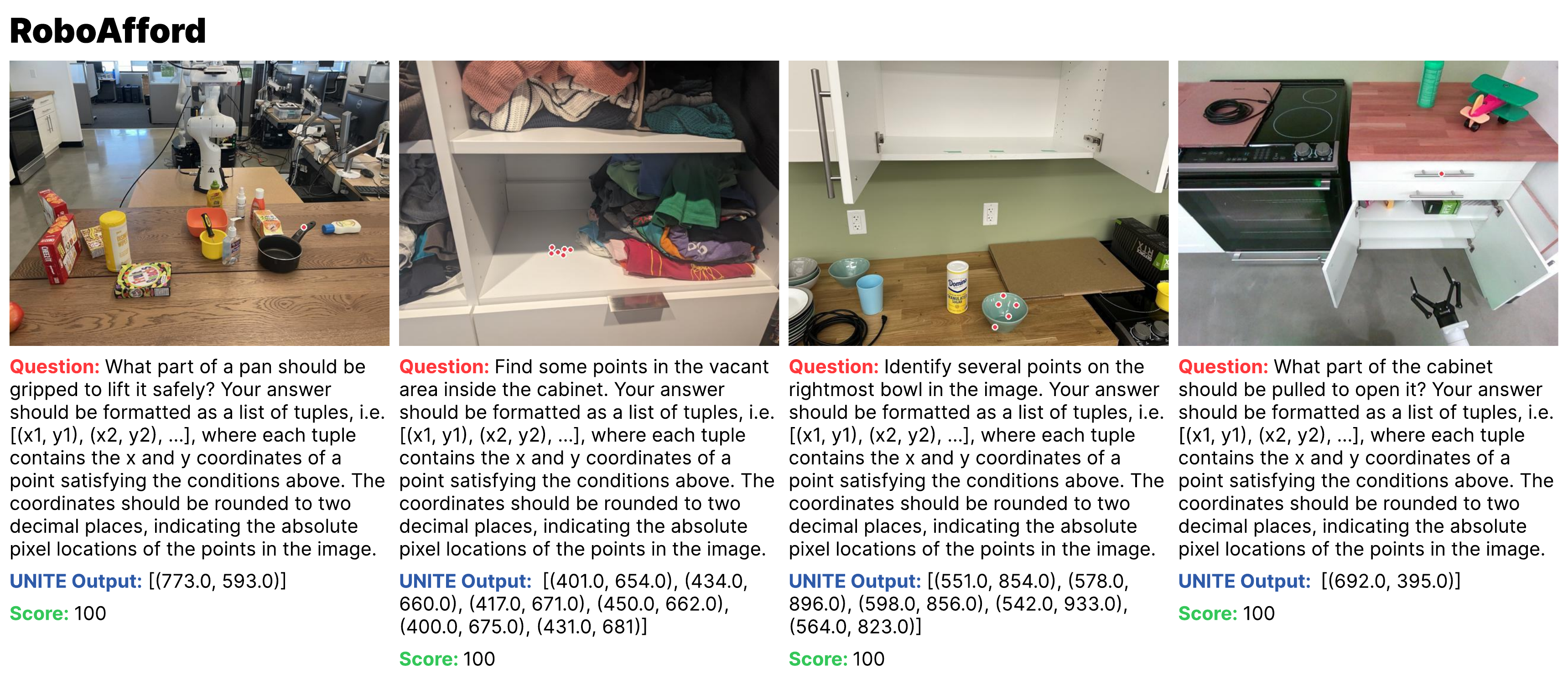}
    \end{minipage}
    \end{adjustbox}
    \caption{Qualitative examples on RoboAfford.}
    \label{fig:qual-roboafford}
\end{figure}

\begin{figure}[!htbp]
    \centering
    \begin{adjustbox}{max totalsize={\textwidth}{0.82\textheight},center}
    \begin{minipage}{\textwidth}
    \includegraphics[width=\textwidth]{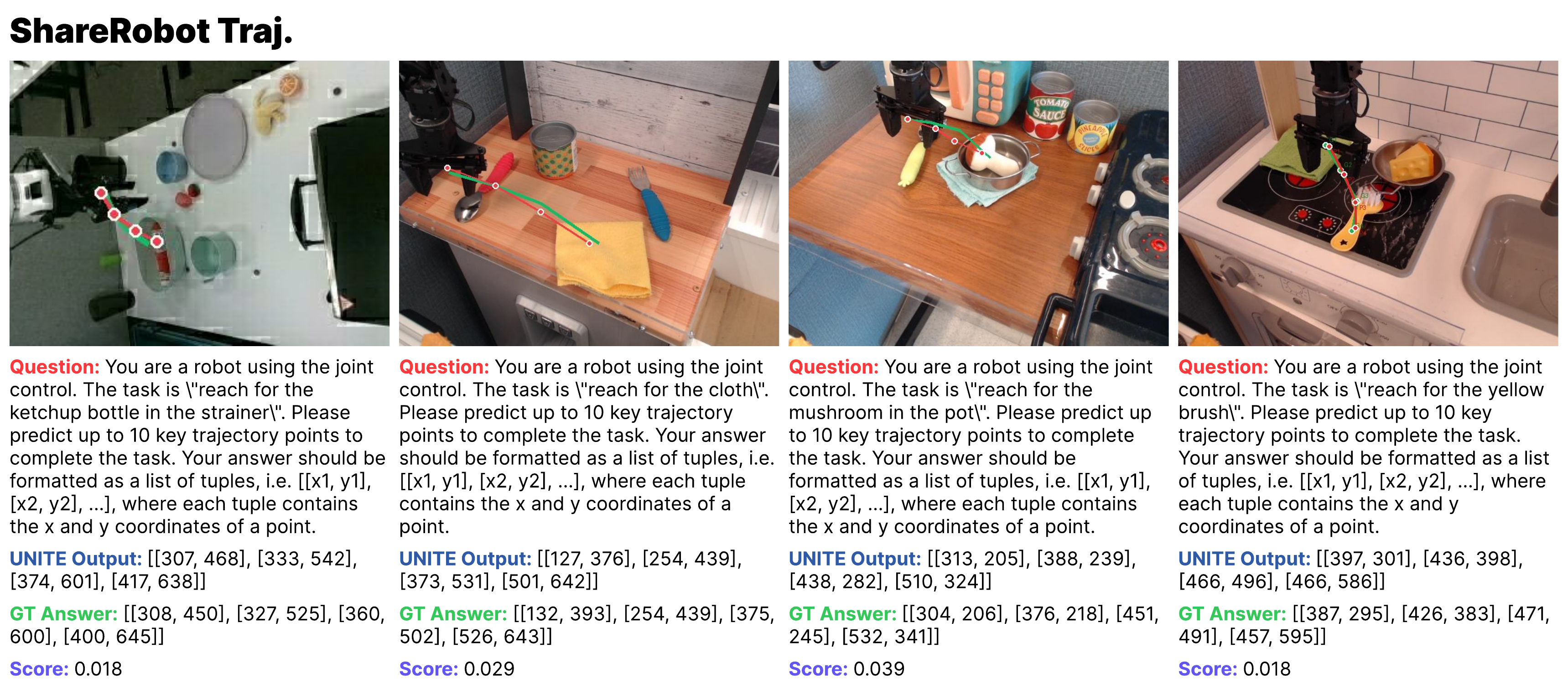}
    \end{minipage}
    \end{adjustbox}
    \caption{Qualitative examples on ShareRobot (Traj.).}
    \label{fig:qual-sharerobot}
\end{figure}

We show examples on RoboSpatial (Fig.~\ref{fig:qual-robospatial}), RefSpatial (Fig.~\ref{fig:qual-refspatial}), PointArena (Fig.~\ref{fig:qual-pointarena}), RoboAfford (Fig.~\ref{fig:qual-roboafford}), and ShareRobot trajectory prediction (Fig.~\ref{fig:qual-sharerobot}).

\FloatBarrier
\subsection{Embodied Spatial}
\label{sec:qualitative-spatial}

\begin{figure}[!htbp]
    \centering
    \begin{adjustbox}{max totalsize={\textwidth}{0.82\textheight},center}
    \begin{minipage}{\textwidth}
    \includegraphics[width=\textwidth]{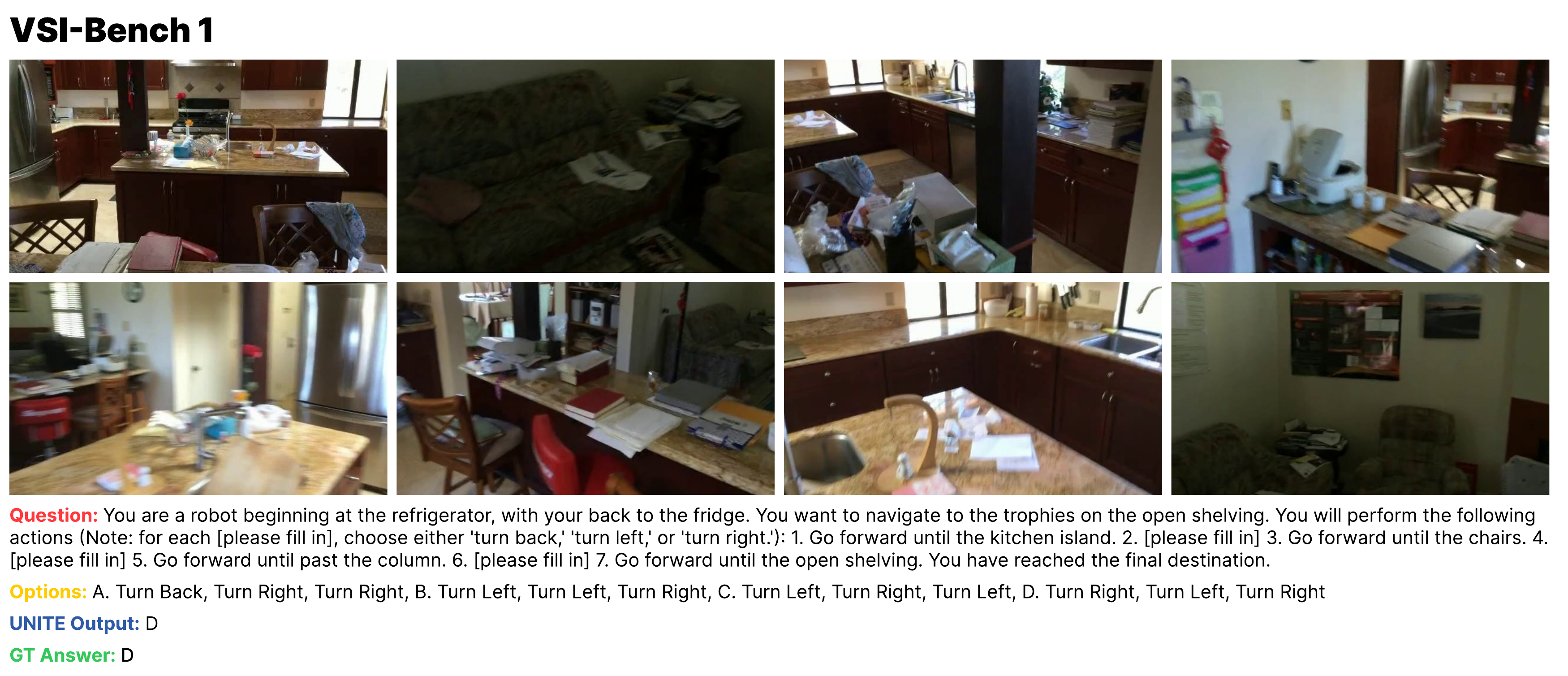}
    \par\smallskip
    \includegraphics[width=\textwidth]{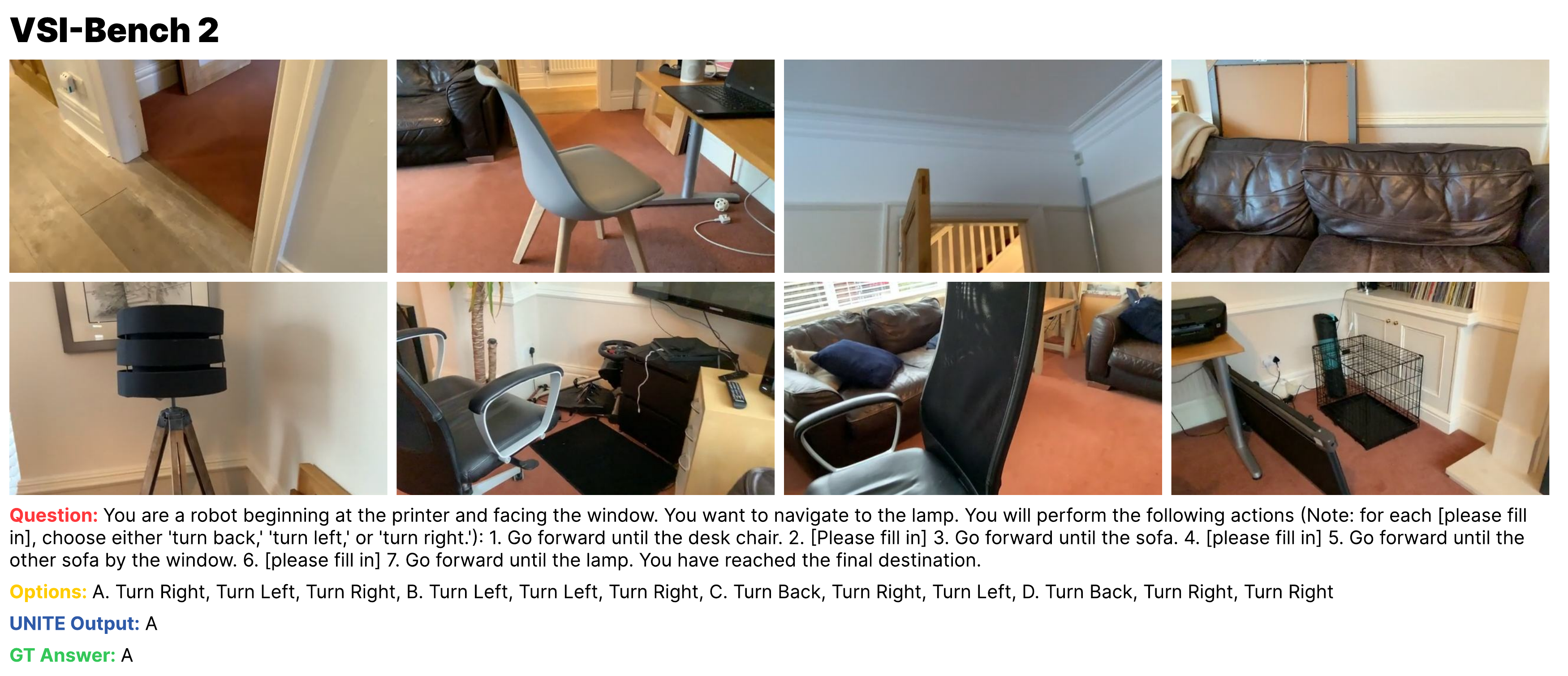}
    \end{minipage}
    \end{adjustbox}
    \caption{Qualitative examples on VSI.}
    \label{fig:qual-vsi}
\end{figure}

\begin{figure}[!htbp]
    \centering
    \begin{adjustbox}{max totalsize={\textwidth}{0.82\textheight},center}
    \begin{minipage}{\textwidth}
    \includegraphics[width=\textwidth]{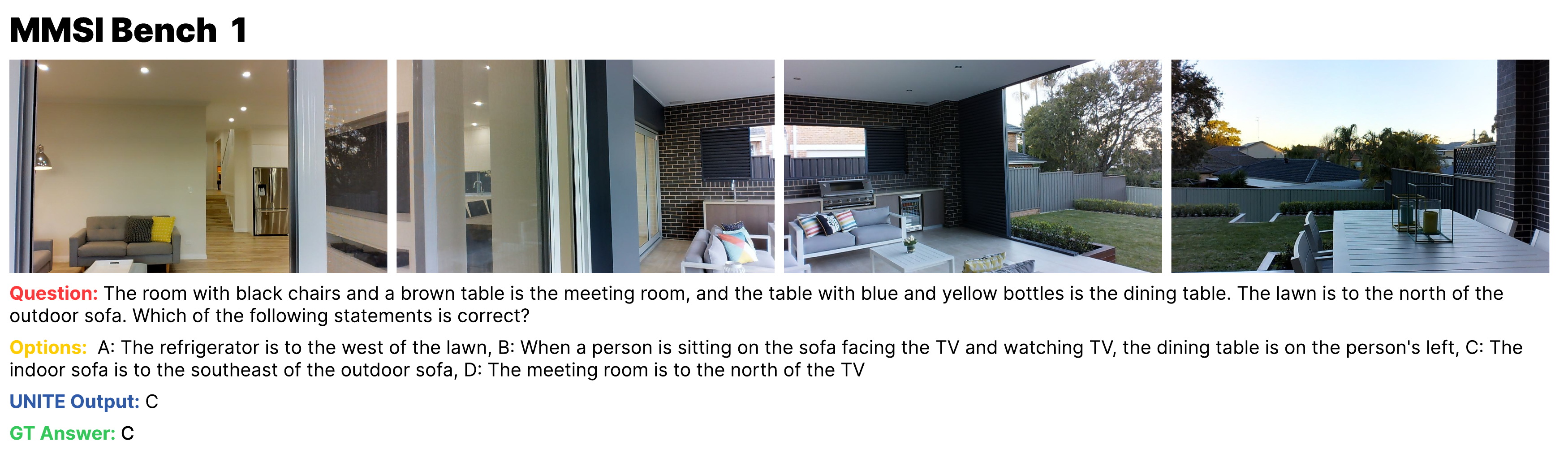}
    \par\smallskip
    \includegraphics[width=\textwidth]{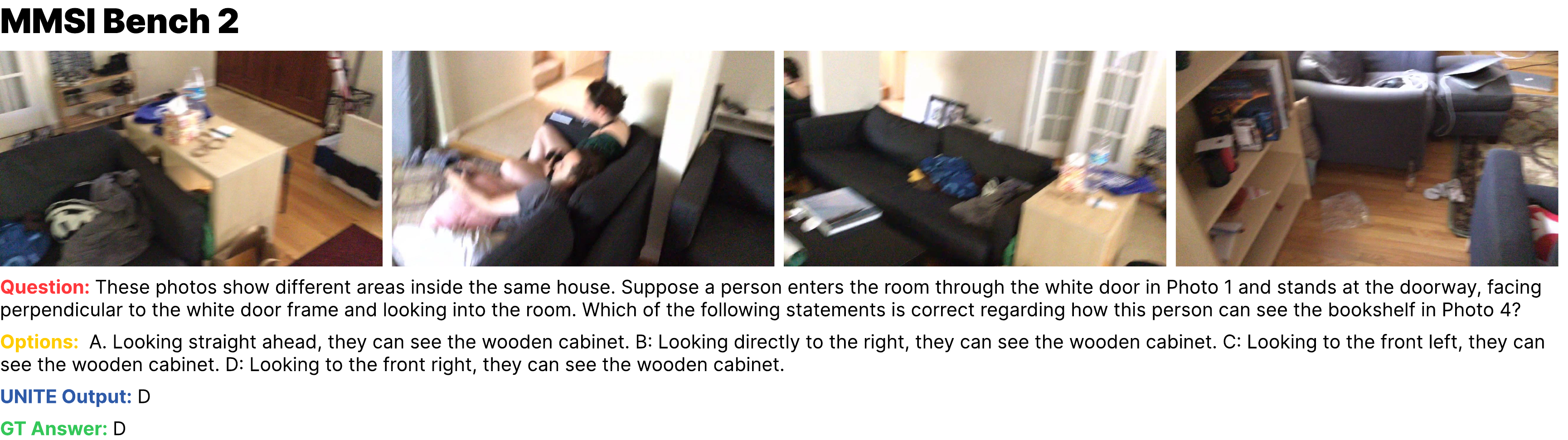}
    \end{minipage}
    \end{adjustbox}
    \caption{Qualitative examples on MMSI.}
    \label{fig:qual-mmsi}
\end{figure}

\begin{figure}[!htbp]
    \centering
    \begin{adjustbox}{max totalsize={\textwidth}{0.82\textheight},center}
    \begin{minipage}{\textwidth}
    \includegraphics[width=\textwidth]{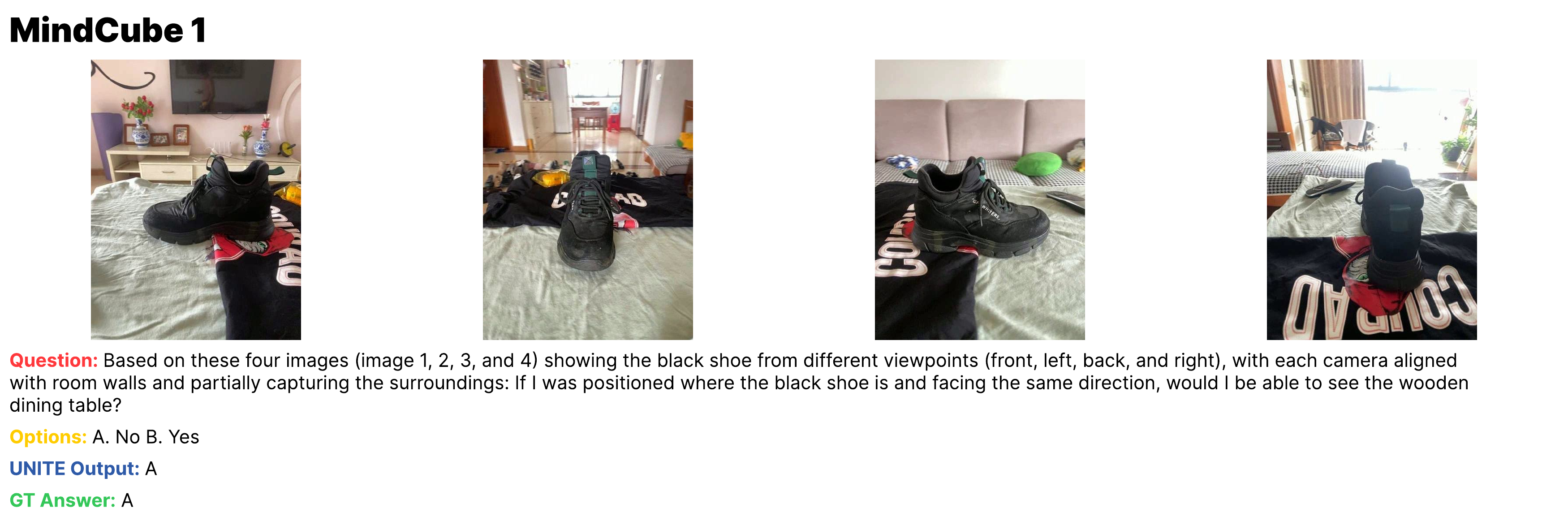}
    \par\smallskip
    \includegraphics[width=\textwidth]{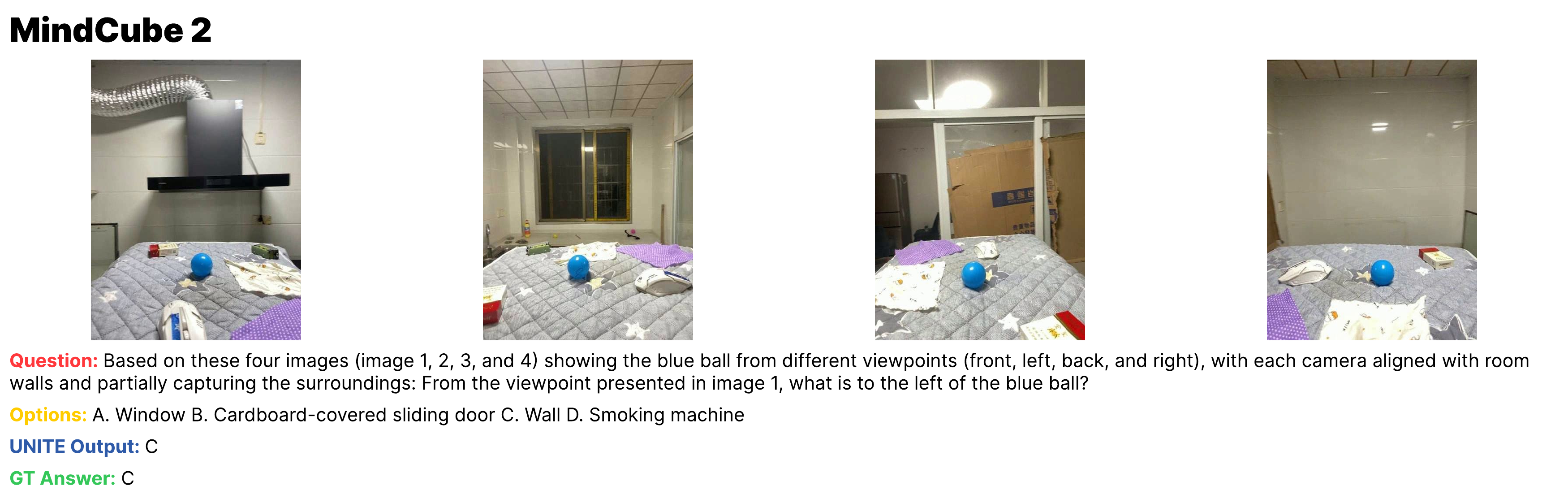}
    \end{minipage}
    \end{adjustbox}
    \caption{Qualitative examples on MindCube.}
    \label{fig:qual-mindcube}
\end{figure}

\begin{figure}[!htbp]
    \centering
    \begin{adjustbox}{max totalsize={\textwidth}{0.82\textheight},center}
    \begin{minipage}{\textwidth}
    \includegraphics[width=\textwidth]{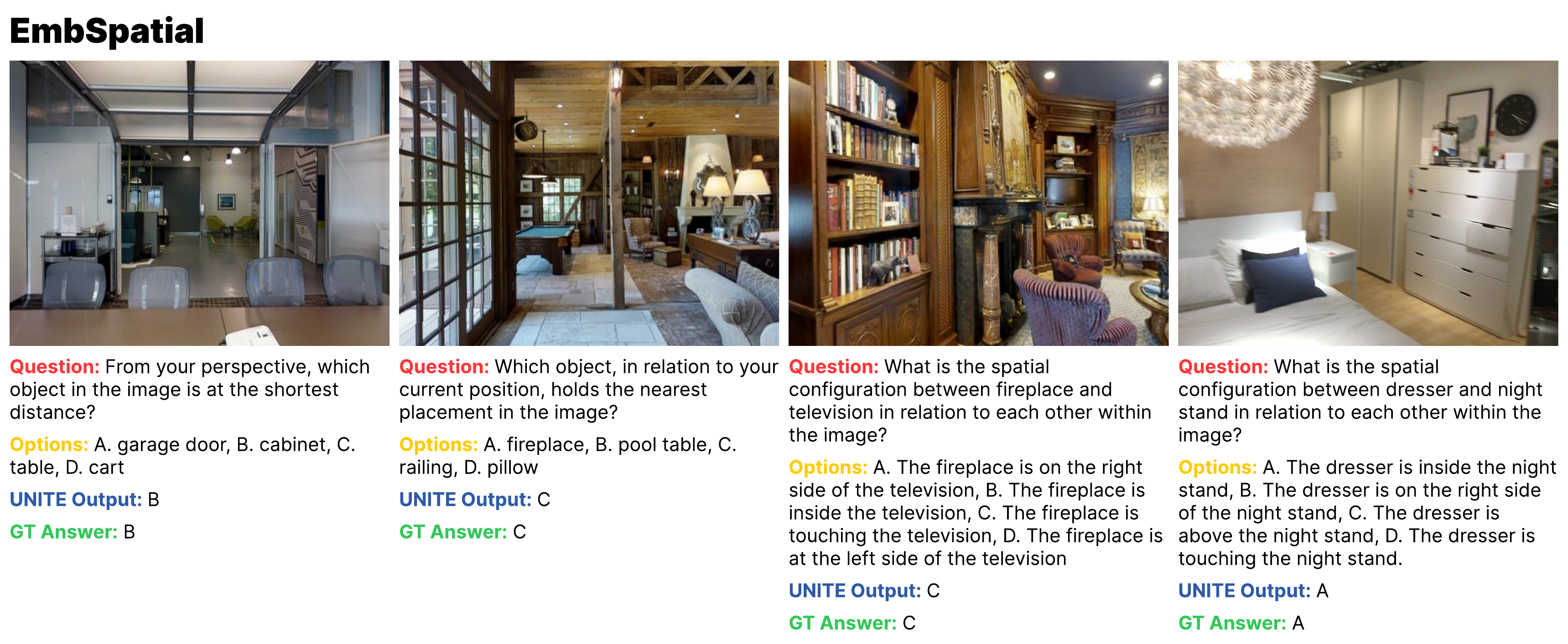}
    \end{minipage}
    \end{adjustbox}
    \caption{Qualitative examples on EmbSpatial.}
    \label{fig:qual-embspatial}
\end{figure}

\begin{figure}[!htbp]
    \centering
    \begin{adjustbox}{max totalsize={\textwidth}{0.82\textheight},center}
    \begin{minipage}{\textwidth}
    \includegraphics[width=\textwidth]{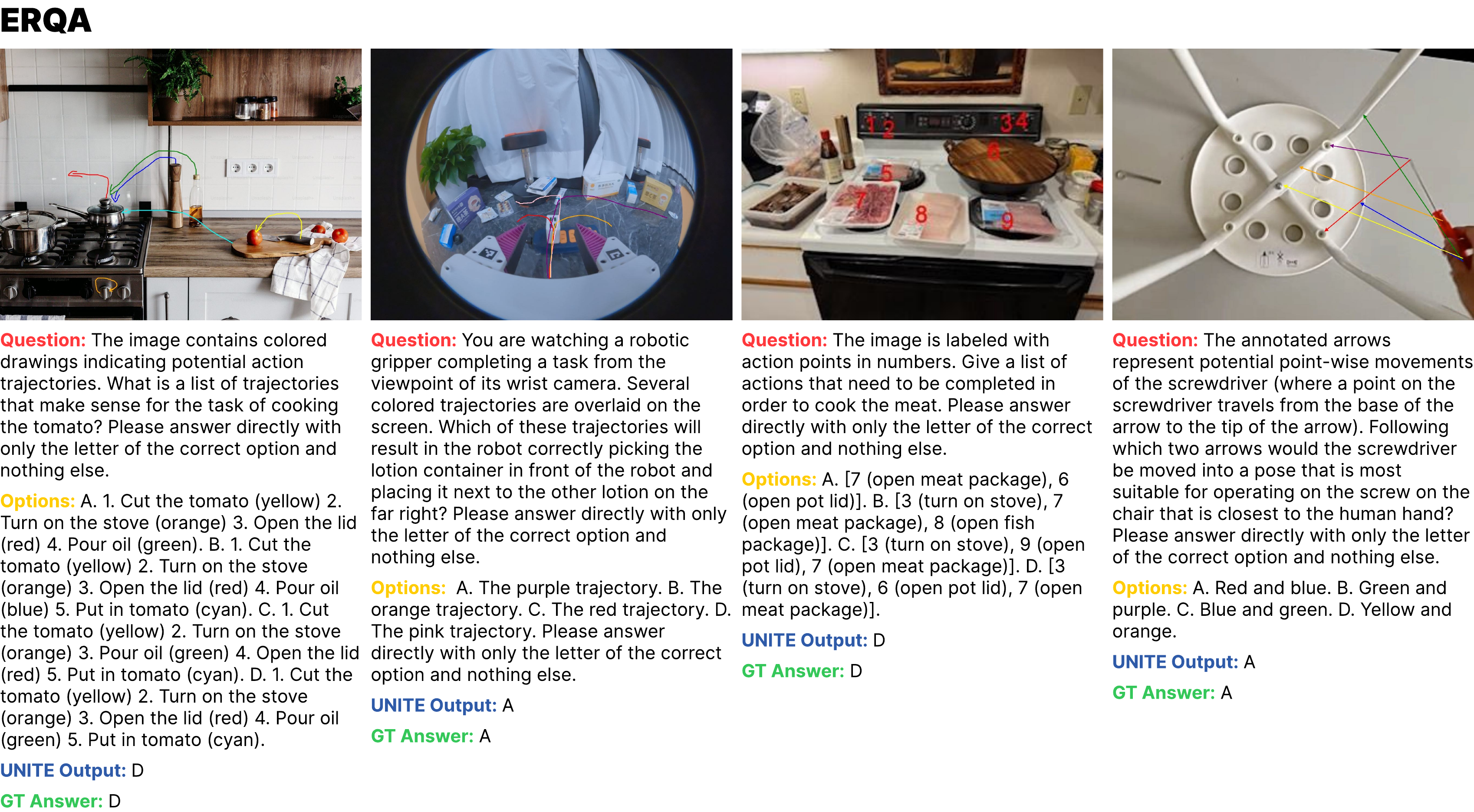}
    \end{minipage}
    \end{adjustbox}
    \caption{Qualitative examples on ERQA.}
    \label{fig:qual-erqa}
\end{figure}

\begin{figure}[!htbp]
    \centering
    \begin{adjustbox}{max totalsize={\textwidth}{0.82\textheight},center}
    \begin{minipage}{\textwidth}
    \includegraphics[width=\textwidth]{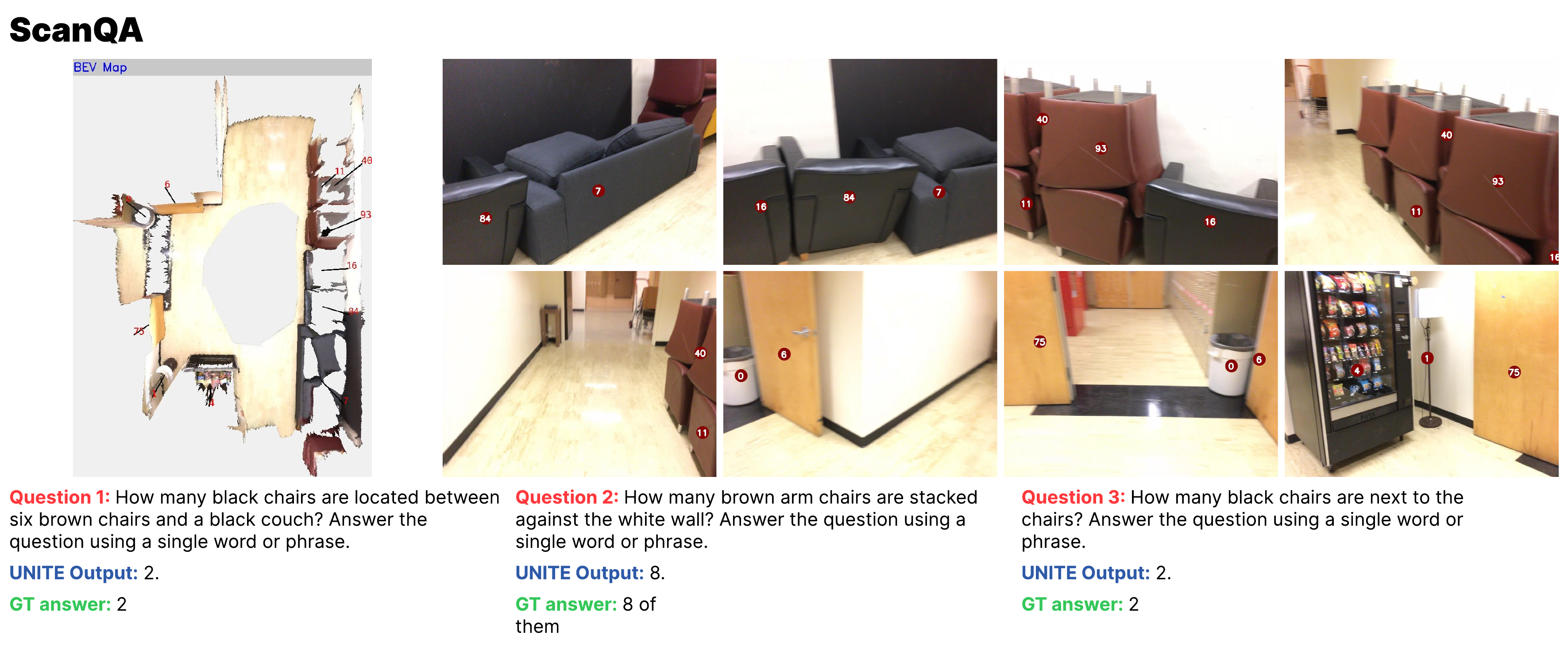}
    \end{minipage}
    \end{adjustbox}
    \caption{Qualitative examples on ScanQA.}
    \label{fig:qual-scanqa}
\end{figure}

\begin{figure}[!htbp]
    \centering
    \begin{adjustbox}{max totalsize={\textwidth}{0.82\textheight},center}
    \begin{minipage}{\textwidth}
    \includegraphics[width=\textwidth]{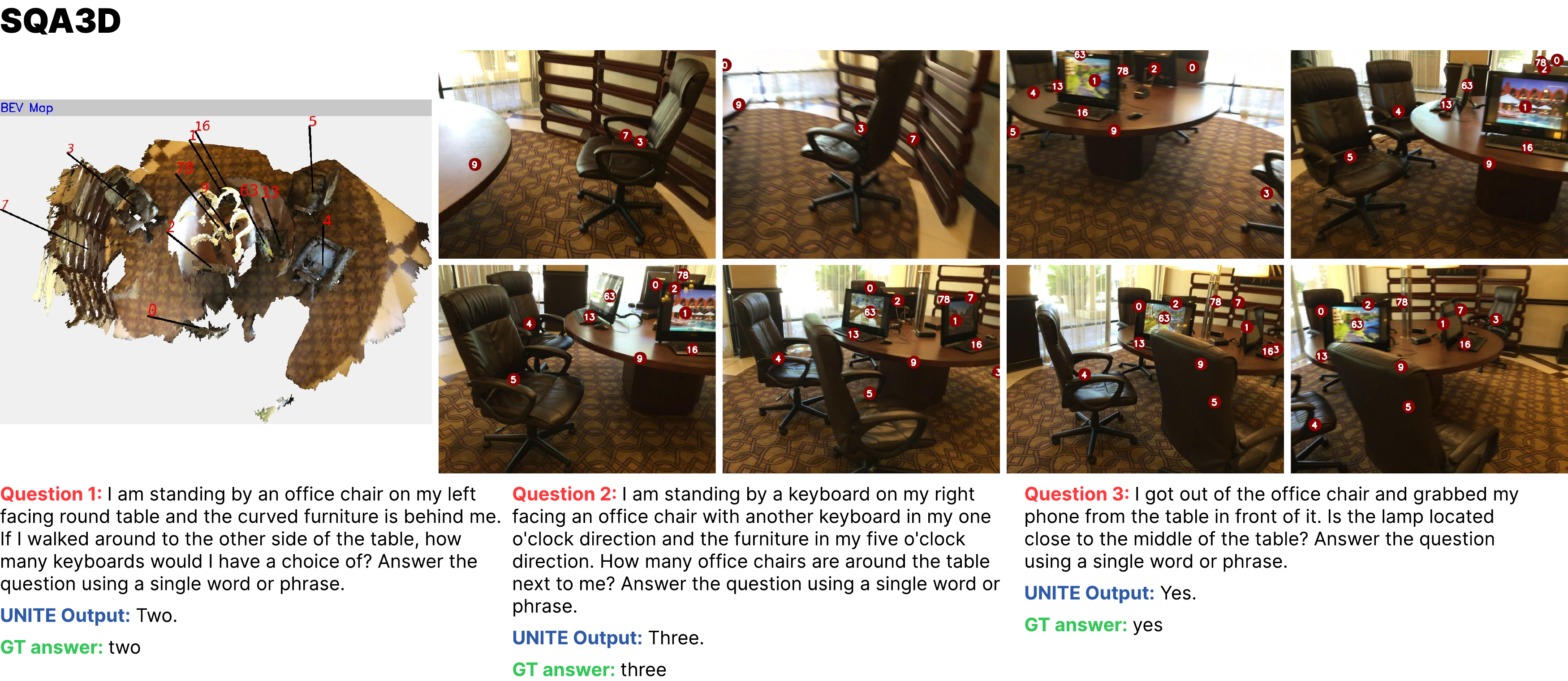}
    \end{minipage}
    \end{adjustbox}
    \caption{Qualitative examples on SQA3D.}
    \label{fig:qual-sqa3d}
\end{figure}

\begin{figure}[!htbp]
    \centering
    \begin{adjustbox}{max totalsize={\textwidth}{0.82\textheight},center}
    \begin{minipage}{\textwidth}
    \includegraphics[width=\textwidth]{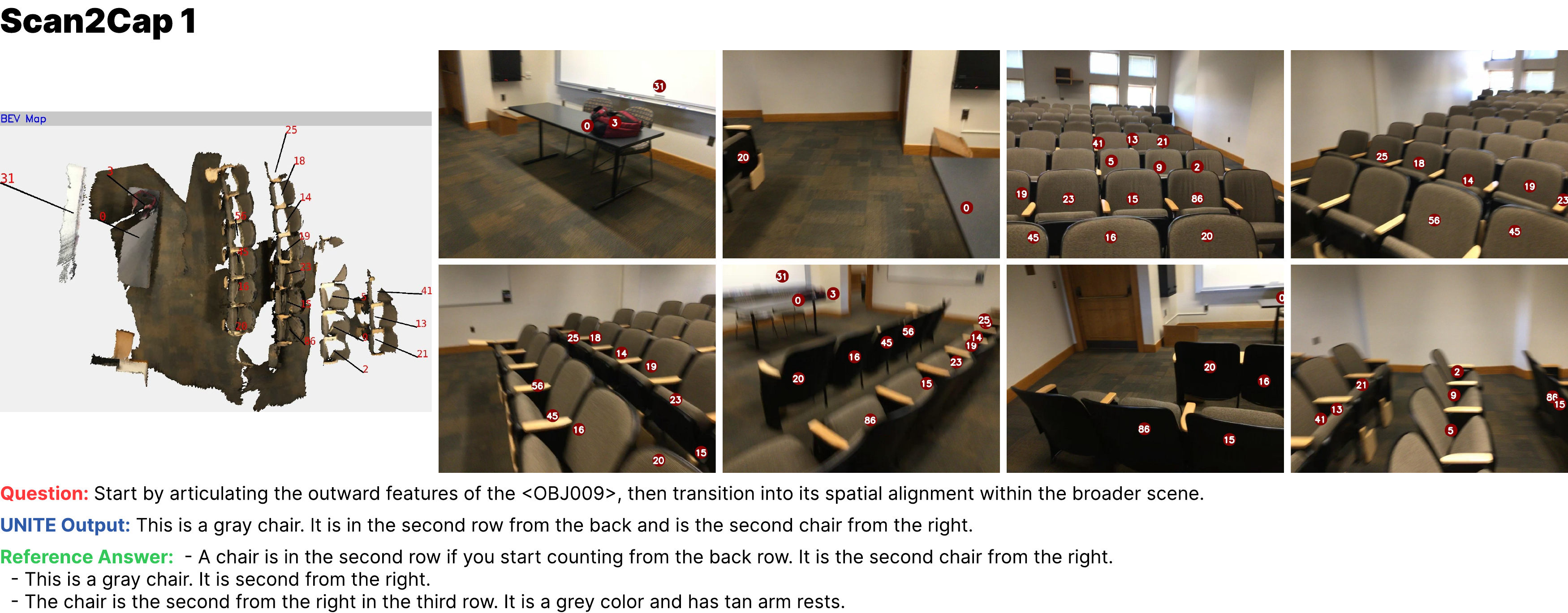}
    \par\smallskip
    \includegraphics[width=\textwidth]{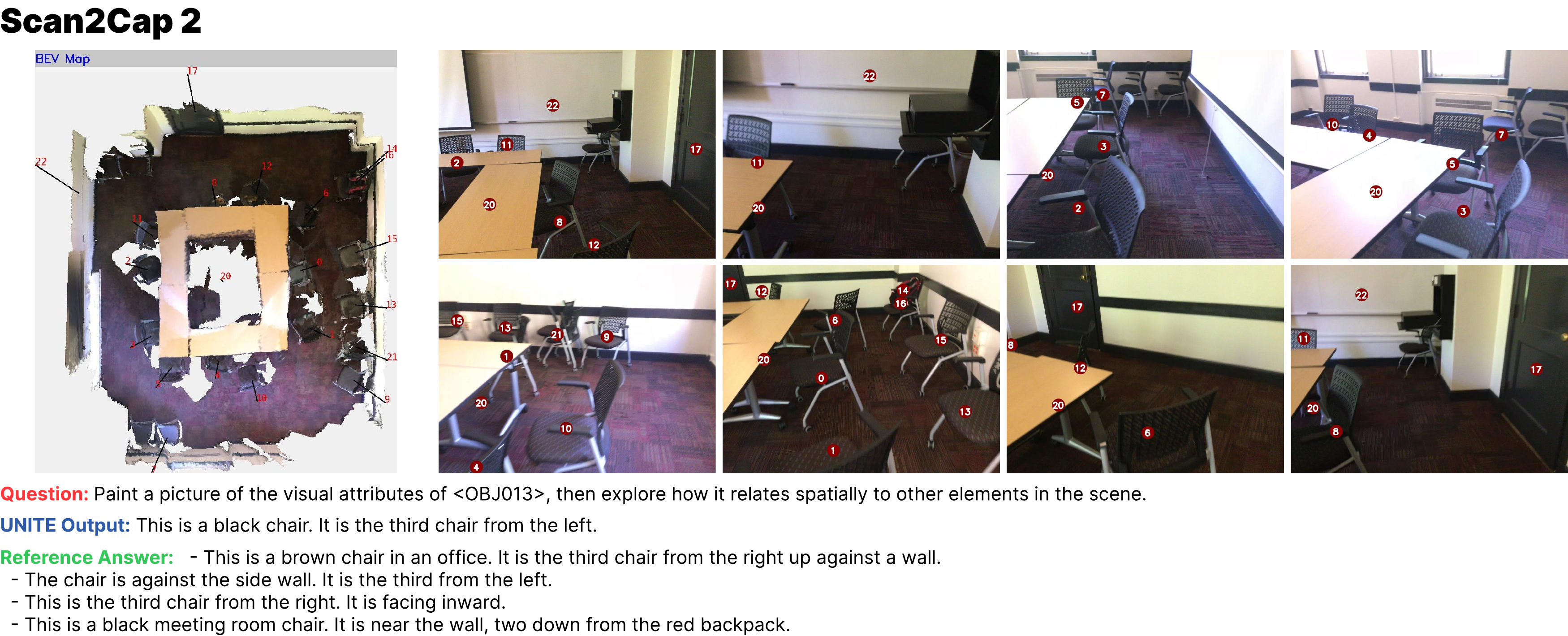}
    \end{minipage}
    \end{adjustbox}
    \caption{Qualitative examples on Scan2Cap.}
    \label{fig:qual-scan2cap}
\end{figure}

\begin{figure}[!htbp]
    \centering
    \begin{adjustbox}{max totalsize={\textwidth}{0.82\textheight},center}
    \begin{minipage}{\textwidth}
    \includegraphics[width=\textwidth]{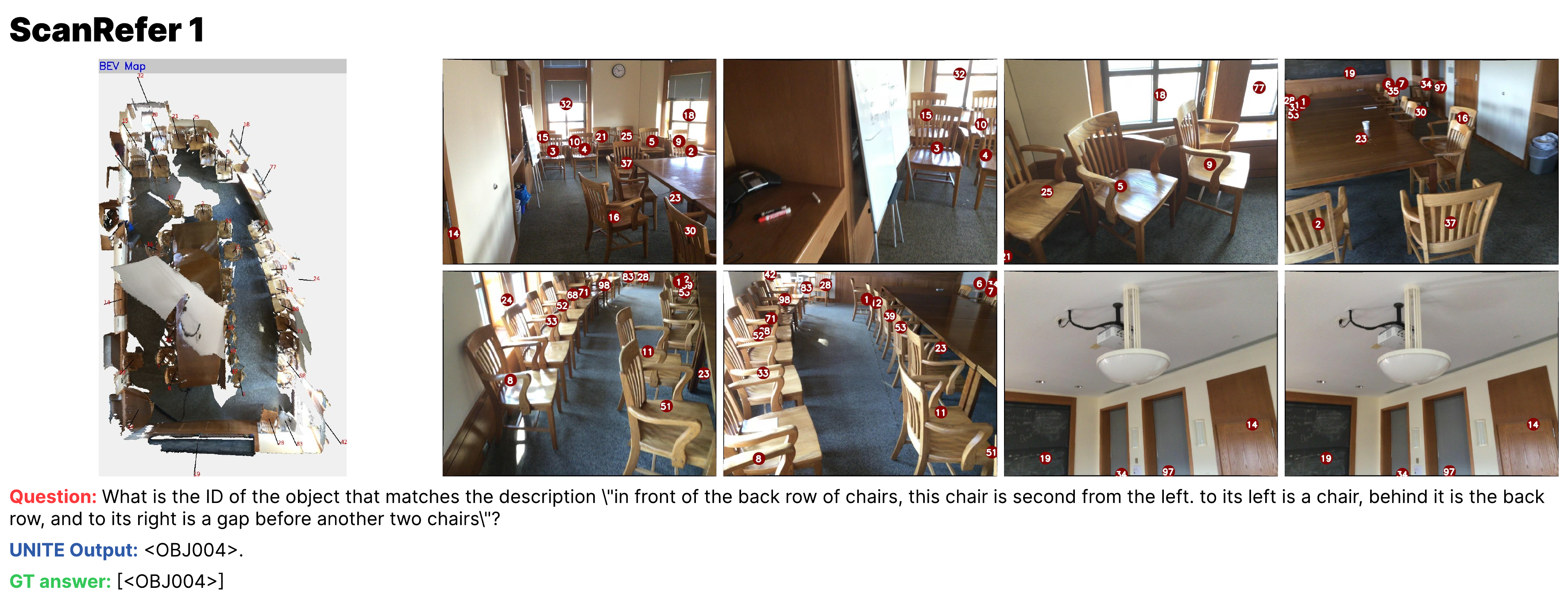}
    \par\smallskip
    \includegraphics[width=\textwidth]{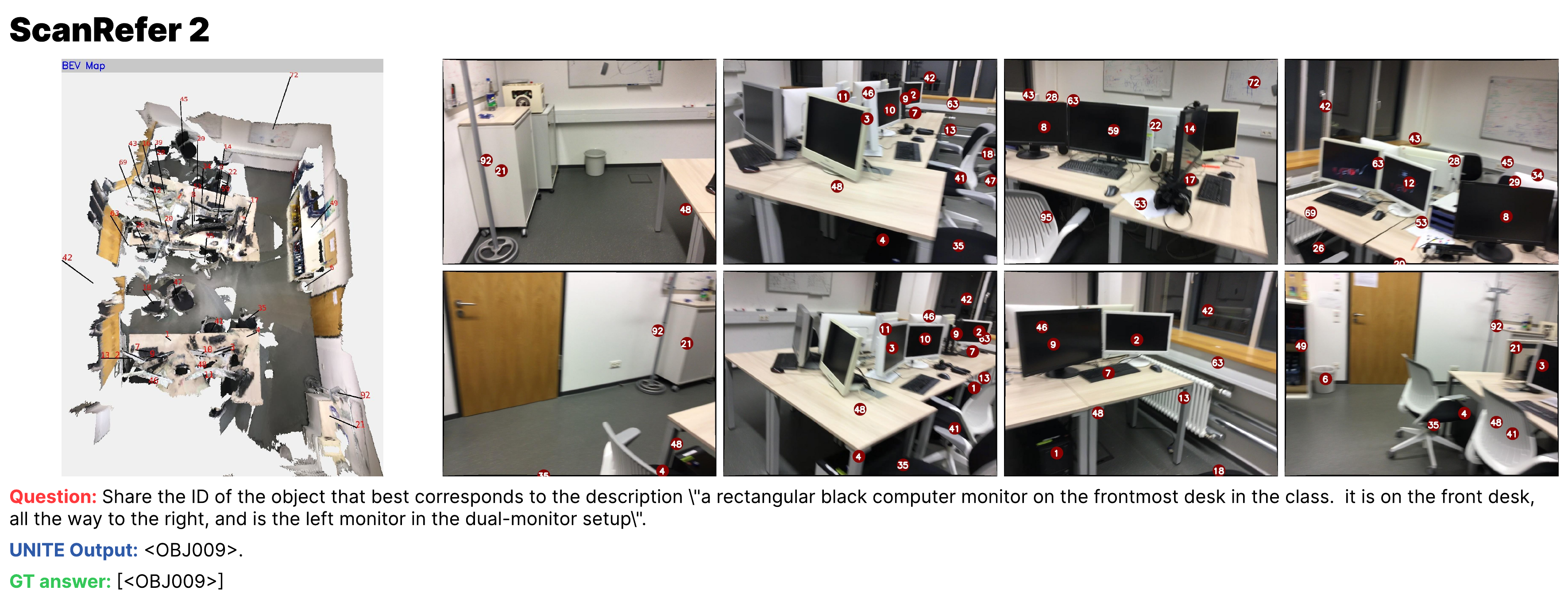}
    \end{minipage}
    \end{adjustbox}
    \caption{Qualitative examples on ScanRefer.}
    \label{fig:qual-scanrefer}
\end{figure}

\begin{figure}[!htbp]
    \centering
    \begin{adjustbox}{max totalsize={\textwidth}{0.82\textheight},center}
    \begin{minipage}{\textwidth}
    \includegraphics[width=\textwidth]{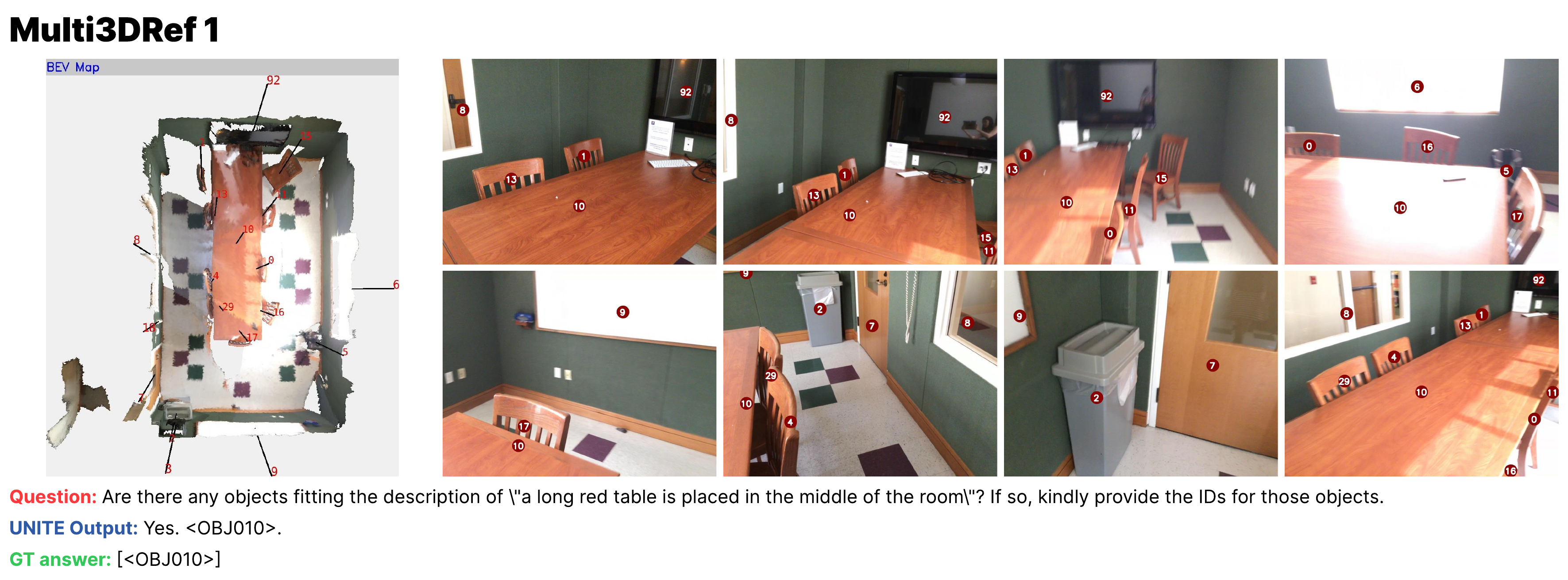}
    \par\smallskip
    \includegraphics[width=\textwidth]{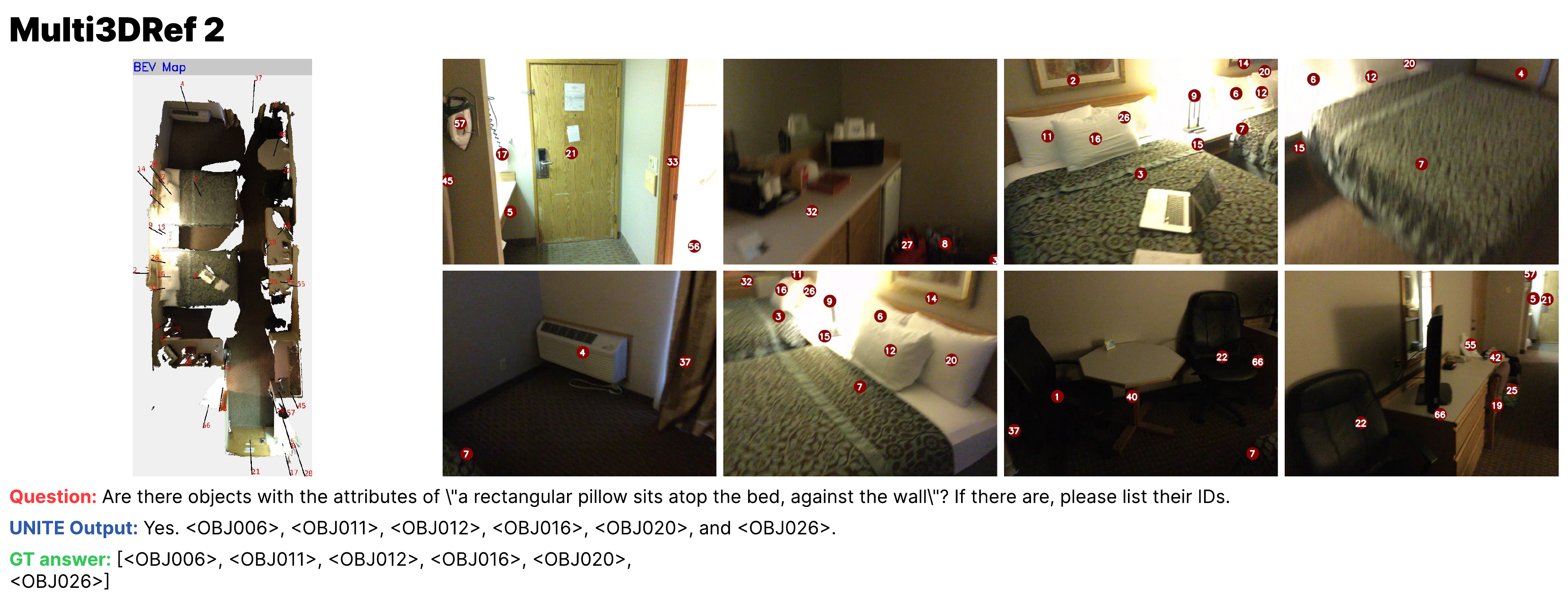}
    \end{minipage}
    \end{adjustbox}
    \caption{Qualitative examples on Multi3DRef.}
    \label{fig:qual-multi3dref}
\end{figure}

\begin{figure}[!htbp]
    \centering
    \begin{adjustbox}{max totalsize={\textwidth}{0.82\textheight},center}
    \begin{minipage}{\textwidth}
    \includegraphics[width=\textwidth]{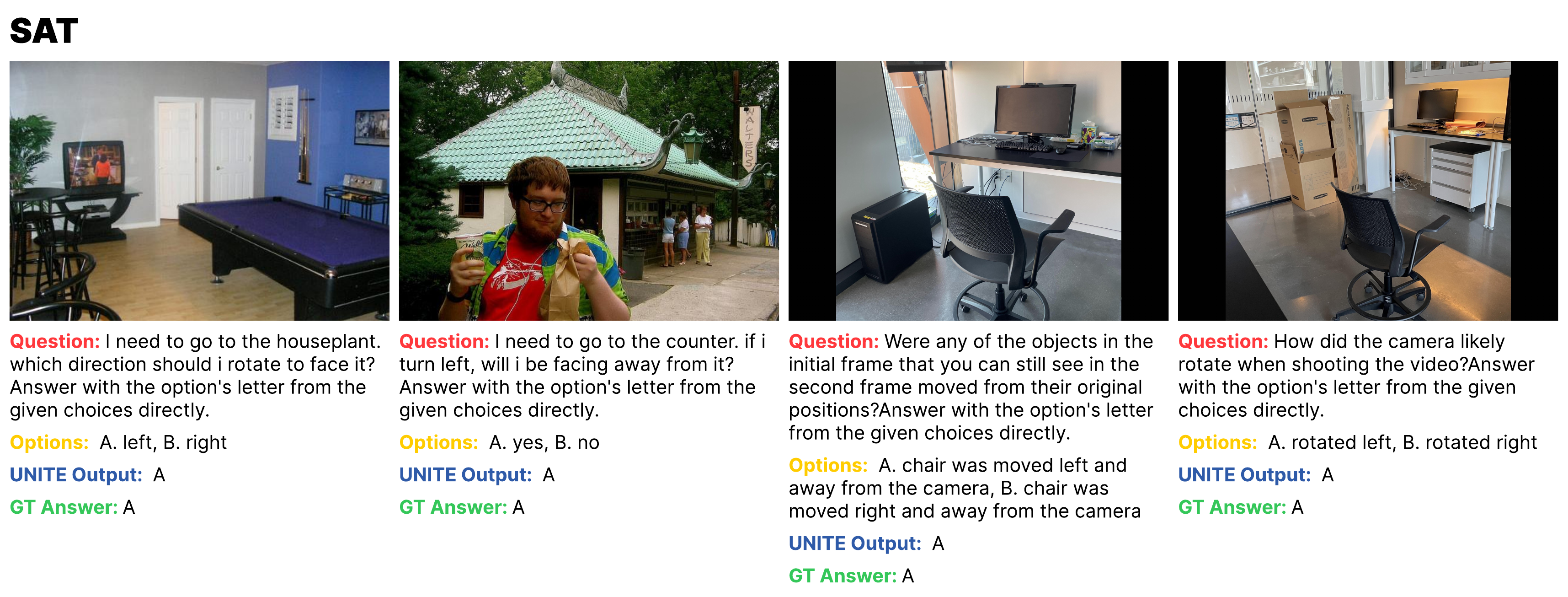}
    \end{minipage}
    \end{adjustbox}
    \caption{Qualitative examples on SAT.}
    \label{fig:qual-sat}
\end{figure}

Examples: VSI (Fig.~\ref{fig:qual-vsi}), MMSI (Fig.~\ref{fig:qual-mmsi}), MindCube (Fig.~\ref{fig:qual-mindcube}), EmbSpatial (Fig.~\ref{fig:qual-embspatial}), ERQA (Fig.~\ref{fig:qual-erqa}), ScanQA (Fig.~\ref{fig:qual-scanqa}), SQA3D (Fig.~\ref{fig:qual-sqa3d}), Scan2Cap (Fig.~\ref{fig:qual-scan2cap}), ScanRefer (Fig.~\ref{fig:qual-scanrefer}), Multi3DRef (Fig.~\ref{fig:qual-multi3dref}), and SAT (Fig.~\ref{fig:qual-sat}).

\FloatBarrier
\subsection{Driving}
\label{sec:qualitative-driving}

\begin{figure}[!htbp]
    \centering
    \begin{adjustbox}{max totalsize={\textwidth}{0.82\textheight},center}
    \begin{minipage}{\textwidth}
    \includegraphics[width=\textwidth]{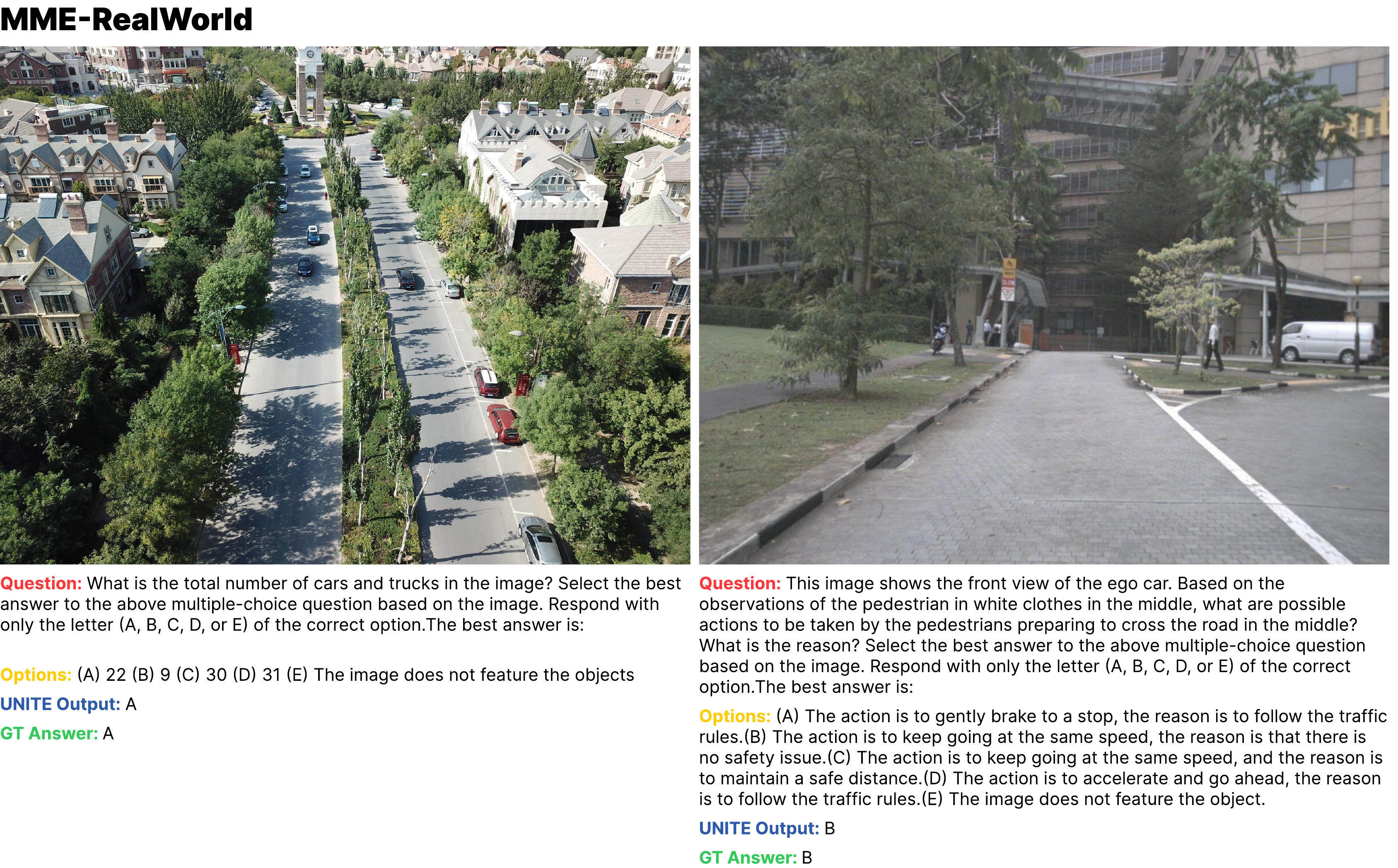}
    \end{minipage}
    \end{adjustbox}
    \caption{Qualitative examples on MME-RealWorld.}
    \label{fig:qual-mme-realworld}
\end{figure}

\begin{figure}[!htbp]
    \centering
    \begin{adjustbox}{max totalsize={\textwidth}{0.82\textheight},center}
    \begin{minipage}{\textwidth}
    \includegraphics[width=\textwidth]{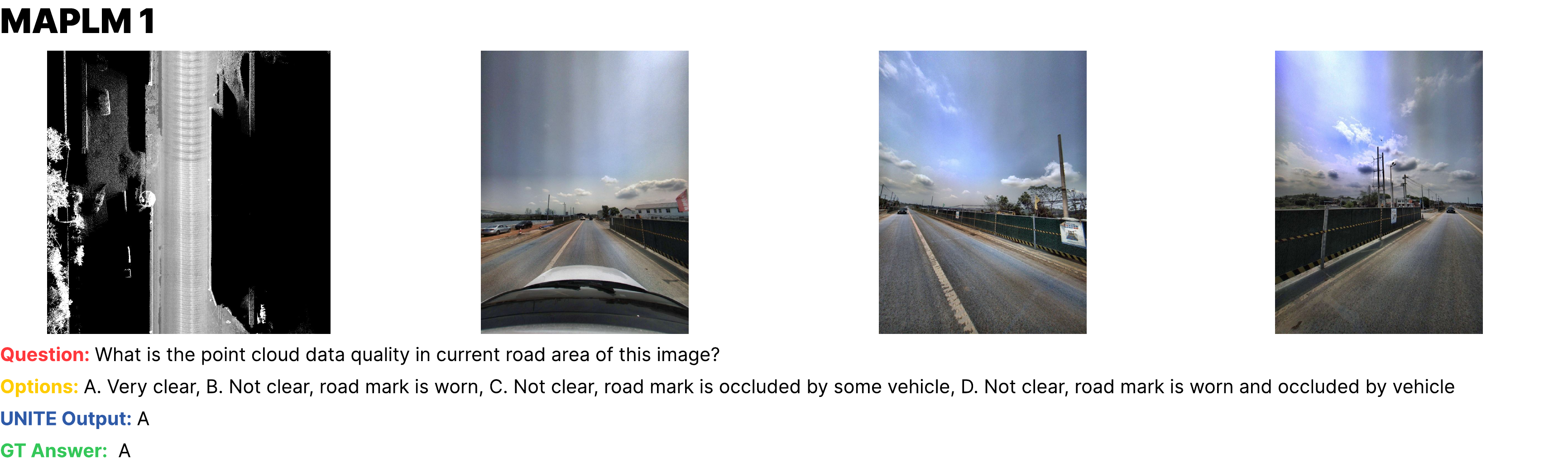}
    \par\smallskip
    \includegraphics[width=\textwidth]{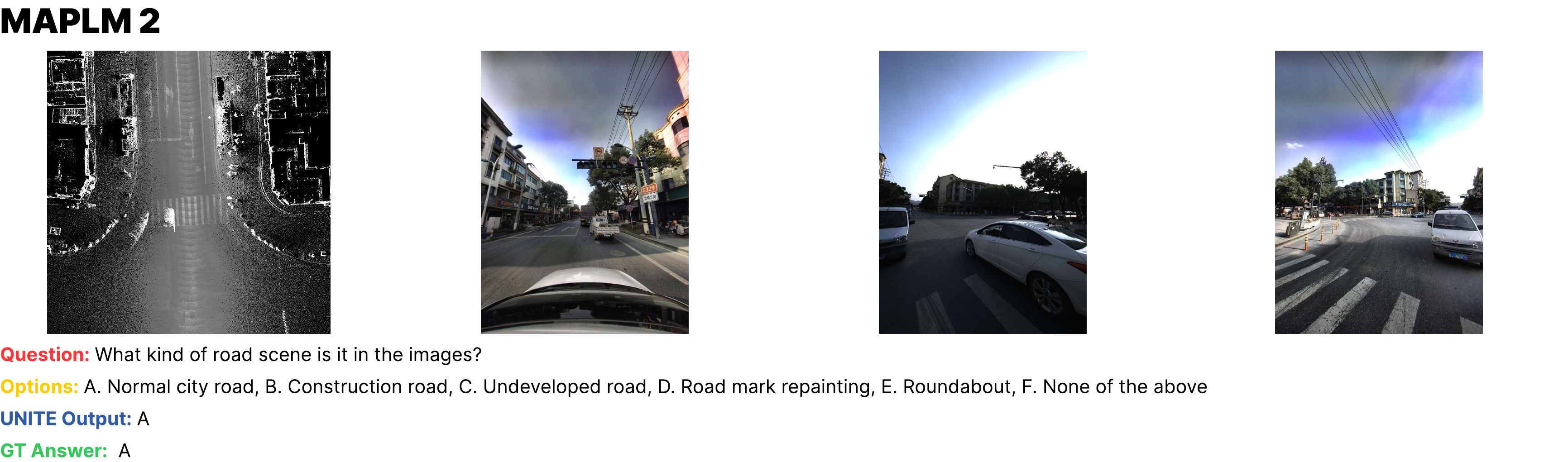}
    \end{minipage}
    \end{adjustbox}
    \caption{Qualitative examples on MAPLM.}
    \label{fig:qual-maplm}
\end{figure}

\begin{figure}[!htbp]
    \centering
    \begin{adjustbox}{max totalsize={\textwidth}{0.82\textheight},center}
    \begin{minipage}{\textwidth}
    \includegraphics[width=\textwidth]{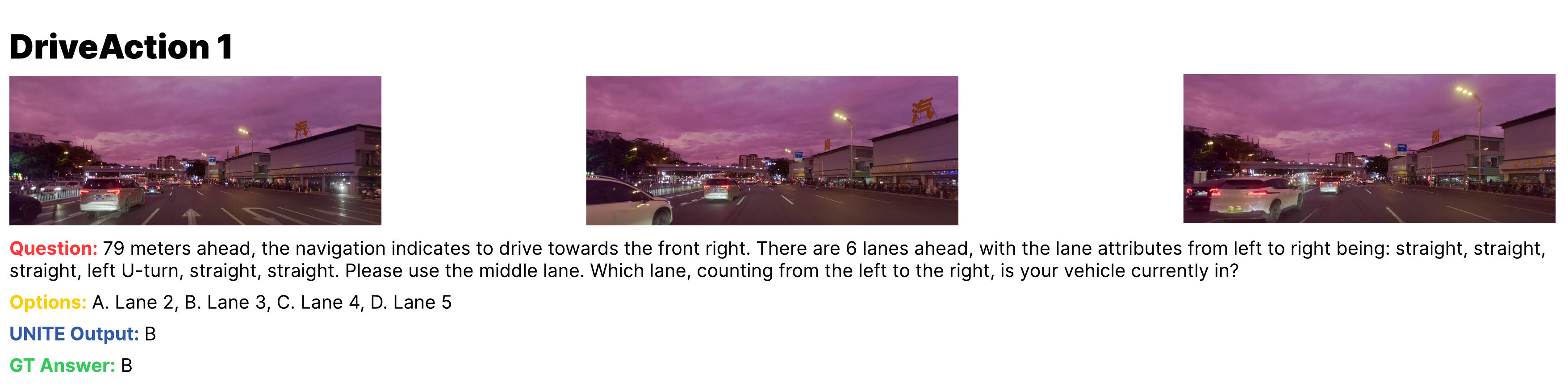}
    \par\smallskip
    \includegraphics[width=\textwidth]{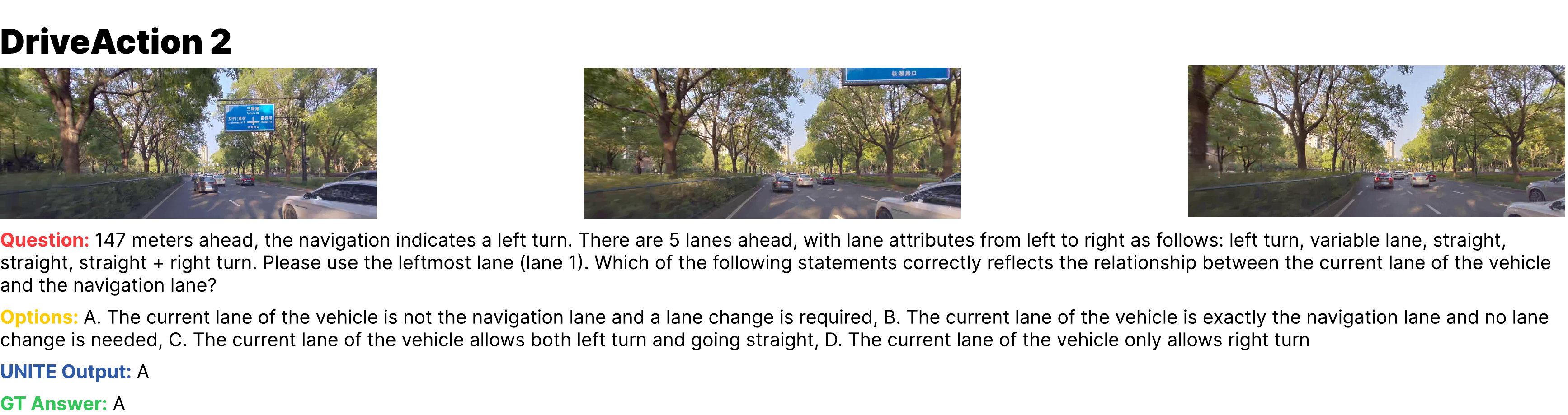}
    \end{minipage}
    \end{adjustbox}
    \caption{Qualitative examples on DriveAction.}
    \label{fig:qual-driveaction}
\end{figure}

\begin{figure}[!htbp]
    \centering
    \begin{adjustbox}{max totalsize={\textwidth}{0.82\textheight},center}
    \begin{minipage}{\textwidth}
    \includegraphics[width=\textwidth]{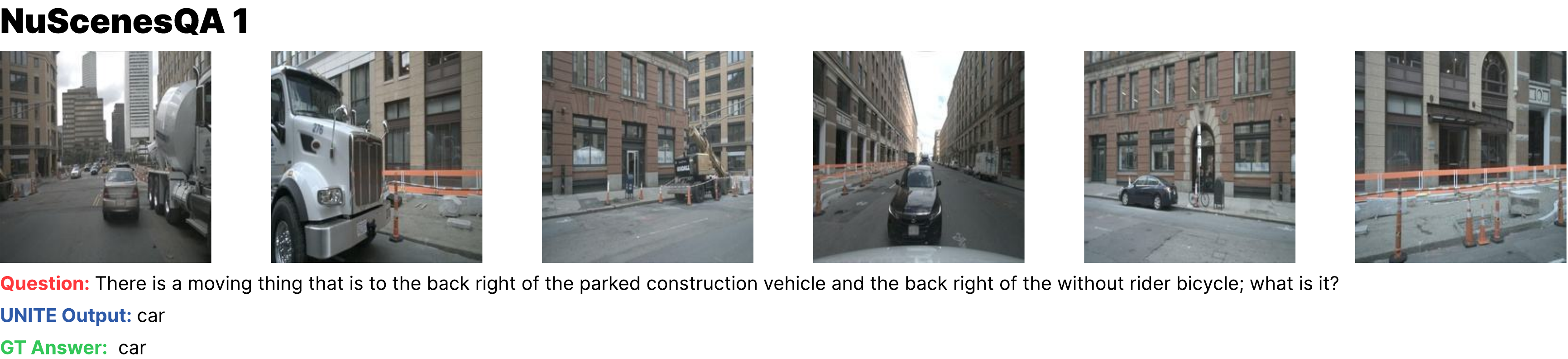}
    \par\smallskip
    \includegraphics[width=\textwidth]{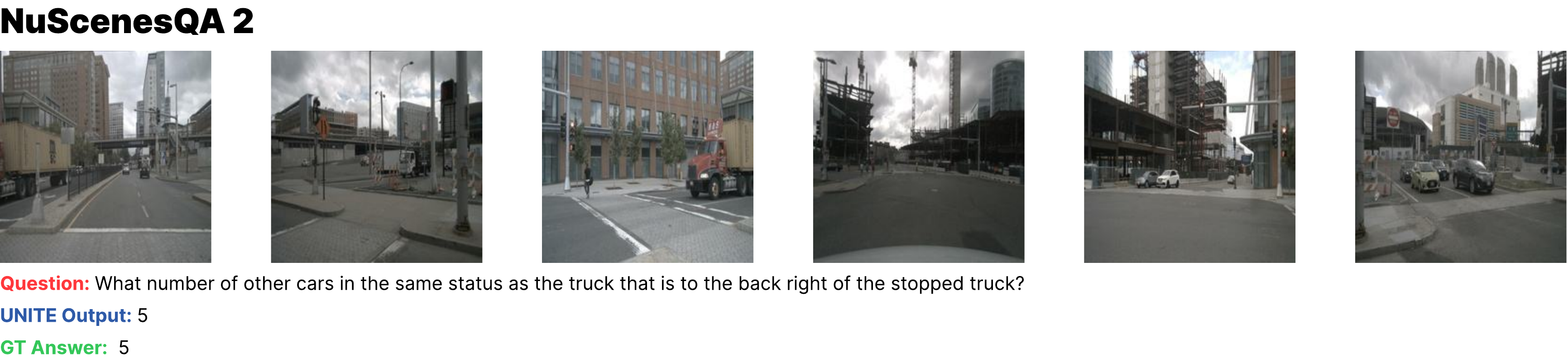}
    \end{minipage}
    \end{adjustbox}
    \caption{Qualitative examples on NuScenesQA.}
    \label{fig:qual-nuscenesqa}
\end{figure}

\begin{figure}[!htbp]
    \centering
    \begin{adjustbox}{max totalsize={\textwidth}{0.82\textheight},center}
    \begin{minipage}{\textwidth}
    \includegraphics[width=\textwidth]{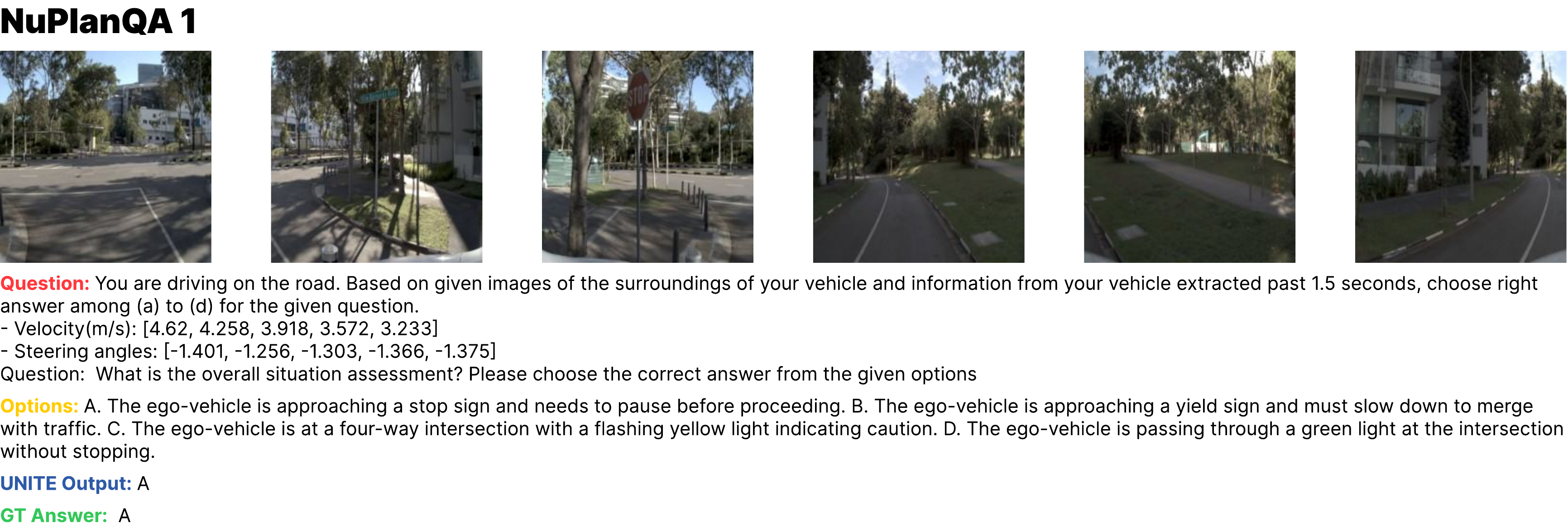}
    \par\smallskip
    \includegraphics[width=\textwidth]{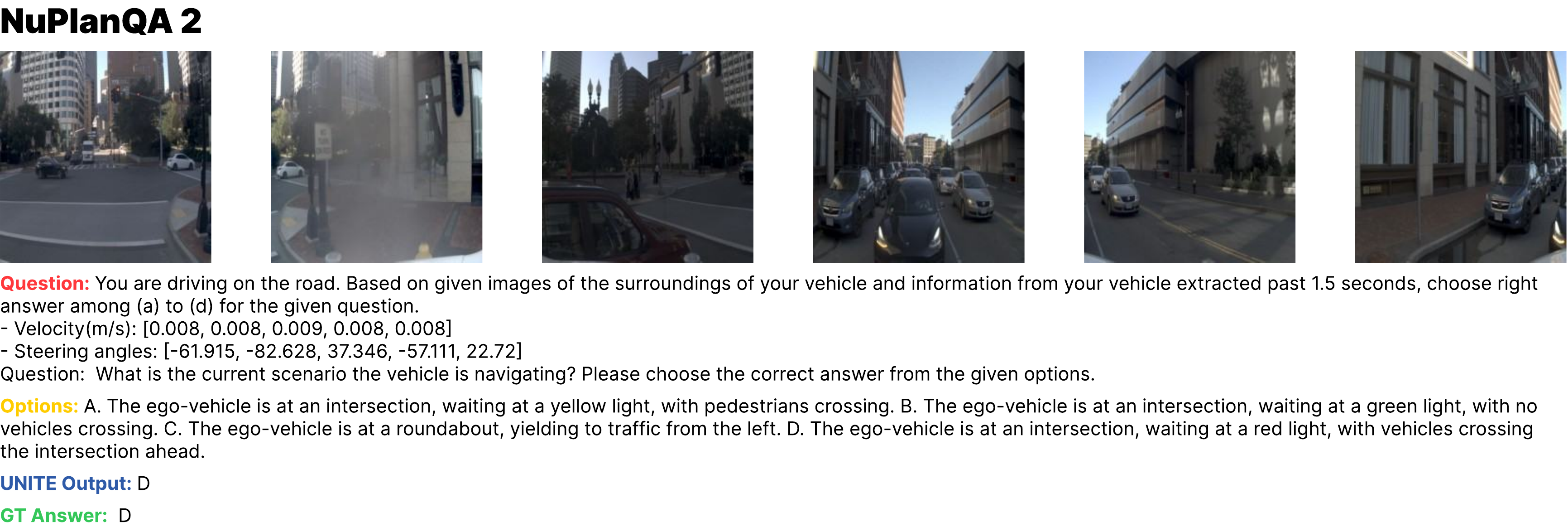}
    \end{minipage}
    \end{adjustbox}
    \caption{Qualitative examples on NuPlanQA.}
    \label{fig:qual-nuplanqa}
\end{figure}

\begin{figure}[!htbp]
    \centering
    \begin{adjustbox}{max totalsize={\textwidth}{0.82\textheight},center}
    \begin{minipage}{\textwidth}
    \includegraphics[width=\textwidth]{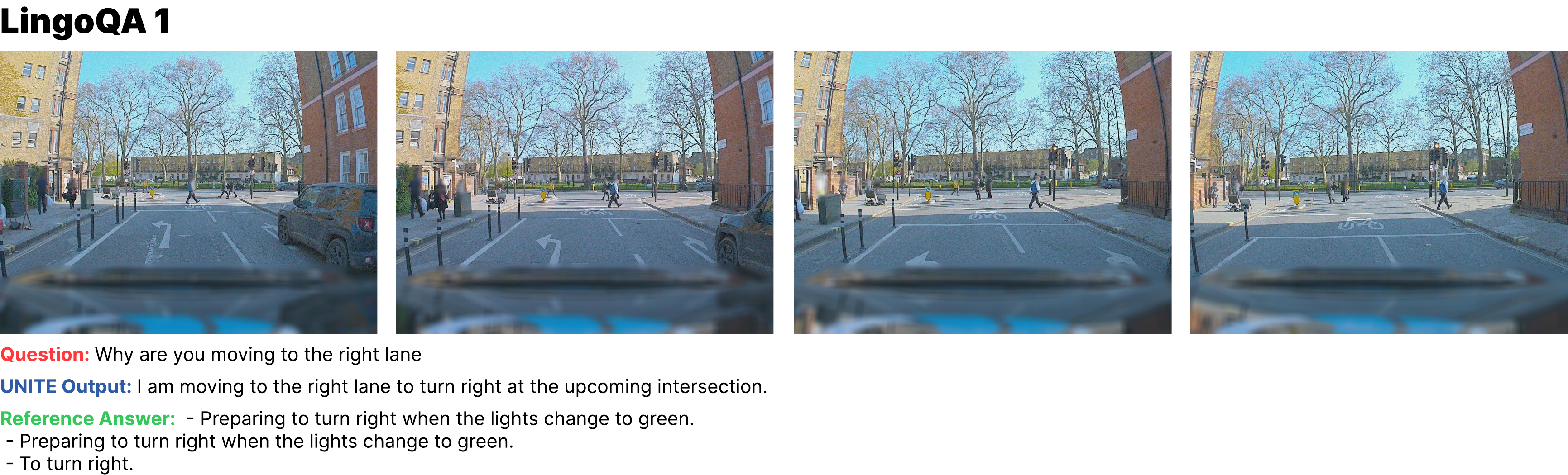}
    \par\smallskip
    \includegraphics[width=\textwidth]{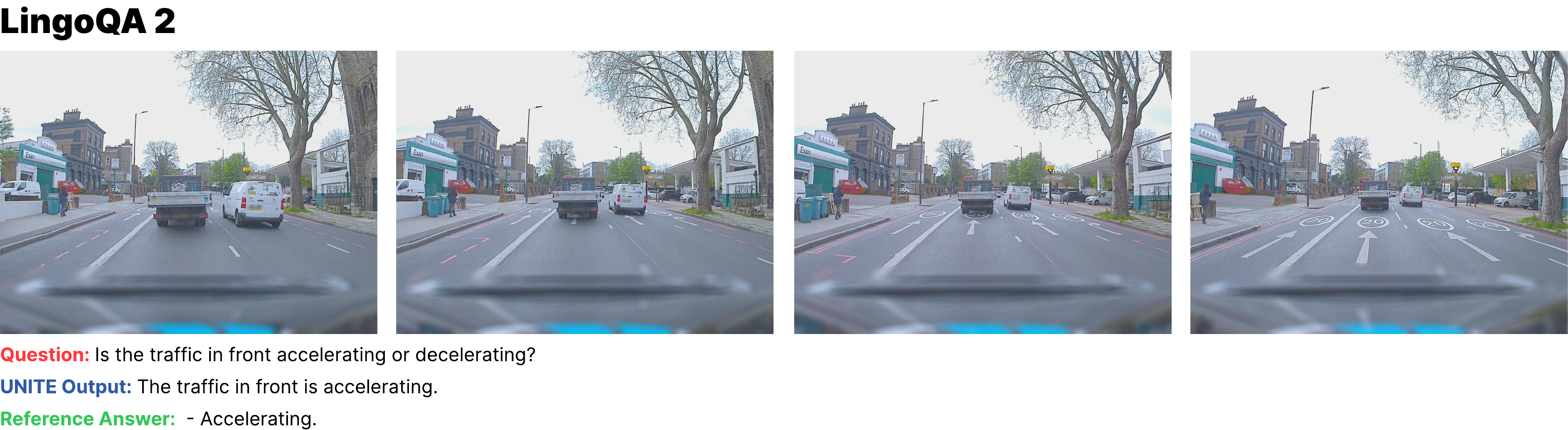}
    \end{minipage}
    \end{adjustbox}
    \caption{Qualitative examples on LingoQA.}
    \label{fig:qual-lingoqa}
\end{figure}

We show examples on MME-RealWorld (Fig.~\ref{fig:qual-mme-realworld}), MAPLM (Fig.~\ref{fig:qual-maplm}), DriveAction (Fig.~\ref{fig:qual-driveaction}), NuScenesQA (Fig.~\ref{fig:qual-nuscenesqa}), NuPlanQA (Fig.~\ref{fig:qual-nuplanqa}), and LingoQA (Fig.~\ref{fig:qual-lingoqa}).

\FloatBarrier